%% file: main.tex
\PassOptionsToPackage{x11names,dvipsnames,table}{xcolor}

\documentclass[11pt, a4paper, google]{google}

\usepackage[authoryear, sort&compress, round]{natbib}
\input{math_commands.tex}

\usepackage{dsfont}
\ifx\hmmax\undefined \chardef\hmmax=0 \fi
\ifx\bmmax\undefined \chardef\bmmax=0 \fi
\usepackage{bm}
\usepackage{mathtools}
\usepackage{mathrsfs}
\usepackage{nicefrac}
\usepackage{latexsym}

\usepackage{array}
\usepackage{multirow}
\usepackage{makecell}
\usepackage{arydshln}          
\usepackage{multicol}
\usepackage{paracol}
\usepackage{wrapfig}
\usepackage{subcaption}
\usepackage{tabu}              

\usepackage[most]{tcolorbox}
\usepackage{stackengine}
\usepackage{tikz}
\usepackage{pgffor}
\usepackage{listings}
\usepackage{algorithm}
\usepackage{algpseudocode}

\usepackage{setspace}
\usepackage{ragged2e}
\usepackage[normalem]{ulem}
\usepackage{xspace}
\usepackage{lineno}
\usepackage[textsize=tiny]{todonotes}
\usepackage{titletoc}

\usepackage[capitalize,noabbrev]{cleveref}

\graphicspath{{./}{figures/}{sections/figures/}}

\tcbuselibrary{theorems}
\newtcbtheorem[number within=section]{mytheo}{Definition}%
{colback=red!5,colframe=red!35!black,fonttitle=\bfseries}{th}

\definecolor{added}{rgb}{0,0.6,0}
\definecolor{deleted}{rgb}{1,0,0}

\newcommand{\hmap}[2]{%
    \pgfmathsetmacro{\percent}{((#1 - 1) / 4) * 100}%
    \ifdim \percent pt > 50 pt
        \def\mytextcol{white}%
    \else
        \def\mytextcol{black}%
    \fi
    \edef\temp{\noexpand\cellcolor{green!\percent!white}}\temp%
    \textcolor{\mytextcol}{$#1_{#2}$}%
}

\theoremstyle{plain}

\theoremstyle{definition}

\theoremstyle{remark}

\tcbuselibrary{breakable}

\definecolor{verylightgray}{gray}{0.9}
\definecolor{light_red}{rgb}{1.0, 0.6, 0.6}
\definecolor{light_green}{rgb}{0.56, 0.93, 0.56}
\definecolor{cldarkgreen}{rgb}{0.0, 0.6, 0.1}

\newcommand{\xmark}{\textcolor{red}{\ding{55}}}
\newcommand{\cmark}{\textcolor{OliveGreen}{\ding{51}}}

\newcommand{\xhdr}[1]{\noindent\textbf{#1.}}

\setlist[itemize]{leftmargin=*,noitemsep}

\newcommand{\leaf}{LEAF\xspace}

\uselogo{}

\title{LEAF: A Living Benchmark\\for Event-Augmented Forecasting}


\correspondingauthor{wtd3gz@virginia.edu, mihirparmar@google.com, palashgoyal@google.com}

\renewcommand{\today}{}

\author[1,2]{\fontsize{10.0pt}{10.0pt}\selectfont Mingtian Tan}
\author[1]{\fontsize{10.0pt}{10.0pt}\selectfont Mihir Parmar}
\author[1]{\fontsize{10.0pt}{10.0pt}\selectfont Palash Goyal}
\author[1]{\fontsize{10.0pt}{10.0pt}\selectfont Chun-Liang Li}
\author[1]{\fontsize{10.0pt}{10.0pt}\selectfont Nanyun Peng}
\author[2]{\fontsize{10.0pt}{10.0pt}\selectfont Thomas Hartvigsen}
\author[1]{\fontsize{10.0pt}{10.0pt}\selectfont Jinsung Yoon}
\author[1]{\fontsize{10.0pt}{10.0pt}\selectfont Tomas Pfister}

\newsavebox{\googlelogobox}
\sbox{\googlelogobox}{%
  \includegraphics[height=1.5ex]{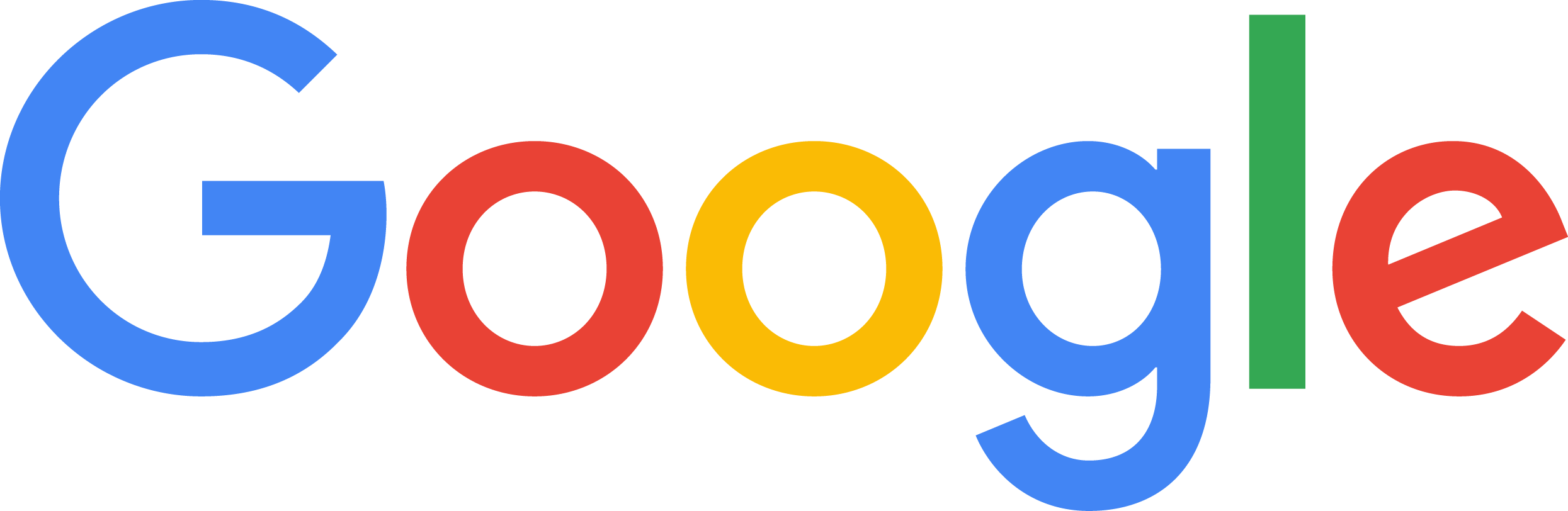}%
}
\DeclareRobustCommand{\googlelogo}{%
  \raisebox{-0.3ex}{\usebox{\googlelogobox}}%
}

\newsavebox{\uvalogobox}
\sbox{\uvalogobox}{%
  \includegraphics[height=1.7ex]{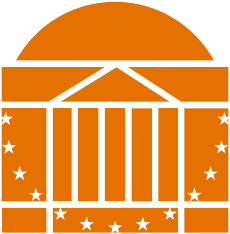}%
}
\DeclareRobustCommand{\uvalogo}{%
  \raisebox{-0.3ex}{\usebox{\uvalogobox}}%
}

\affil[1]{\fontsize{9.0pt}{9.0pt}\selectfont \protect\googlelogo\hspace{0.1em} Google Cloud AI Research}
\affil[2]{\fontsize{9.0pt}{9.0pt}\selectfont \protect\uvalogo\hspace{0.1em} University of Virginia}

\begin{abstract}
Large Language Models (LLMs) are increasingly applied to real-world forecasting tasks, yet evaluating their true predictive capability remains compromised by pre-training data contamination and look-ahead leakage in automated search. Existing benchmarks either rely on static contexts, restrict evaluations to narrow environments, or fail to audit auxiliary textual events for future information leakage. To establish a rigorous evaluation paradigm, we propose \leaf, the first living benchmark for event-augmented forecasting tasks, including trend, event, and time series forecasting. \leaf couples a recursive retrieval agent system with dual-agent cross-validation to gather comprehensive, relevant, and temporally aligned auxiliary context. A comprehensive audit across 500 tasks by 47 domain specialists demonstrates that our pipeline suppresses future information leakage from 8.6\% to 1.6\%. Across extensive evaluations of 16 frontier proprietary and open-weight models on our benchmark, we show that LLMs effectively extract signals from verified events to boost trend and event forecasting.

\end{abstract}

\begin{document}

\maketitle
\suppressfloats[t]

\input{sections/intro}

\input{sections/related}
\input{sections/benchmark}

\input{sections/experiment}

\input{sections/conclusion}
\bibliography{main}

\clearpage
\appendix
\input{sections/appendix}

\end{document}

%% file: math_commands.tex
\usepackage{amsmath,amsfonts,bm}

\def\eqref#1{equation~\ref{#1}}

\def\1{\bm{1}}

\DeclareMathAlphabet{\mathsfit}{\encodingdefault}{\sfdefault}{m}{sl}
\SetMathAlphabet{\mathsfit}{bold}{\encodingdefault}{\sfdefault}{bx}{n}



%% file: sections/intro.tex
\section{Introduction}\label{sec:introduction}
As Large Language Models (LLMs) have demonstrated increasingly sophisticated reasoning capabilities~\citep{deepmind2026gemini31pro,anthropic2026opus46,openai2026gpt54}, researchers have explored their potential in decision-making and forecasting tasks, including financial trading~\citep{xiao2025trade,fan2025ai}, event forecasting~\citep{karger2024forecastbench,halawi2024approaching}, and multimodal time series forecasting~\citep{merrill2024language,williams2024context}. Unlike deterministic reasoning benchmarks, forecasting requires predicting uncertain futures rather than extracting established facts. This probabilistic task requires interpreting weak signals within dynamic environments, serving as a challenging testbed to evaluate LLMs' temporal reasoning and decision-making under uncertainty.

While numerous benchmarks have been introduced to evaluate the predictive abilities of LLMs, existing frameworks suffer from two fundamental vulnerabilities: i) \emph{pre-training contamination in static benchmarks} and ii) \emph{look-ahead leakage in living benchmarks}. Static benchmarks~\citep{chen2025mtbench,yu2026tsrbench,liutime,li2024cryptotrade} relying on fixed historical sequences or manually curated context~\citep{williams2024context} fall within the pre-training corpora of newer LLMs. As we demonstrate, evaluating on data preceding training cutoffs artificially inflates apparent forecasting accuracy by up to 147\% (as detailed in \textsection \ref{sec:integrity}), measuring training memorization rather than true predictive reasoning. To avoid contamination, recent ``living'' benchmarks continuously update test sets~\citep{tan2024language,xu2025fidel,fan2025ai}. However, dynamically retrieving external news via automated search introduces severe look-ahead leakage, such as post-event summaries, retrospective analyses, or misdated articles that reveal post-cutoff outcomes to the model. Furthermore, current living benchmarks remain restricted to narrow domains or lack verified contextual diversity~\citep{karger2024forecastbench,li2025time}.



To address these challenges, we introduce \leaf, a living benchmark for event-augmented forecasting across diverse event-driven domains, including politico-economic events, weather, electricity, stock markets, and commodities. As illustrated in \autoref{fig:pipline}, \leaf mitigates pre-training data contamination via a continuously updated ``living'' test set, employing a retrieval and dual-agent validation system to ensure event breadth and temporal alignment while filtering out fabricated information. To validate benchmark integrity, we conducted an exhaustive evaluation across 500 tasks with 47 professional domain specialists (averaging 13.6 years of professional experience across finance, meteorology, power systems, and geopolitics). Our human-expert audit reveals that while unverified automated search exhibits an 8.6\% future look-ahead leakage rate, our dual-agent cross-validation pipeline suppresses future information leakage down to 1.6\%, with over 60\% of the remaining minor discrepancies bounded by a single day.


\begin{figure*}[t]
    \centering
    \includegraphics[width=.98\linewidth]{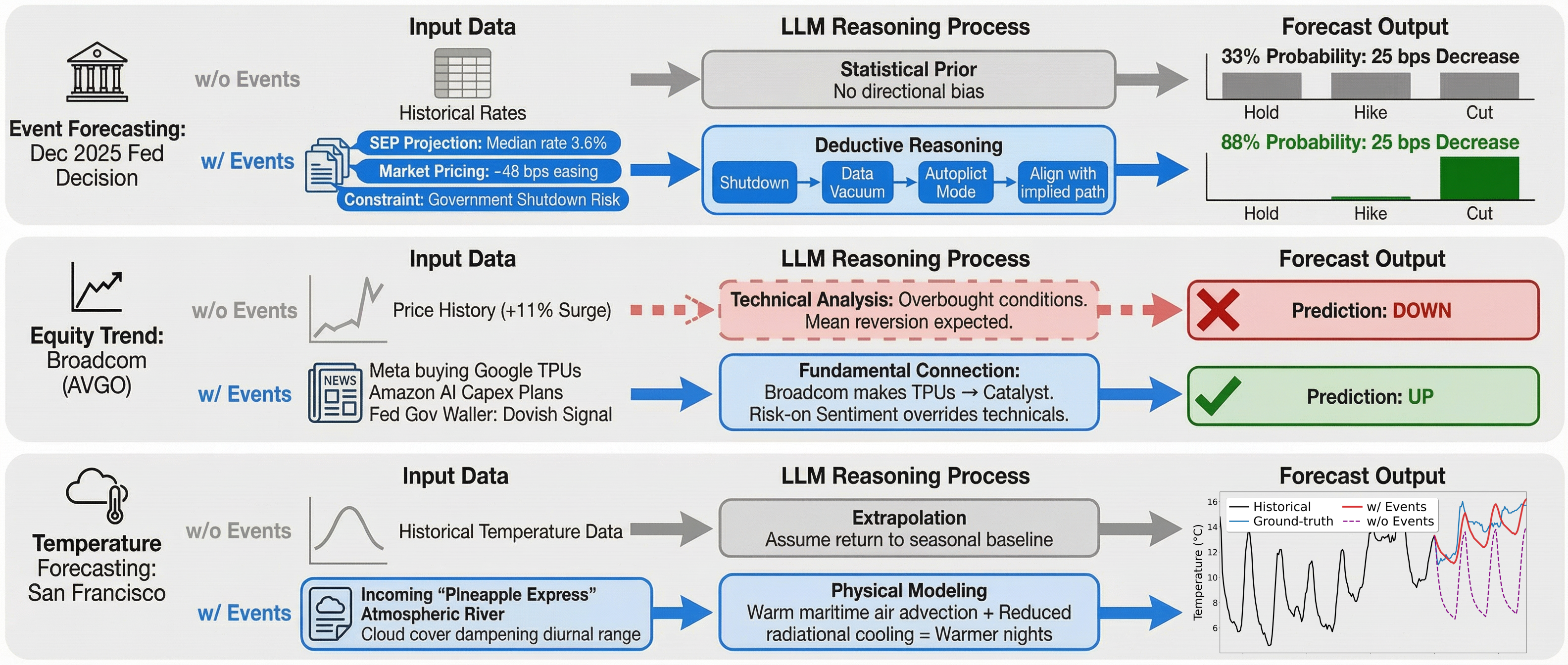}
    \caption{\textbf{Experimental case studies from \leaf} across its three core tasks: discrete event forecasting (top), trend forecasting (middle), and time-series forecasting (bottom). LLMs struggle to make effective predictions without event inputs (w/o Event); however, their performance is enhanced with events provided by our retrieval agents. Further details are provided in \autoref{fig:ef_case}, \ref{fig:trend_case}, and \ref{fig:tsf_case} in Appendix~\ref{sec:case_studies}.}
    \label{fig:context_aid_cases}
\end{figure*}

Through evaluating sixteen LLMs, including proprietary (Gemini-, Grok-, Claude-Family) and open-weight models (Qwen-, Gemma-Family), we find that proprietary models effectively extract signals from verified events, achieving relative gains of 30.03\% in trend forecasting  and 25.02\% in event probability forecasting over event-free baselines (\textsection \ref{sec:event_trend_forecast_results}), and illustrative examples are provided in \autoref{fig:context_aid_cases}. Our main contributions are:


\begin{itemize}
\item We introduce \leaf, a living benchmark to evaluate LLMs' ability to leverage multidimensional signals for event-driven forecasting in complex environments (\textsection\ref{sec:problem_define}).

\item We propose a multi-agent retrieval system to collect events, augmented by a dual-agent cross-validation pipeline that ensures authenticity and temporal alignment (\textsection\ref{sec:method}), which significantly improves event date accuracy and mitigates look-ahead leakage (\textsection\ref{sec:integrity}).


\item Evaluating 16 LLMs reveals fundamental limitations: even with informative text, most models perform merely on par with time-series foundation models (TSFMs) and fail to outperform naive baselines in market price forecasting, empirically confirming the latter task's infeasibility (\textsection\ref{sec:tsf_results}).

\item In market trend predictability tests, despite the Efficient Market Hypothesis (EMH)~\citep{fama1970efficient}, we observe that LLMs exhibit capacity to distinguish between strong signals and those already priced in, yielding forecasting improvements in high-confidence scenarios (\textsection\ref{sec:predictability_validations}).
\end{itemize}

%% file: sections/related.tex
\begin{table}[htbp]
\centering
\small
\caption{Comparison of forecasting benchmarks. Our benchmark includes \textbf{three forecast tasks}: event (EF), time series (TSF), and trend forecasting. Instead of relying on APIs or static datasets for text conditions, we employ a \textbf{multi-agent system} to ensure comprehensive text coverage. We further compare the benchmarks based on Diversity (\textbf{Div}): whether the event incorporates various sentiments and from multiple sources; Quality Control (\textbf{QC}): whether textual relevance is validated; and whether the benchmark is \textbf{Living}.}
\label{tab:benchmark_comparison}
\begin{tabular}{ll p{4cm} c c c}
\toprule
\textbf{Benchmark} & \textbf{Forecast} & \textbf{Context Source} & \textbf{Div} & \textbf{QC} & \textbf{Living} \\
\midrule
\cite{karger2024forecastbench} & EF             & API (e.g., \cite{polymarket2026api}) & \xmark   & \xmark   & \cmark \\
\cite{fan2025ai}               & Trend          & API (e.g., \cite{jinaai2026reader})  & \cmark & \xmark   & \cmark \\
\cite{nof1ai2026alphaarena}    & Trend          & None (Season 1 \& 2)                 & \xmark   & \xmark   & \cmark \\
\cite{xu2025fidel}             & TSF            & Weather API                          & \cmark & \xmark   & \cmark \\
\cite{yu2026tsrbench}          & TSF            & Records \& Synthetic                 & \cmark & \xmark   & \xmark   \\
\cite{chen2025mtbench}         & TSF \& Trend     & Records \& Dataset                   & \xmark   & \cmark & \xmark   \\
\cite{williams2024context}     & TSF            & Manually Crafted                     & \xmark   & \cmark & \xmark   \\
\cite{tan2025inferring} & EF \& TSF & API & \xmark & \cmark & \cmark \\
\midrule
\leaf (ours)                  & EF \& TSF \& Trend & Multi-Agent System                   & \cmark & \cmark & \cmark \\
\bottomrule
\end{tabular}
\end{table}

\section{Related Work}\label{sec:related}


\xhdr{Forecasting with LLMs} Many studies employ LLMs for time series forecasting based solely on historical numerical data~\citep{gruver2023llmtime,zhou2024one,jin2023time}, or incorporate textual information~\citep{wang2024newsforecast,luo2025time}. For instance, \citet{williams2024context} uses manually crafted expected future events to assist LLMs in predicting time series trends and \citet{merrill2024language} uses synthetic text and time series data for context-aided forecasting. Other studies combine LLMs with news to predict the probability of future events~\citep{halawi2024approaching,yang2026llm}. \citet{karger2024forecastbench} leverages odds from prediction platform like Polymarket~\citep{polymarket2026api} to aid LLM forecasting. \citet{yang2026llm} utilizes GPT-4o to retrieve the top-$k$ news articles to aid LLM live predictions. Additionally, some works directly deploy LLMs for forecasting and trading in markets~\citep{nof1ai2026alphaarena,fan2025ai,li2025time}. However, the textual information provided in these works is often synthesized or constrained by narrow data sources (e.g., specific APIs), which struggles to guarantee comprehensiveness. Moreover, some existing forecasting benchmarks are static, making it difficult to prevent data contamination~\citep{chen2025mtbench,liutime,yu2026tsrbench}. To evaluate LLMs' ability to extract informative signals from complex contexts for event-driven forecasting, we propose a living benchmark featuring comprehensive event data.

\xhdr{Living Benchmarks} To mitigate data contamination, several living benchmarks dynamically construct test sets to evaluate general model performance. For instance, LMSYS~\citep{zheng2024lmsys} continuously ranks LLMs via crowdsourced human pairwise evaluations, while LiveBench~\citep{livebench} regularly updates its questions using recent objective data to prevent memorization. Extending this dynamic paradigm, other benchmarks focus on specific capabilities, such as reasoning~\citep{fan2023nphardeval,glazer2024frontiermath}, coding~\citep{jain2024livecodebench,zheng2025livecodebench}, and multimodal reasoning~\citep{zou2024dynamathdynamicvisualbenchmark,bitton2023visitbench,yang2024dynamic}. Living benchmarks and evaluations, as summarized in \autoref{tab:benchmark_comparison}, have also been proposed for forecasting tasks~\citep{nof1ai2026alphaarena,karger2024forecastbench,li2025time,tan2025inferring}. For instance, \citet{xu2025fidel} evaluates LLMs in predicting power generation or traffic flow using weather forecasts and synthetic sensor data. \citet{fan2025ai} collects real-time news via APIs for market forecasting. Additionally, \citet{williams2024context} incorporates manually crafted text for time-series forecasting, relying on human annotation to ensure relevance. However, constructing comprehensive context-aided test sets in complex open environments remains challenging due to data scarcity. To bridge this gap, we introduce a real-world dataset and empirically validate the auxiliary text's relevance.


%% file: sections/benchmark.tex
\section{LEAF: A Living Benchmark}\label{sec:benchmark}



\subsection{Problem Definition}\label{sec:problem_define}
Let $X_{\le T} = \{x_1, x_2, \dots, x_{T}\}$ denote a real-valued historical time series observed up to the present timestamp $T$. In traditional time series analysis, forecasting models often predict future states by relying solely on historical numerical sequences. 
Our living event-augmented forecasting benchmark evaluates LLMs' ability to extract informative signals from historical values and events for forecasting. Specifically, the models are conditioned on real-valued inputs $X_{\le T}$ (when available) and historical events $\mathcal{C}_{\le T}$ to perform three primary prediction tasks: 

\begin{itemize}
    \item \textbf{Event Forecasting (EF)} evaluates the LLM's ability to predict the probability of a discrete future event $P(E_{T+h})$, scheduled to resolve at a future timestamp $T+h$, using available priors. Our evaluation consists exclusively of textual events.
    \item \textbf{Time Series Forecasting (TSF)} evaluates the ability to predict exact future values ${X}_{T+1:T+h}$ over a horizon $h$, conditioned on both numerical and textual histories $(X_{\le T}, \mathcal{C}_{\le T})$.
    \item \textbf{Trend Forecasting (TF)} converts exact numerical prediction into a categorical task, given that LLMs excel at logical classification over precise regression. The model predicts the future movement direction $y_{T+1}$ as a 3-class or 5-class classification using the prior $(X_{\le T}, \mathcal{C}_{\le T})$, where the ground truth is derived from the differential $\Delta x = x_{T+1} - x_{T}$. Details refer to Appendix \ref{sec:trend_boundary}. 
    

\end{itemize}

\begin{figure*}[t]
    \centering
    \includegraphics[width=.98\linewidth]{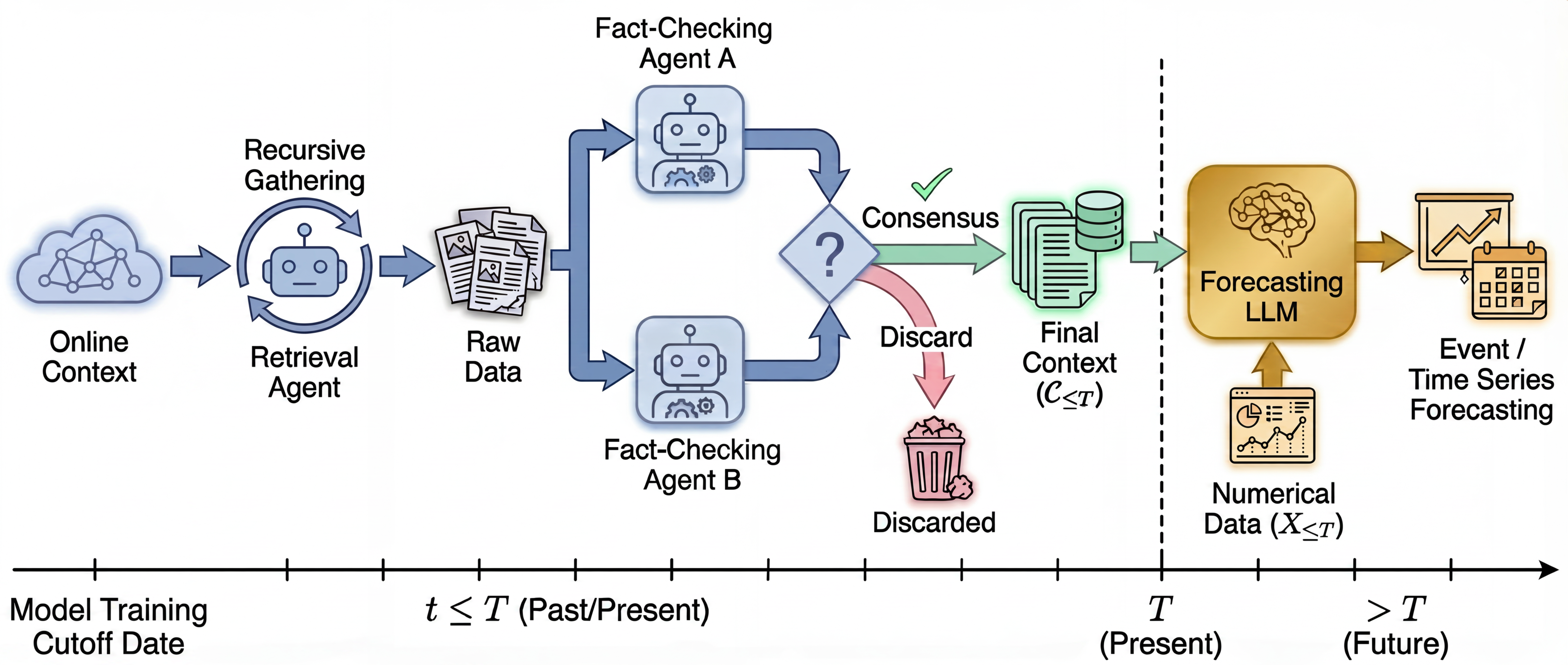} 
    \caption{The data retrieval and cross-validation framework enables predictive evaluation in realistic and complex environments. To provide events relevant to the forecasting target, an LLM-based \emph{Retrieval Agent} recursively gathers temporally constrained online context ($t \le T$). A dual-agent \emph{Fact-Checking Module} then filters this data, discarding unverified information and retaining only events with a strict consensus on both publication time and factual authenticity. The test data is restricted to periods after the models' training cutoffs to prevent pre-training data leakage.}
    \label{fig:pipline} 
\end{figure*}


\begin{table*}[htbp] 
\centering
\small
\setlength{\tabcolsep}{5pt}
\caption{Evaluation data statistics in \leaf across domains, entities, and forecasting tasks. We report the number of raw events prior to filtering, the filtering rate of the dual-agent fact-checking module, the number of days with active events (\# Days), the evaluation frequency, and the evaluation horizon.}
\label{tab:dataset_statistics}
\resizebox{\textwidth}{!}{%
\begin{tabular}{l c c c c c c c}
\toprule
\textbf{Domain} & \textbf{Tasks} & \textbf{\# Entities} & \textbf{\# Events} & \textbf{Filter Rate} & \textbf{\# Days} & \textbf{Forecast Freq} & \textbf{Horizon} \\
\midrule
Polymarket & EF & 66 & 12,162 & 39.5\% & 2,796 & 10\% & 04.2026--09.2026 \\
Large-cap Stocks    & Trend, TSF & 30 & 37,343 & 41.9\% & 3,959 & Daily & 04.2026--09.2026 \\
Mid-cap Stocks      & Trend, TSF & 40 & 13,427 & 51.7\% & 3,136 & Monthly & 03.2026--08.2026 \\
Commodities         & Trend & 10 & 4,240  & 72.8\% & 509   & Daily & 10.2025--02.2026 \\
Electricity         & TSF & 3  & 1,518  & 26.3\% & 353   & Hourly/Daily & 12.2025--04.2026 \\
Weather             & TSF & 10 & 5,271  & 24.0\% & 1,177 & Hourly/Daily & 10.2025--02.2026 \\
\bottomrule
\end{tabular}%
}
\end{table*}

\subsection{Methodology}\label{sec:method}
\xhdr{Retrieval Agent and Fact-checking Module}
In our benchmark, the textual context is gathered recursively by an LLM-based retrieval agent. This agent is augmented with Gemini Grounded Search\footnote{\url{https://ai.google.dev/gemini-api/docs/google-search}}. As illustrated in~\autoref{fig:pipline}, for each specified time interval (e.g., daily or weekly), we collect all published information relevant to the forecasting target. To ensure comprehensiveness and avoid duplication, we conduct multiple search iterations, appending previously retrieved event as context for subsequent rounds. Beyond detailed \emph{event descriptions} and \emph{reference date}, the retrieval agent outputs a ``\emph{causality}'' rationale explaining why the event affects the target, a ``\emph{sentiment}'' label indicating the impact direction (Positive or Negative), and an ``\emph{impact\_type}'' classification (Direct or Indirect). Furthermore, to ensure data authenticity and temporal alignment, we utilize a dual-agent cross-validation mechanism. Retrieved events are retained to form the final context if and only if two independent fact-checking agents reach a strict consensus regarding their authenticity and public release time; any discordant information is discarded. For the detailed searching and fact-checking prompt, refer to Appendix \ref{sec:forecast_prompts} (\autoref{fig:stock_event_search_prompt} and \ref{fig:fact_check_prompt}).

\xhdr{Human Expert Data Annotation} We engaged 47 domain experts (averaging 13.6 years of experience across finance, meteorology, energy, geopolitics, etc.) to annotate 500 events. 
With the implementation of the fact-checking agent, only 1.6\% (8 events) exhibited future information leakage, and five of these cases preceded the actual date by only a single day, which has limited impact given our 3-day and 7-day forecasting horizons. In contrast, without the fact-checking agent, 43 events suffered from future information leakage. Additionally, experts identified 1.6\% of the events as hallucinations. For further details on the expert annotations, refer to Appendix \ref{app:expert_anno}.

\xhdr{Stock Market Problem Formulation}
Stock market forecasting is notoriously challenging under the Efficient Market Hypothesis (EMH)~\citep{fama1970efficient}, as most public events are rapidly priced in. Nevertheless, we include equities in \leaf, hypothesizing that rare, strong signals persist that are not immediately absorbed or require time to unfold. To determine whether LLMs can distinguish between these strong signals and priced-in events, we present the models with all stocks at each prediction step, conditioned on $(X_{\le T}, \mathcal{C}_{\le T})$. The LLMs rate the signal strength of these 30 (40) stocks on 5 levels. We observe that higher LLM confidence correlates with better performance (\textsection\ref{sec:predictability_validations}). This improvement is evident in trend forecasting for short-term (1 or 2 days) large-cap stocks and long-term (3 or 6 months) mid-cap stocks, though accurately predicting specific prices remains infeasible (\textsection\ref{sec:tsf_results}). Detailed prompts and classification criteria are provided in \autoref{fig:predictability_test_prompt} in the Appendix \ref{sec:forecast_prompts}.

\xhdr{Retrieval Events Validations}
To validate our event retrieval pipeline, we construct a "consistent" textual context to assist predictions during evaluation, similar to~\citet{chen2025mtbench} in stock domain. Since the ground-truth change $\Delta x = x_{T+h} - x_T$ is known during benchmark formulation, we use it to systematically select events $e \in \mathcal{C}_{\le T}$ whose expected ``\emph{sentiment}'', from retrieval agent, aligns with the actual future trajectory. 

Specifically, we control the mixture of the provided context using a consistency ratio, $\alpha \in [0, 1]$. For instance, if the time series exhibits a downward trend in the future ($\Delta x < 0$), we assemble a context where an $\alpha$ proportion consists of ``consistent'' (negative sentiment) events, and the remaining $1-\alpha$ proportion consists of ``inconsistent'' events. We then condition the LLM on this controlled context to perform the forecasting task. If the forecasting error monotonically decreases as the consistency ratio $\alpha$ increases, it empirically validates two critical aspects: (1) the retrieved textual events are genuinely correlated with the time series dynamics, and (2) our annotated ``\emph{sentiment}'' labels are accurate. To verify this, evaluating $\alpha \in \{0, 0.2, 0.4, 0.6, 0.8, 1.0\}$ achieves a $-0.967$ correlation, which drops to $-0.537$ in an ablation study using sentiment-consistent text from alternative targets.



\subsection{Domains and Retrieval}
\xhdr{Domains} Our living benchmark focuses on event-driven forecasting targets that are inherently volatile and continuously shaped by real-world occurrences~\citep{karger2024forecastbench,chen2025mtbench,fan2025ai,li2025time}. Specifically, as listed in~\autoref{tab:dataset_statistics}, we cover \textbf{1) Stock Market}, tracking both large-cap (e.g., AAPL, NVDA) and mid-cap equities (e.g., HOOD, NET); and \textbf{2) Commodities}, monitoring raw materials influenced by global supply chains and geopolitics (e.g., GC=F, CL=F). Furthermore, we include \textbf{3) Electricity} and \textbf{4) Meteorological} time series for major cities (e.g., New York, Chicago) to reflect complex physical systems, alongside \textbf{5) Event Forecasting} for real-world predictive tasks spanning economics, politics, and technology (e.g., elections, interest rate decisions). For detailed descriptions and examples of each domain, please refer to Appendix \ref{sec:domain_details} and \ref{sec:taxonomy}.

\xhdr{Data Retrieval} \leaf gathers at least one quarter of event data for each domain. Specifically, as detailed in \autoref{tab:dataset_statistics}, we collect target-relevant events daily for daily forecasting domains, and weekly for both monthly and event forecasting tasks. For time series data, we utilize the Yahoo Finance API~\citep{aroussi2026yfinance} for real-time stock and commodity pricing, the Meteostat API~\citep{meteostat2026python} for weather observations, the Grid Status API~\citep{kanter2026gridstatus} for electricity load data and the Polymarket API~\citep{polymarket2026api} to track trending predictive events and outcomes. Finally, our retrieval agent is powered by Gemini 3.1 Pro~\citep{deepmind2026gemini31pro}, while the dual fact-checking agents are built upon Gemini 3.1 Pro and Claude 4.6 Sonnet~\citep{anthropic2025claudesonnet45}.

\subsection{Benchmark Integrity and Expert Verification}\label{sec:integrity}

A core vulnerability of static and automated forecasting benchmarks is evaluation failure driven by either pre-training memorization or search-induced look-ahead information leakage.

\paragraph{Quantifying Pre-Training Contamination.} 
Evaluating models on data predating their training cut-offs (Jan 2023--Jun 2024) reveals severe performance inflation due to pre-training data leakage. Across equity trend tasks, models exhibit an average 147\% relative performance gain on historical data compared to out-of-sample living sequences, underscoring that static benchmarks reward memorization over genuine predictive reasoning (detailed breakdowns in Appendix~\ref{app:extra_exps}, \autoref{tab:pure_ts_classification_living}).

\begin{wrapfigure}{R}{0.50\textwidth}
    \centering
    \includegraphics[width=\linewidth]{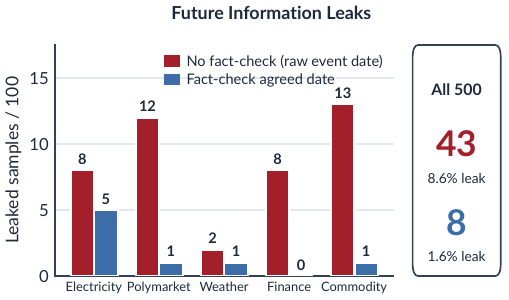}
    \caption{Human-expert-evaluated look-ahead information leakage rates across domains. Without fact-checking, 8.6\% of events exhibit leakage; applying the module suppresses this to 1.6\%.}
    \label{fig:future_info_leakage}
\end{wrapfigure}

\paragraph{Mitigating Look-Ahead Information Leakage.} 
Automated search often surfaces retrospective analyses or misdated articles that reveal post-cutoff outcomes, causing severe look-ahead information leakage. To audit benchmark hygiene, we engaged 47 vetted domain specialists (averaging 13.6 years of professional industry experience across finance, meteorology, power systems, and geopolitics) to evaluate 500 tasks across all five domains (100 per domain). 

Without our dual-agent cross-validation mechanism, automated scrapers matched verified publication dates in only 305/500 tasks (61.0\%) and introduced look-ahead information leakage in 43/500 tasks (8.6\%). Implementing our cross-validation module reconciled date accuracy to 405/500 tasks (81.0\%) and suppressed look-ahead leakage down to just 8/500 tasks (1.6\%). Crucially, as shown in \autoref{fig:future_info_leakage}, leakage was heavily concentrated in the Electricity domain, where 5 of the 8 cases deviated by only a single calendar day, a bounded error with negligible impact over our 3-day and 7-day forecasting horizons. 

\paragraph{Factuality and Semantic Grounding.} 
Expert verification confirmed that only 1.6\% of cross-validated events contained factual fabrications, with a 0\% false negative rate (indicating no valid events were erroneously excluded). In semantic extraction, automated ``Sentiment'' and ``Impact Type'' achieved 89.8\% and 89.1\% expert agreement, respectively. Conversely, ``Causality'' achieved only 17.3\% agreement, as models frequently make unsupported assumptions linking macro news to micro assets (e.g., global wafer expansions affecting DDR5 prices). These semantic attributes are excluded from our inputs, ensuring models are tested purely on their ability to filter noise and infer causal dynamics. Expert demographics, rubrics, and the failure taxonomy are detailed in Appendix~\ref{app:expert_anno}.


%% file: sections/experiment.tex
\section{Experiments and Results}\label{sec:exp_res}


In this section, we evaluate LLMs on our benchmark to investigate three research questions: 
\textbf{RQ1}: What is the overall performance of LLMs on forecasting tasks, and can they leverage events to enhance forecasting performance? (\textsection\ref{sec:event_trend_forecast_results}) 
\textbf{RQ2}: Can LLMs outperform baselines in time series forecasting? (\textsection\ref{sec:tsf_results}) 
\textbf{RQ3}: Are market trends predictable, and are events relevant to market forecasting? (\textsection\ref{sec:predictability_validations})

\begin{figure*}[t]
    \centering
    \includegraphics[width=.98\linewidth]{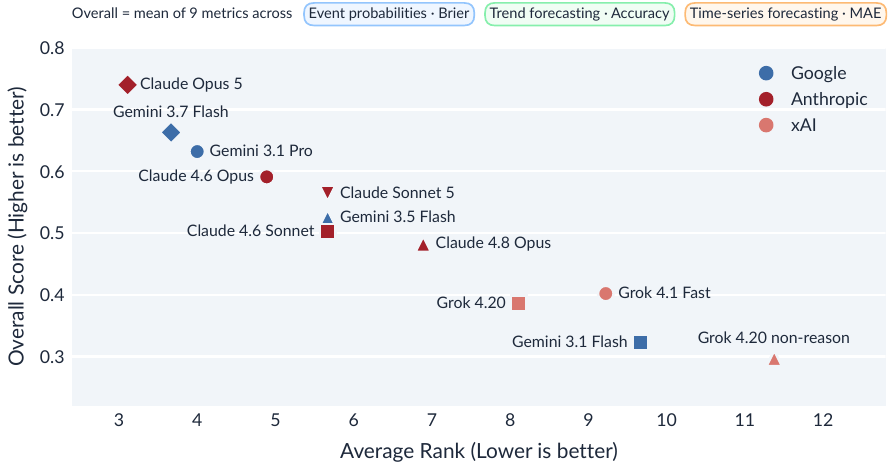} 
    \caption{Overall performance of the evaluated LLMs. The x-axis denotes the average rank across tasks, while the y-axis represents the overall score. The two metrics demonstrate a strong correlation.}
    \label{fig:overall} 
    \vspace{-5mm}
\end{figure*}

\begin{table*}[thbp]
\centering
\scriptsize
\setlength{\tabcolsep}{4.2pt}
\renewcommand{\arraystretch}{0.92}
\caption{Accuracy comparison for 3-class trend forecasting across domains (weather, electricity, and long-term mid-cap / short-term large-cap stocks), using only time series (TS) vs.\ incorporating events (w/ Evts). \colorbox{green!15}{Green} (\colorbox{red!15}{Red}) indicates improvement (degradation) over the TS-only baseline. Complete results (quinary targets, 95\% CIs, additional baselines) are in Appendix~\ref{app:extra_exps} (\autoref{tab:completed_trend_weather_elec}, \ref{tab:completed_trend_large_mid_cap_3way}, \ref{tab:completed_trend_large_mid_cap_5way}).}
\label{tab:main_trend_all_domains}
\begin{tabular*}{\linewidth}{@{\extracolsep{\fill}}l cccc cccc@{}}
\toprule
\multirow{3}{*}{\textbf{Model}} & \multicolumn{4}{c}{\textbf{Weather}} & \multicolumn{4}{c}{\textbf{Electricity}} \\
\cmidrule(lr){2-5} \cmidrule(lr){6-9}
 & \multicolumn{2}{c}{Horizon: 3 Days} & \multicolumn{2}{c}{Horizon: 7 Days} & \multicolumn{2}{c}{Horizon: 3 Days} & \multicolumn{2}{c}{Horizon: 7 Days} \\
\cmidrule(lr){2-3} \cmidrule(lr){4-5} \cmidrule(lr){6-7} \cmidrule(lr){8-9}
 & TS & w/ Evts & TS & w/ Evts & TS & w/ Evts & TS & w/ Evts \\
\midrule
Gemini 3.1 Pro   & 49.25 & \cellcolor{green!15}68.30 & 71.18 & \cellcolor{green!15}77.84 & 51.76 & \cellcolor{green!15}61.76 & 62.80 & \cellcolor{green!15}70.73 \\
Gemini 3.7 Flash & 51.67 & \cellcolor{green!15}69.81 & 70.96 & \cellcolor{green!15}78.65 & 51.76 & \cellcolor{green!15}59.41 & 71.34 & \cellcolor{red!15}70.73 \\
Claude Opus 4.6  & 47.22 & \cellcolor{green!15}61.30 & 59.23 & \cellcolor{green!15}66.92 & 44.71 & \cellcolor{green!15}60.59 & 54.88 & \cellcolor{green!15}60.37 \\
Claude Opus 5    & 45.93 & \cellcolor{green!15}69.26 & 59.42 & \cellcolor{green!15}69.23 & 50.59 & \cellcolor{green!15}64.71 & 68.90 & \cellcolor{green!15}77.44 \\
Grok 4.20        & 37.78 & \cellcolor{green!15}62.59 & 36.92 & \cellcolor{green!15}65.96 & 40.59 & \cellcolor{green!15}51.76 & 44.51 & \cellcolor{green!15}59.76 \\
\cmidrule(lr){1-9}
Gemma3-27B-it    & 31.13 & \cellcolor{green!15}52.26 & 30.59 & \cellcolor{green!15}57.84 & 36.47 & \cellcolor{green!15}47.06 & 45.73 & \cellcolor{green!15}59.15 \\
Gemma3-12B-it    & 35.09 & \cellcolor{green!15}52.45 & 42.75 & \cellcolor{green!15}63.33 & 33.53 & \cellcolor{green!15}41.76 & 43.29 & \cellcolor{green!15}60.98 \\
\midrule
\multirow{3}{*}{\textbf{Model}} & \multicolumn{4}{c}{\textbf{Mid-Cap Stocks (Long-Term)}} & \multicolumn{4}{c}{\textbf{Large-Cap Stocks (Short-Term)}} \\
\cmidrule(lr){2-5} \cmidrule(lr){6-9}
 & \multicolumn{2}{c}{Horizon: 3 Months} & \multicolumn{2}{c}{Horizon: 6 Months} & \multicolumn{2}{c}{Horizon: 1 Day} & \multicolumn{2}{c}{Horizon: 2 Days} \\
\cmidrule(lr){2-3} \cmidrule(lr){4-5} \cmidrule(lr){6-7} \cmidrule(lr){8-9}
 & TS & w/ Evts & TS & w/ Evts & TS & w/ Evts & TS & w/ Evts \\
\midrule
Gemini 3.1 Pro   & 33.76 & \cellcolor{green!15}35.90 & 40.17 & \cellcolor{green!15}49.15 & 38.72 & \cellcolor{green!15}40.85 & 42.34 & \cellcolor{green!15}44.89 \\
Gemini 3.7 Flash & 32.91 & \cellcolor{green!15}35.04 & 34.19 & \cellcolor{green!15}38.03 & 37.09 & \cellcolor{green!15}39.01 & 43.26 & \cellcolor{green!15}46.24 \\
Claude Opus 4.6  & 31.20 & \cellcolor{green!15}32.05 & 26.92 & \cellcolor{green!15}28.63 & 36.03 & \cellcolor{green!15}40.21 & 41.77 & \cellcolor{green!15}45.53 \\
Claude Opus 5    & 41.45 & \cellcolor{green!15}42.31 & 54.70 & \cellcolor{green!15}59.83 & 39.72 & \cellcolor{red!15}38.94 & 43.97 & \cellcolor{green!15}45.74 \\
Grok 4.20        & 14.10 & \cellcolor{green!15}20.51 &  6.41 & \cellcolor{green!15}26.92 & 38.79 & \cellcolor{green!15}41.13 & 39.57 & \cellcolor{green!15}42.27 \\
\cmidrule(lr){1-9}
Gemma3-27B-it    & 26.07 & \cellcolor{green!15}33.76 & 20.09 & \cellcolor{green!15}33.33 & 33.48 & \cellcolor{green!15}38.79 & 34.61 & \cellcolor{green!15}39.36 \\
Gemma3-12B-it    & 17.95 & \cellcolor{green!15}31.20 & 13.25 & \cellcolor{green!15}32.48 & 33.26 & \cellcolor{green!15}39.72 & 35.74 & \cellcolor{green!15}37.87 \\
\bottomrule
\end{tabular*}
\end{table*}

\subsection{Experimental Setup}\label{sec:setup}
\xhdr{Models and Baseline} We evaluate a total of 16 models, including both the latest proprietary models (e.g., the Claude 4.6 and 5 families, Gemini 3.x series, and Grok 4 variants) and prominent open-weight models (e.g., Qwen3.5-27B/-9B~\citep{qwen3_5_2026}, Gemma-3-27B-it/-12B-it~\citep{gemma3_2025}). We employ two time-series foundation models, TimesFM 3.0~\citep{das2024decoder} and Chronos-2~\citep{ansari2024chronos}, alongside naive time series baselines: the Last Value Baseline (LVB) and Historical Mean Baseline (HMB). Depending on the domain, LVB repeats the most recent observation or 24-hour sequence, while HMB predicts the overall or 14-day rolling average (see Appendix \ref{app:model_compute} for details).



\xhdr{Statistics and Evaluation Settings} For time-series forecasting, we use a 14-day look-back window to predict 3- and 7-day horizons. Large-cap stocks are evaluated on daily closing prices. For weather and electricity, we evaluate two distinct targets: hourly values and daily max/min values. Consequently, the test sample counts for the 3- and 7-day horizons are: commodities (340/320), stocks (1,380/1,320), electricity ($170\times 2$ / $164\times 2$), and weather ($540\times 2$ / $520\times 2$), where the $\times 2$ accounts for these dual targets. For trend forecasting, we add 3- and 6-month horizons for mid-cap stocks using a 3-month look-back (234/234 samples), and 1- and 2-day horizons for large-cap stocks (1,410/1,410 samples) to reflect rapid information pricing. Finally, we allocate an 8,192-token budget per model, discarding any baseline that yields over 33\% invalid outputs to ensure fair comparison.


\xhdr{Metrics} For market price forecasting with $n$ timestamps, we evaluate performance using Mean Absolute Percentage Error (MAPE, $\frac{1}{n}\sum_{i=1}^{n}\left|\frac{y_i-\hat{y}_i}{y_i}\right|$). For weather forecasting, we employ Mean Absolute Error (MAE) and Mean Squared Error (MSE) calculated on z-score normalized data. For event forecasting, we evaluate predictions at $6$ distinct points across the forecasting window, ranging from 100\% down to 10\% of the time-to-resolution. This window spans from the prediction start time, aligned with \cite{polymarket2026api} settings, to the final event resolution. In the evaluation, as mentioned in \autoref{sec:problem_define}, we prompt the model for the predicted probability of the actual ground-truth outcome. For evaluation metric, we utilize the average Brier Score of $m$ forecasting cutoffs ($\frac{1}{m}\sum_{t=1}^{m}(f_t-o_t)^2$), where $f_t$ is the predicted probability and $o_t$ is the outcome\footnote{\url{https://en.wikipedia.org/wiki/Brier\_score}}. 

We define an \textbf{Overall Score} as the unweighted mean of normalized $[0, 1]$ performance metrics across 9 available tasks for each model, mapping diverse raw metrics (accuracy, Brier score, and MAE) to a common ``higher is better'' scale. This aggregate metric combines scaled accuracy for trend classification, $1 - \text{Brier}$ score for event forecasting, and inverted normalized MAE for time series point forecasting. Full calculation details are provided in Appendix \ref{app:overall}.

\subsection{Event and Trend Forecasting with Context}\label{sec:event_trend_forecast_results}
Figure~\ref{fig:overall} illustrates the overall performance of the evaluated LLMs. The average rank reflects the models' relative standing across diverse tasks, whereas the overall score quantifies the specific performance gaps between them. Generally, the forecasting capability of an LLM correlates with its general-purpose abilities; for instance, Claude 5 Opus outperforms Claude 4.6 Opus, which in turn surpasses Claude 4.6 Sonnet. Since smaller-parameter models still lack the capacity to perform certain tasks, they are excluded from these aggregate results. 

\begin{figure*}[t]
    \centering
    \includegraphics[width=.98\linewidth]{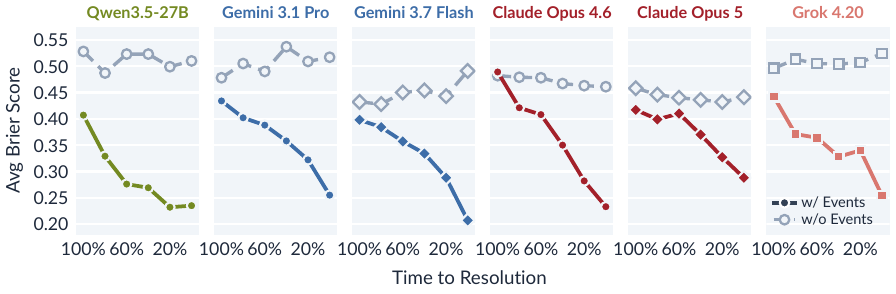} 
     \caption{Event forecasting performance across six time-to-resolution cut-offs from 100\% to 10\% based on the timeframes in \citet{polymarket2026api}. The prediction error of context-augmented models consistently decreases as the resolution date approaches. Details can be seen in \autoref{tab:completed_event_forecasting} in Appendix \ref{app:extra_exps}.}
    \label{fig:event_forecasting} 
\end{figure*}

In addition, \autoref{tab:main_trend_all_domains} shows trend forecasting accuracy for weather, electricity, and stock. Text-augmented proprietary models consistently outperform historical-data-only baselines across all settings, proving the benefit of auxiliary events. For event forecasting (\autoref{fig:event_forecasting}), textual context (solid lines) enhances performance over event-free inference (dashed lines) across all models, with prediction errors decreasing near resolution.  Detailed prompts and complete results for all 16 models, including 95\% confidence intervals, are provided in Appendices~\ref{sec:forecast_prompts} and \ref{app:extra_exps}, confirming the universal benefit of textual context in forecasting. Notably, open-weight models such as Qwen are excluded due to frequent invalid outputs in certain tasks (e.g., weather and electricity trend forecasting), whereas proprietary models yield a 0\% failure rate.

\begin{table*}
\centering
\scriptsize
\setlength{\tabcolsep}{3.8pt}
\renewcommand{\arraystretch}{0.92}
\caption{Forecasting error (MAE on $z$-score normalized data) using time series only (TS) vs.\ incorporating events (w/ Evts): hourly point forecasting and daily Max/Min forecasting of the same series. The best value per column is \underline{underlined}; shaded cells mark numerical time-series models that cannot ingest text. Complete MSE metrics and baselines are in \autoref{tab:completed_weather_elec_tsf} and \ref{tab:completed_weather_elec_maxmin} in Appendix~\ref{app:extra_exps}.}
\label{tab:main_weather_elec_mae}
\begin{tabular*}{\linewidth}{@{\extracolsep{\fill}}cl cccc cccc@{}}
\toprule
\multirow{3}{*}{\textbf{Domain}} & \multirow{3}{*}{\textbf{Model}} & \multicolumn{4}{c}{\textbf{Point Forecasting (Hourly)}} & \multicolumn{4}{c}{\textbf{Daily Max/Min Forecasting}} \\
\cmidrule(lr){3-6} \cmidrule(lr){7-10}
 & & \multicolumn{2}{c}{Horizon: 3 Days} & \multicolumn{2}{c}{Horizon: 7 Days} & \multicolumn{2}{c}{Horizon: 3 Days} & \multicolumn{2}{c}{Horizon: 7 Days} \\
\cmidrule(lr){3-4} \cmidrule(lr){5-6} \cmidrule(lr){7-8} \cmidrule(lr){9-10}
 & & TS & w/ Evts & TS & w/ Evts & TS & w/ Evts & TS & w/ Evts \\
\midrule
\multirow{7}{*}{\rotatebox{90}{\textbf{Weather}}}
 & Gemini 3.1 Pro   & 0.810 & \cellcolor{green!15}0.780 & 0.992 & \cellcolor{green!15}0.924 & 1.036 & \cellcolor{green!15}0.769 & 1.163 & \cellcolor{green!15}0.899 \\
 & Gemini 3.7 Flash & 0.917 & \cellcolor{green!15}0.865 & 1.049 & \cellcolor{green!15}1.006 & 0.989 & \underline{\cellcolor{green!15}0.675} & 1.119 & \underline{\cellcolor{green!15}0.818} \\
 & Claude Opus 4.6  & 0.847 & \cellcolor{green!15}0.735 & 0.988 & \cellcolor{green!15}0.855 & 1.006 & \cellcolor{green!15}0.764 & 1.143 & \cellcolor{green!15}0.894 \\
 & Claude Opus 5    & 0.808 & \underline{\cellcolor{green!15}0.639} & 0.973 & \underline{\cellcolor{green!15}0.774} & 1.016 & \cellcolor{green!15}0.707 & 1.149 & \cellcolor{green!15}0.819 \\
 & Grok 4.20        & 1.077 & \cellcolor{green!15}1.000 & 1.170 & \cellcolor{green!15}1.098 & 1.020 & \cellcolor{green!15}0.827 & 1.182 & \cellcolor{green!15}0.950 \\
\cmidrule(lr){2-10}
 & TimesFM 3.0      & 0.699 & \cellcolor{gray!15}-- & 0.858 & \cellcolor{gray!15}-- & 1.035 & \cellcolor{gray!15}-- & 1.228 & \cellcolor{gray!15}-- \\
 & Chronos-2        & 0.693 & \cellcolor{gray!15}-- & 0.859 & \cellcolor{gray!15}-- & 1.032 & \cellcolor{gray!15}-- & 1.216 & \cellcolor{gray!15}-- \\
\midrule
\multirow{7}{*}{\rotatebox{90}{\textbf{Electricity}}}
 & Gemini 3.1 Pro   & 0.571 & \cellcolor{green!15}0.486 & 0.695 & \cellcolor{green!15}0.590 & 1.057 & \cellcolor{green!15}0.815 & 1.178 & \cellcolor{green!15}0.955 \\
 & Gemini 3.7 Flash & 0.584 & \cellcolor{green!15}0.563 & 0.695 & \cellcolor{green!15}0.650 & 0.957 & \cellcolor{green!15}0.766 & 1.151 & \cellcolor{green!15}0.932 \\
 & Claude Opus 4.6  & 0.738 & \cellcolor{green!15}0.675 & 0.842 & \cellcolor{green!15}0.774 & 1.017 & \cellcolor{green!15}0.831 & 1.206 & \cellcolor{green!15}1.010 \\
 & Claude Opus 5    & 0.484 & \underline{\cellcolor{green!15}0.423} & 0.575 & \underline{\cellcolor{green!15}0.496} & 0.926 & \underline{\cellcolor{green!15}0.753} & 1.089 & \underline{\cellcolor{green!15}0.897} \\
 & Grok 4.20        & 0.836 & \cellcolor{green!15}0.772 & 0.860 & \cellcolor{green!15}0.842 & 1.080 & \cellcolor{green!15}1.044 & 1.226 & \cellcolor{green!15}1.177 \\
\cmidrule(lr){2-10}
 & TimesFM 3.0      & 0.491 & \cellcolor{gray!15}-- & 0.570 & \cellcolor{gray!15}-- & 1.015 & \cellcolor{gray!15}-- & 1.164 & \cellcolor{gray!15}-- \\
 & Chronos-2        & 0.474 & \cellcolor{gray!15}-- & 0.563 & \cellcolor{gray!15}-- & 1.014 & \cellcolor{gray!15}-- & 1.158 & \cellcolor{gray!15}-- \\
\bottomrule
\end{tabular*}
\end{table*}

\subsection{Time Series Forecasting with Context}\label{sec:tsf_results}

Given that time series forecasting is notoriously difficult for LLMs~\citep{merrill2024language,tan2024language}, our evaluation primarily focuses on advanced proprietary models. As \autoref{tab:main_weather_elec_mae} illustrates, while these models effectively leverage event data to enhance forecasting performance, most still perform only on par with specialized time series foundation models like TimesFM 3.0~\citep{das2024decoder}, with the exception of Claude Opus 5. Furthermore, while market price prediction is frequently included in time series forecasting tasks~\citep{chen2025mtbench,wang2024news,dong2024fnspid_acm}, our results (\autoref{tab:market_price_forecast} in Appendix~\ref{app:extra_exps}) reveal that a naive Baseline Last Value (BLV) outperforms all other methods. Given the Efficient Market Hypothesis (EMH), we conclude that this task is inherently infeasible. Detailed explanations are provided in Appendix \ref{app:infeasible}.



\subsection{Identifying Forecastable Signals and Events Validations}\label{sec:predictability_validations}
\begin{figure*}[t]
    \centering
    \includegraphics[width=.98\linewidth]{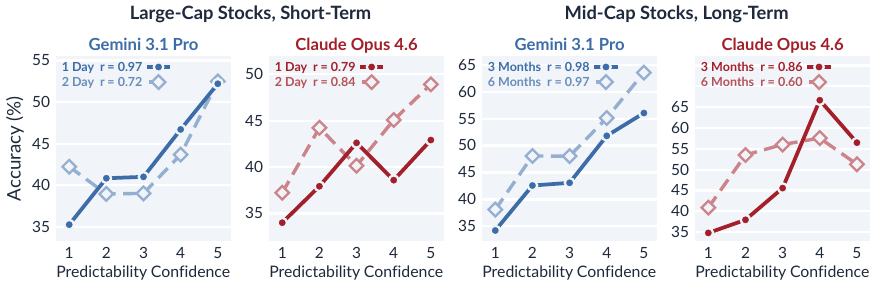} 
     \caption{Forecasting performance of large-cap and mid-cap stocks across varying levels of predictability confidence of Gemini-3.1 Pro (Gemini) and Claude Opus 4.6 (Claude). Higher forecasting accuracy is achieved on stocks assigned higher predictability confidence by the LLMs.}
    \label{fig:predictablitiy_test}
\end{figure*}

We use stock trend forecasting as a benchmark task, hypothesizing that strong, unpriced signals persist (\textsection\ref{sec:method}). We prompt LLMs to classify the daily context (events and historical prices) of 40 mid-cap or 30 large-cap stocks into five predictability confidence levels, and aggregate the daily accuracy for each level. \textbf{We observe a strong correlation between accuracy and confidence}; for example, Gemini 3.1 Pro achieves correlations of 0.97 and 0.98 in short- and long-term stock price forecasting, respectively. The main results are presented in \autoref{fig:predictablitiy_test}, with additional details and ablation studies provided in \autoref{tab:predictability_test} (Appendix~\ref{app:extra_exps}). To further validate whether our events correlate with prediction targets, we introduce ``sentiment consistent'' text~\citep{chen2025mtbench} to aid predictions (\textsection\ref{sec:method}). By controlling the proportion ($\alpha$) of this text, \autoref{fig:cases_alpha} (Appendix \ref{app:extra_exps}) shows that forecasting performance correlates strongly with $\alpha$ when target-relevant text is used. Specifically, evaluating 15 out of 30 instances in the market domain, target-relevant text yields a correlation of -0.967, whereas using text from unrelated companies yields only -0.537.

%% file: sections/conclusion.tex
\section{Conclusions and Future Works}
In summary, we introduce \leaf, a novel living benchmark for event-augmented LLM forecasting. To construct this benchmark, we propose a multi-agent retrieval and cross-validation system. Our human expert evaluation confirms that this system guarantees authentic and temporally aligned events, successfully suppressing future information leakage to just 1.6\%, with over 60\% of these remaining discrepancies bounded by a single day. Our evaluation of 16 LLMs demonstrates that incorporating these events consistently enhances forecasting performance across diverse domains. Additionally, we find that market trends exhibit predictability and correlate well with external events. Furthermore, we reveal that evaluating on outdated data causes severe performance inflation (e.g., a 147\% relative improvement in market forecasting), underscoring the critical necessity of live testing.

\xhdr{Future Work} We aim to expand the benchmark with a broader spectrum of event-driven forecasting targets and train agents or predictive models within its dynamic environment.

%% file: sections/appendix.tex
\titlecontents{section}[1.5em]
  {\vspace{0.5em}\bfseries\color{blue}}
  {\contentslabel{1.5em}}
  {\hspace*{-1.5em}}
  {\titlerule*[0.5pc]{.}\contentspage}

\titlecontents{subsection}[3.8em]
  {\normalfont\color{blue}}
  {\contentslabel{2.3em}}
  {\hspace*{-2.3em}}
  {\titlerule*[0.5pc]{.}\contentspage}

\vspace{2em}
\noindent{\Large\textbf{\textsf{Table of Contents for Appendix}}}\par
\vspace{1em}

\startcontents[appendix]
\printcontents[appendix]{}{1}{\setcounter{tocdepth}{2}}
\vspace{2em}


\clearpage

\section{Limitations}\label{app:limit}
Here, we discuss primary limitations of our paper. First, given our living benchmark requires retrieval agents to dynamically collect auxiliary events for LLM evaluation. The associated API costs constrain our domain coverage, preventing an exhaustive assessment across all forecasting fields. Second, our event forecasting setup does not incorporate historical multimodal data (e.g., time-series values). Due to data collection constraints, we rely solely on textual events as auxiliary inputs. Evaluations of other forecasting tasks in the paper incorporate both textual events and time-series values.

\section{Broader Impacts}\label{app:broader_impacts}
Our paper introduces a living benchmark to evaluate LLM forecasting capabilities across complex domains like market trends, weather, and electricity loads. By utilizing dynamically updated test data, this work addresses pre-training data contamination in LLM evaluation. Furthermore, enhancing forecasting in physical domains can positively impact societal infrastructure, such as optimizing energy grids and environmental preparedness. However, exploring financial market predictions carries the risk of unintended misuse, as users might over-trust models for algorithmic trading or personal decisions, risking financial loss. To mitigate this, we explicitly state that our research does not constitute financial advice or endorse commercial deployment, remaining strictly intended for academic evaluation.


\section{Human Expert Data Annotation}\label{app:expert_anno}

To ensure the high quality and factual authenticity of our benchmark, we conducted an exhaustive human-expert audit across 500 randomly sampled event-driven forecasting instances (100 per domain across Finance, Meteorology, Energy/Electricity, Geopolitics, and Hardware/Commodities).

\subsection{Expert Pool Demographics}
The annotation pool comprised 47 veteran specialists across seven operational domains\footnote{Annotation guidelines are provided in \url{https://anonymous.4open.science/r/LEAF-4565}}. These specialists possess extensive professional backgrounds, with domain experience ranging from 2 to 35 years (averaging 13.6 years, with a median of 10 years; 30 specialists have $\ge$10 years experience). To ensure fair compensation commensurate with domain expertise, annotators were paid an average rate of \$140 per hour. The vast majority (68\%) hold advanced postgraduate degrees (8 Doctorates, 24 Master's degrees) and serve in senior roles such as Chief Meteorologists, Quantitative Modellers, Institutional Credit Underwriters, Energy Systems Data Scientists, and Geopolitical Risk Consultants:
\begin{itemize}
    \item \textbf{Finance \& Markets (15 specialists):} CFAs, institutional credit underwriters, quantitative modellers, and long/short equity research associates (avg. 12.5 years of experience).
    \item \textbf{Weather \& Meteorology (10 specialists):} Operational forecasters, chief meteorologists, Doppler radar analysts, and numerical weather prediction (NWP) researchers (avg. 18.2 years of experience).
    \item \textbf{Energy \& Electricity (8 specialists):} Power systems engineers, grid contingency and transmission modelers, and commercial power pricing analysts (avg. 9.3 years of experience).
    \item \textbf{Geopolitics (4 specialists):} International political economists, trade sanction and export control consultants, and open-source intelligence (OSINT) researchers (avg. 11.8 years of experience).
    \item \textbf{Infrastructure, Sales \& Mobility (10 specialists):} Cloud observability engineers, transport demand modelers, and enterprise S\&OP planning directors (avg. 10.7 years of experience).
\end{itemize}

\subsection{Temporal Alignment and Leakage Mitigation}
As outlined in \autoref{sec:method}, expert evaluation reveals a major discrepancy between raw automated news retrieval and actual event occurrence dates. As illustrated in \autoref{fig:future_info_leakage}, raw retrieval dates suffer from look-ahead information leakage in 43 out of 500 cases (8.6\%), with only 305 out of 500 dates matching the true first-public release. After applying our dual-agent fact-checking module, verified publication dates match ground truth in 405 out of 500 tasks (81.0\%), while future information leakage drops to just 8 out of 500 cases (1.6\%). 

The exact profiles of these 8 residual leakage instances are documented in \autoref{tab:anno_leak}. Notably, 5 of these 8 cases originate in the Electricity domain, and 5 exhibit a temporal slippage of only a single calendar day (e.g., an outlook filed on day $T$ quoting a day $T+1$ morning schedule)—representing a bounded discrepancy with minimal impact given our 3-day and 7-day forecast horizons.

\subsection{Event Authenticity and Semantic Attributes}
Regarding factual authenticity, expert inspection confirmed that only 8 events (1.6\%) were completely fabricated or directly contradicted cited primary sources after passing through the dual fact-checking filters. As detailed in \autoref{tab:anno_fabricated}, these residual errors were predominantly concentrated in the Weather domain, where fine-grained localized forecasts are often poorly archived or superseded online. 

Finally, as shown in \autoref{tab:anno_summary}, experts audited extracted semantic labels (``\emph{causality}'', ``\emph{sentiment}'', and ``\emph{impact\_type}''). While models accurately captured ``\emph{sentiment}'' (89.5\% acceptance) and ``\emph{impact\_type}'' (89.1\% acceptance), the ``\emph{causality}'' labels achieved only 17.3\% expert acceptance (52/300 eligible tasks). Reviewers noted that models routinely make unsupported causal leaps (70.8\% of tasks), such as assuming broad industry investments immediately dictate micro-asset price movements. Consequently, \textbf{these noisy causality labels were excluded from downstream model inputs}, ensuring LLMs must independently deduce event impacts.

\subsection{Fact-Check Module Case Study}
\label{sec:fact_check_case_study}
\begin{wrapfigure}{R}{0.75\textwidth}
    \centering
    \includegraphics[width=\linewidth]{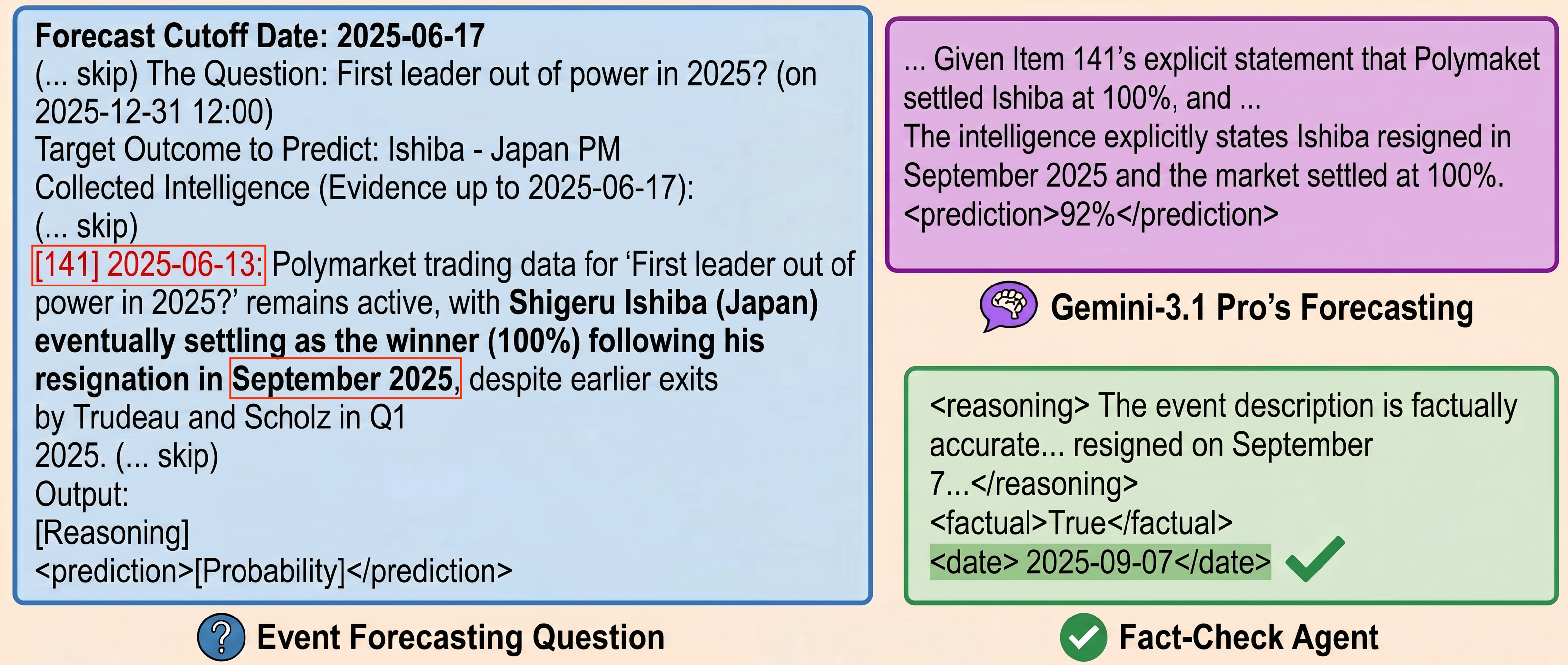}
    \caption{Example of event forecasting without Fact-Check Agents. Event \#141 leaks future information (published in 09/2025), inflating the 2025-06-13 prediction to 92\%, compared to 0\% when the event is filtered. The bottom-right panel shows the Fact-Check Agent's filtering analysis triggered by this chronological discrepancy.}
    \label{fig:fact-check}
\end{wrapfigure}
To explicitly illustrate the consequences of omitting the fact-checking module, \autoref{fig:fact-check} presents an example of severe performance inflation in Gemini-3.1-Pro caused by look-ahead information leakage. In the event forecasting task ``the first leader to depart by 2025-12-31'' (cut-off date: 2025-06-17), an unfiltered search query surfaced a leaked retrospective article (dated 2025-09-07), artificially jumping the predicted probability from 0\% to 92\%. In contrast, the fact-check agent identifies the chronological mismatch and suppresses the leaked record, preserving true out-of-sample evaluation hygiene.

\subsection{Evaluation of the Fact-Check Agent}
\label{sec:eval_fact_check_agent}
We define ``false negative cases'' for the fact-check agent as invalid instances that bypass the filter, potentially allowing the LLM to gain an unfair advantage during prediction and inflate performance. Based on human expert evaluation of 500 examples, this rate is 3.2\% in total (1.6\% future information leakage and 1.6\% fabricated events). Conversely, we define ``false positive cases'' as authentic events collected by Gemini-3.1-Pro with correct raw dates that are mistakenly discarded by the fact-check agent. Human experts manually inspected 40 randomly selected events filtered out by the mechanism and found no such instances, resulting in a false positive rate of 0\%.

\begin{table*}[t]
\centering
\small
\caption{Future-information leaks relative to the fact-check-agreed \texttt{date\_val}: 8 of 500 expert-annotated samples (1.6\%). A sample is listed when the expert's annotation states that a specific part of the event description was first published after \texttt{date\_val}. ``Gap'' is the number of days between \texttt{date\_val} and the first public release of the leaked content.}
\label{tab:anno_leak}
\resizebox{\textwidth}{!}{%
\begin{tabular}{@{}c l c p{3.3cm} p{6.2cm} c c@{}}
\toprule
\# & Task & \texttt{date\_val} & Leaked element & Leaked content in description & First public & Gap \\
\midrule
1 & Electricity\_1\_San\_Francisco & 2026-01-21 & quotes the Ascend Analytics weekly market outlook, posted the next day & ``\ldots (Source: CAISO / Ascend Analytics Market Outlook) \ldots solar output projected to strictly eclipse 10 GW beginning on Friday, January 23 \ldots'' & 2026-01-22 & 1 d \\
2 & Electricity\_2\_New\_York & 2026-01-22 & attendance figure taken from the post-event press release & ``\ldots gathered over 1,100 real estate, government, and civic leaders \ldots'' & 2026-01-23 & 1 d \\
3 & Electricity\_3\_San\_Francisco & 2026-03-17 & observed midnight temperature, first published in the 02:00 NWS discussion of the next day & ``\ldots downtown San Francisco recording 70$^\circ$F around midnight \ldots'' & 2026-03-18 & 1 d \\
4 & Electricity\_4\_New\_York & 2025-12-09 & content of an Amperon white paper whose earliest verifiable publication is 14 days later & ``\ldots Amperon published a NYISO-specific reliability report \ldots highlights shrinking reserve margins and winter fuel-security constraints \ldots'' & 2025-12-23 & 14 d \\
5 & Electricity\_5\_San\_Francisco & 2025-12-23 & forecast wording that first appears in the next day's NWS discussion & ``\ldots Saturated soils and stressed trees \ldots are forecast to cause compounding tree failures \ldots on December 25-26 \ldots'' & 2025-12-24 & 1 d \\
6 & Weather\_1\_Washington\_DC & 2025-12-04 & sounding / freezing-drizzle sentence from the advisory issued the next day & ``\ldots Forecast soundings indicate a transition to patchy freezing drizzle by the afternoon \ldots'' & 2025-12-05 & 1 d \\
7 & Event\,Forecasting\_1 & 2025-12-06 & incident first reported in IAEA Update 332, five days after the event & ``\ldots ZNPP lost all off-site power for 30 minutes, marking the 11th such blackout \ldots'' & 2025-12-11 & 5 d \\
8 & Commodity\_1 & 2025-07-15 & export licences asserted as granted; they were issued 24 days later & ``The U.S. government approved the sale of NVIDIA's H20 GPUs and AMD's MI308 AI chips to China \ldots'' & 2025-08-08 & 24 d \\
\bottomrule
\end{tabular}
}
\end{table*}

\begin{table*}[t]
\centering
\small
\caption{Expert-identified hallucinations where the core event was completely fabricated or the central claim directly contradicted the source material (8 of 500 samples, 1.6\%; 7 in the Weather domain). The dual fact-check agents failed to verify these 8 cases as factual. Both text columns provide verbatim excerpts from the raw dataset; [\ldots] denotes elided text.}
\label{tab:anno_fabricated}
\resizebox{\textwidth}{!}{%
\begin{tabular}{@{}c l c p{6.5cm} p{7.0cm}@{}}
\toprule
\# & Task & Date & Description (verbatim excerpt) & Expert reviewer (verbatim excerpt) \\
\midrule

1 & Weather\_1\_New\_York & 2025-12-23 & 
``Official National Weather Service (NWS) winter storm warnings from December 23 establish a cold thermal baseline [\ldots] recording an exact overnight low temperature forecast of -5$^\circ$C (23$^\circ$F).'' & 
``This is a completely fabricated event. There were no winter storm warnings issued on December 23 [\ldots] temperatures were actually warming [\ldots] Thus, the 23 degree F temperature in the explanation is wrong.'' \\

2 & Weather\_2\_Denver & 2025-11-10 & 
``NWS Boulder issued a Red Flag Warning for the Interstate 25 urban corridor [\ldots] valid for Tuesday, November 11.'' & 
``This is a made up event by the model. The NWS Boulder office did not issue any Red Flag warnings [\ldots] valid for Tuesday, November 11th.'' \\

3 & Weather\_3\_Denver & 2025-12-13 & 
``The December 17 cold front will lack deep moisture [\ldots] Northern Colorado will be positioned in the left-exit region of the upper-level jet [\ldots] producing localized snow squalls'' & 
``This is a completely fabricated event [\ldots] There is zero mention of a cold front on December 17th [\ldots] there is no mention of northern Colorado being in the left exit region of the jet''. \\

4 & Weather\_4\_Seattle & 2025-11-21 & 
``[\ldots] The thermal outlook calculates a 33\% to 50\% probability of above-normal temperatures across the Seattle region [\ldots] The precipitation outlook concurrently projects a 33\% to 50\% probability of above-normal precipitation'' & 
``This event was far from authentic. [\ldots] the temperature outlook [\ldots] for the Seattle area was below average temperatures and the precipitation was below avearge. This is actually opposite of what was stated in the event.'' \\

5 & Weather\_5\_New\_York & 2026-01-11 & 
``[\ldots] Published on January 11, 2026, the long-range hazards assessment highlights a `Slight Risk' contour [\ldots] for heavy snow accumulation across the Northeast region, including New York [\ldots]'' & 
``[\ldots] the CPC hazard map from the 11th does not include the slight risk of heavy snowfall to New York [\ldots] Eventually this event happened but not through the process described in this event.'' \\

6 & Weather\_6\_San\_Francisco & 2025-10-16 & 
``For the 6-10 day period [\ldots] the CPC forecasts temperatures to average near climatological normals for the West Coast [\ldots] The 8-14 day temperature outlook [\ldots] indicates a slight probability for above-average thermal anomalies across California.'' & 
``[\ldots] the temperature forecasts are incorrect. The 6-10 day outlook issed that day has temperatures mostly above normal [\ldots] The 8-14 day was near average. It's like the event and models flipped the forecasts [\ldots]'' \\

7 & Weather\_7\_Boston & 2025-12-23 & 
``[\ldots] 8-14 day extended-range forecast (ERF) indicating a macro-scale thermal anomaly characterized by strongly above-normal temperatures building across the eastern United States [\ldots] as a larger cyclonic system approaches the Northeast'' & 
``The 8-14 day outlook [\ldots] was actually near to below normal right along the East Coast. [\ldots] There also did not end up being a large cylonic system along the east coast [\ldots]'' \\

8 & Electricity\_1\_Boston & 2026-02-19 & 
``ISO New England (ISO-NE) published its Participant Readiness Project Outlook, detailing the upcoming implementation of the Inventoried Energy Program (IEP) and new Day-Ahead Ancillary Services (DASC) reserve requirements [\ldots]'' & 
``This event has been mixed up with the previous year's news. [\ldots] The Inventoried Energy Program [\ldots] was not extended into 2026, so there was no upcoming IEP implementation to report. [\ldots] the record contains no citation with ID 48.'' \\

\bottomrule
\end{tabular}%
}
\end{table*}

\begin{table}[t]
\centering
\small
\caption{Expert acceptance of generated event attributes by domain. Each value denotes the share of samples whose attribute was accepted by expert reviewers, computed over eligible samples carrying the attribute; ``--'' marks domains whose schema does not evaluate that field.}
\label{tab:anno_summary}
\setlength{\tabcolsep}{5pt}
\begin{tabular}{@{}lcccccc@{}}
\toprule
\textbf{Attribute} & \textbf{RAM / Comm.} & \textbf{Electricity} & \textbf{Geopolitical} & \textbf{Meteorology} & \textbf{Finance} & \textbf{Overall Pass} \\
\midrule
Evaluated Tasks    & 100    & 100   & 100    & 100   & 100    & 500 \\
Date Validity Pass & 83.0\% & 46.0\% & 91.0\% & 92.0\% & 93.0\% & 81.0\% (405/500) \\
Sentiment          & 84.0\% & --    & --     & --    & 95.9\% & 89.5\% (179/200) \\
Impact Type        & 87.0\% & --    & --     & --    & 92.8\% & 89.1\% (179/201) \\
Causality          & 13.0\% & --    & 23.0\% & --    & 14.4\% & 17.3\% (52/300)  \\
\bottomrule
\end{tabular}
\end{table}

\section{Methodology Supplementary}
\label{app:more_method}

\subsection{Trend Boundary}\label{sec:trend_boundary}
To define the trend boundaries for stocks and commodities, we follow an approach similar to \citep{chen2025mtbench} by categorizing based on the Rate of Change (ROC), calculated as $\text{ROC} = \frac{x_{T+N} - x_T}{x_T}$, where $x_T$ denotes the price on the last observed day $T$, and $x_{T+N}$ represents the price on the $N$-th day in the future. We adopt the boundary thresholds detailed in \autoref{tab:trend_boundary} to ensure a more balanced class distribution. Due to the long forecasting horizon and elevated volatility of mid-cap stocks, their trend boundaries are widened relative to short-term predictions. Specifically, the neutral and strong boundaries are expanded from $\pm$0.5\% to $\pm$5\% and from $\pm$5\% to $\pm$15\%, respectively.

For Weather trend classification, we avoid the rate-of-change method \citep{chen2025mtbench} since our dataset includes winter temperatures near 0 $^\circ\text{C}$, where percentage changes become ill-defined. Instead, we use absolute temperature differences based on human perception. Defining the difference between the minimum temperature over the future $N$ days ($T_{target}$) and the previous day's minimum temperature ($T_{reference}$) as $\Delta T = T_{target} - T_{reference}$ (measured in $^\circ\text{C}$), we categorize the labels as follows: Strong Up ($\Delta T > 3$), Up ($1.5 < \Delta T \le 3$), Neutral ($-1.5 \le \Delta T \le 1.5$), Down ($-3 \le \Delta T < -1.5$), and Strong Down ($\Delta T < -3$).  

For electricity load forecasting, we evaluate the absolute load change $\Delta L$ (MW) by categorizing the labels based on their observed distribution, ensuring symmetric thresholds while maintaining a relatively balanced class distribution: Strong Up ($\Delta L > 1000$, 36.70\%), Up ($400 < \Delta L \le 1000$, 17.32\%), Neutral ($-400 \le \Delta L \le 400$, 25.98\%), Down ($-1000 \le \Delta L < -400$, 13.40\%), and Strong Down ($\Delta L < -1000$, 6.60\%). Notably, $\Delta L$ is computed based on daily peak electricity loads, as accurate peak demand forecasting is essential for optimizing grid stability, managing energy resources efficiently, and preventing system overloads.

\begin{table}[htbp]
\centering
\caption{Comparison of trend label distribution and variance between our boundary and MTBench~\cite{chen2025mtbench}.}
\label{tab:trend_boundary}
\begin{tabular}{clccc}
\toprule
\multirow{2}{*}{\textbf{Benchmark}} & \multirow{2}{*}{\textbf{Trend Label}} & \multirow{2}{*}{\textbf{Boundary}} & \textbf{Large-Cap Stock} & \textbf{Commodities} \\
\cmidrule(lr){4-4} \cmidrule(lr){5-5}
 & & & \textbf{Proportion (\%)} & \textbf{Proportion (\%)} \\
\midrule
\multirow{6}{*}{\rotatebox{90}{Ours}} 
 & Strong Up   & $> 2\%$             & 23.94 & 27.69 \\
 & Up          & $0.5\% \sim 2\%$    & 16.23 & 16.93 \\
 & Neutral     & $-0.5\% \sim 0.5\%$ & 18.05 & 15.74 \\
 & Down        & $-2\% \sim -0.5\%$  & 20.59 & 13.15 \\
 & Strong Down & $< -2\%$            & 21.20 & 26.49 \\
\cmidrule(lr){2-5}
 & \textit{Variance} & -             & \textit{7.06}  & \textit{35.17} \\
\midrule
\multirow{6}{*}{\rotatebox{90}{MTBench}} 
 & Strong Up   & $> 4\%$          & 8.62  & 14.71 \\
 & Up          & $2\% \sim 4\%$   & 15.31 & 15.59 \\
 & Neutral     & $-2\% \sim 2\%$  & 54.87 & 48.53 \\
 & Down        & $-4\% \sim -2\%$ & 11.97 & 13.82 \\
 & Strong Down & $< -4\%$         & 9.23  & 7.35 \\
\cmidrule(lr){2-5}
 & \textit{Variance} & -          & \textit{309.55} & \textit{211.90} \\
\bottomrule
\end{tabular}
\end{table}

\subsection{Forecasting Domains}\label{sec:domain_details}
This section provides a comprehensive breakdown of the specific targets continuously monitored within our living benchmark, categorized across four distinct domains.

\subsubsection{Stock Market}
We track 30 \textbf{large-cap equities} representing seven critical sectors to capture diverse market dynamics, corporate announcements, and macroeconomic shifts. The selected tickers are as follows:
\begin{itemize}
    \item \textbf{Tech Giants (Magnificent Seven):} Meta (META), Apple (AAPL), NVIDIA (NVDA), Microsoft (MSFT), Alphabet (GOOGL), Amazon (AMZN), Tesla (TSLA).
    \item \textbf{Semiconductors:} AMD (AMD), Intel (INTC), TSMC (TSM), Broadcom (AVGO), Qualcomm (QCOM).
    \item \textbf{Software \& Media:} Netflix (NFLX), Disney (DIS), Adobe (ADBE), Salesforce (CRM), Oracle (ORCL).
    \item \textbf{Finance \& Payments:} JPMorgan Chase (JPM), Bank of America (BAC), Visa (V), Mastercard (MA).
    \item \textbf{Consumer \& Retail:} Walmart (WMT), Costco (COST), Coca-Cola (KO), PepsiCo (PEP), Procter \& Gamble (PG).
    \item \textbf{Healthcare:} Johnson \& Johnson (JNJ), Pfizer (PFE), Eli Lilly (LLY), UnitedHealth Group (UNH).
\end{itemize}

We extend our benchmark to include 40 mid-cap equities, characterized by higher volatility and concentration in emerging technology sectors. The tickers are categorized as follows:

\begin{itemize}
    \item \textbf{Cloud, AI \& Data Analytics:} Snowflake (SNOW), MongoDB (MDB), Datadog (DDOG), Palantir (PLTR), Super Micro Computer (SMCI), Vertiv (VRT).
    \item \textbf{Software, Security \& Interactive Media:} Cloudflare (NET), Zscaler (ZS), SentinelOne (S), AppLovin (APP), Duolingo (DUOL), Roblox (RBLX), Unity (U), Reddit (RDDT), UiPath (PATH).
    \item \textbf{Semiconductors \& Industrial Automation:} Marvell (MRVL), Astera Labs (ALAB), Coherent (COHR), Symbotic (SYM).
    \item \textbf{Fintech \& Digital Assets:} Robinhood (HOOD), SoFi (SOFI), Affirm (AFRM), Coinbase (COIN), MicroStrategy (MSTR), Toast (TOST).
    \item \textbf{Consumer, E-commerce \& Lifestyle:} Carvana (CVNA), Coupang (CPNG), Sea Limited (SE), e.l.f. Beauty (ELF), On Holding (ONON), CAVA Group (CAVA), Celsius (CELH), DraftKings (DKNG).
    \item \textbf{Frontier Tech, Biotech \& Energy:} AST SpaceMobile (ASTS), Rocket Lab (RKLB), Joby Aviation (JOBY), Rivian (RIVN), CRISPR Therapeutics (CRSP), Viking Therapeutics (VKTX), Constellation Energy (CEG).
\end{itemize}

\subsubsection{Commodities}
We dynamically monitor 10 major commodities, encompassing energy, metals, and agricultural products, to reflect vulnerabilities to global supply chains and geopolitical events:
\begin{itemize}
    \item \textbf{Energy:} Crude Oil WTI (CL=F), Brent Crude (BZ=F), Natural Gas (NG=F).
    \item \textbf{Metals:} Gold (GC=F), Silver (SI=F), Copper (HG=F).
    \item \textbf{Agriculture:} Corn (ZC=F), Soybeans (ZS=F), Wheat (ZW=F), Coffee (KC=F).
\end{itemize}

\subsubsection{Meteorological Time Series}
To evaluate predictions on complex physical systems, we collect time-series weather data from 10 major U.S. cities, selected for their diverse geographical and climatic profiles. The locations (with latitude and longitude) include:
Chicago (41.8781, -87.6298), New York (40.7128, -74.0060), San Francisco (37.7749, -122.4194), Washington D.C. (38.9072, -77.0369), Miami (25.7617, -80.1918), Los Angeles (34.0522, -118.2437), Houston (29.7604, -95.3698), Seattle (47.6062, -122.3321), Boston (42.3601, -71.0589), and Denver (39.7392, -104.9903).

\subsubsection{Electricity Time Series}
To assess forecasting capabilities on critical energy infrastructure, we retrieve electricity load data via the Grid Status API~\citep{kanter2026gridstatus}. The dataset includes load measurements from three major metropolitan areas: New York, Boston, and San Francisco. The data is collected at a 5-minute frequency, providing a high-resolution record of power demand fluctuations across diverse urban environments. While the raw data is sampled at 5-minute intervals, we aggregate it into hourly time series for the forecasting task, where each hourly value represents the mean of all observations within that hour.

\subsubsection{Event Forecasting}
Our benchmark incorporates, trending 30 diverse, real-world predictive tasks to test LLMs on reasoning over dynamic, high-uncertainty scenarios, sourcing from \cite{polymarket2026api}. Rather than relying on static question-answering, we carefully curate these events across three core categories to evaluate distinct facets of real-world decision-making:

\begin{itemize}
    \item \textbf{Tech \& AI:} Aimed at testing the models' responsiveness to rapid technological advancements and digital culture. This involves predicting industry milestones such as AI model supremacy, major tech-focused public addresses, and broader internet trends (e.g., top search volumes and best AI Models).
        
    \item \textbf{Economy \& Business:} To assess the models' grasp of macroeconomic policies and market-driving fiscal shifts, this category tracks critical financial indicators. Specific targets include consecutive Federal Reserve interest rate decisions and the regulatory rollout of major prediction platforms.
    
    \item \textbf{Politics \& World Affairs:} Designed to evaluate reasoning capabilities within evolving international landscapes. This category encompasses diverse global electoral outcomes across multiple jurisdictions, key diplomatic developments impacting regional stability, and notable administrative shifts in global governance.
    
\end{itemize}

\subsection{Prompts}\label{sec:forecast_prompts}
This section details the specific prompt templates utilized across our diverse evaluation tasks. We categorize these prompts into two primary groups based on their forecasting objectives:

\begin{itemize}
    \item \textbf{Time Series and Event Forecasting:} The prompt designs for exact point and probability predictions are presented in \autoref{fig:weather_tsf_prompt} (weather), \autoref{fig:stock_tsf_prompt} (commodities and stocks price forecasting), and \autoref{fig:event_forecast_prompt} (events forecasting).
    \item \textbf{Trend Prediction:} The templates designed for directional classification are illustrated in \autoref{fig:trend_market_3way} (3-class market trend), \autoref{fig:trend_market_5way} (5-class market trend), and \autoref{fig:weather_trend_prompt} (weather trend).
     \item \textbf{Retrieval Agent and Predictability Prompt:} The Search and Fact-Check agent prompt for stock events and Predictability Tests prompt are illustrated in \autoref{fig:stock_event_search_prompt}, \ref{fig:fact_check_example} and \ref{fig:predictability_test_prompt}.
\end{itemize}

Note that redundant or similar prompts are omitted here. The complete set of prompts is available in our repository\footnote{\url{https://anonymous.4open.science/r/LEAF-4565}}.

\section{Experiment Supplementary}
\label{app:more_results}

\subsection{Models and Computation Resources}\label{app:model_compute}

Our study evaluates a total of 16 large language models (LLMs), ranging from cutting-edge proprietary models to prominent open-weight models, alongside two time-series foundation models (TSFMs) as baselines. 

\begin{itemize}
    \item \textbf{Proprietary Models (13 in total):} We evaluate a comprehensive suite of the latest proprietary models, including the Claude family (Claude 4.6 Sonnet, Claude 5 Sonnet, Claude 4.6 Opus, Claude 4.8 Opus, Claude 5 Opus), the Gemini series (Gemini, Gemini 3.1 Flash, Gemini 3.5 Flash, Gemini 3.7 Flash), the Grok 4 variants (Grok 4.1, Grok 4.20, Grok 4.20 Non-Reasoning). These state-of-the-art models exhibit frontier performance in reasoning and zero-shot tasks.
    
    \item \textbf{Open-weight Models (4 in total):} To examine the effectiveness of publicly accessible architectures, we include four prominent models: the Qwen 3.5 series (Qwen3.5-27B and Qwen3.5-9B)~\citep{qwen3_5_2026}, alongside the Gemma 3 series (Gemma-3-27B-it and Gemma-3-12B-it)~\citep{gemma3_2025}.
    
    \item \textbf{Time-Series Foundation Models (TSFMs):} To compare the text-driven forecasting capabilities of LLMs against purely data-driven quantitative approaches, we establish robust baselines using two prominent TSFMs: Chronos-2~\cite{ansari2024chronos} and TimesFM 3.0~\cite{das2024decoder}, a decoder-only foundation model pre-trained on massive datasets. 
\end{itemize}

The retrieval agent is powered by Gemini 3.1 Pro, while the fact-checking agents utilize Gemini 3.1 Pro and Claude 4.6 Sonnet. Inference for proprietary models is conducted via their respective APIs. The primary token consumption depends on the domain and temporal coverage; on average, processing a single event consumes approximately 3,864 tokens, inclusive of the fact-checking process. The open-weight models are deployed locally on NVIDIA A100 GPUs. Further implementation details and configurations are available in our repository.

\subsection{Overall Score}\label{app:overall}
The Overall score is a single scalar in $[0, 1]$ that aggregates per-domain metrics onto a common ``higher is better'' scale. We evaluate models across nine metrics, denoted by index $k$: Stock H1D, Stock H3M, Polymarket, Electricity 3-Class, Electricity TS, Electricity Max/Min, Weather 3-Class, Weather TS, and Weather Max/Min. Given the inherent difficulty of real-world forecasting, our current evaluation focuses on these specific tasks as they exhibit relatively strong predictability.

Let $x_k(m)$ represent the raw performance value of model $m$ on metric $k$. We first map each raw metric to a normalized score, $\text{score}_k(m) \in [0, 1]$, based on its evaluation type:
\begin{equation}
    \text{score}_k(m) = 
    \begin{cases} 
        \frac{x_k(m)}{100}, & \text{if } k \text{ is an accuracy metric;} \\
        1 - x_k(m), & \text{if } k \text{ is the Polymarket Brier score;} \\
        1 - \frac{x_k(m) - \min_{m'} x_k(m')}{\max_{m'} x_k(m') - \min_{m'} x_k(m')}, & \text{if } k \text{ is an MAE metric.}
    \end{cases}
\end{equation}
Specifically, the accuracy metrics include Stock H1D, Stock H3M, Electricity 3-Class, and Weather 3-Class. The MAE metrics consist of Electricity TS, Electricity Max/Min, Weather TS, and Weather Max/Min. For MAE metrics, values are min-max normalized across all evaluated models $m'$ and subsequently inverted.

Finally, the Overall score for model $m$ is computed as the unweighted mean of its available normalized metric scores:
\begin{equation}
    \text{Overall}(m) = \frac{1}{|K(m)|} \sum_{k \in K(m)} \text{score}_k(m),
\end{equation}
where $K(m)$ denotes the subset of metrics evaluated for model $m$ (excluding missing or unavailable tasks). A higher Overall score indicates better aggregate performance.

\subsection{Case Studies}\label{sec:case_studies}
To display the reasoning process of how LLMs extract signals from complex events to aid forecasting, we present cases here. First, \autoref{fig:ef_case} presents an event forecasting scenario where the model predicts the Federal Reserve's interest rate decision by synthesizing policy projections, labor market data, and macroeconomic constraints. Second, \autoref{fig:trend_case} demonstrates a market trend prediction for Broadcom Inc., highlighting the model's ability to extract bullish signals from AI hardware partnerships and broader market sentiment. Finally, \autoref{fig:tsf_case} provides an example of meteorological time series forecasting guided by relevant weather events, such as "Pineapple Express".  To further illustrate the necessity of our dual-agent cross-validation mechanism, we present a qualitative case study demonstrating how information leakage occurs when the fact-check module is disabled. In the event forecasting task shown in \autoref{fig:fact_check_leakage}, the strict temporal cut-off is set to \textbf{2025-06-17}, while the retrieved evidence  ``[141]'' claims a valid timestamp, its actual information originates from post-cutoff events on 2025-09-07. The fact-checking process for this event, which successfully detects and filters out this temporal leakage, is detailed in \autoref{fig:fact_check_example}.

\subsection{Dataset Taxonomy}\label{sec:taxonomy}
We provide a detailed breakdown of our dataset taxonomy. \autoref{lst:dataset_taxonomy1} and \ref{lst:dataset_taxonomy2} illustrates examples of the event-augmented inputs across the four evaluated domains: Event Forecasting, Weather, Stock, and Commodities. These inputs combine textual event descriptions with historical numerical time-series data to form comprehensive multimodal prompts for the forecasting models. Due to space constraints, the complete data structures are omitted here.

\begin{table*}
\footnotesize
\centering
\caption{The market price forecasting performance of LLMs. The results demonstrate that while LLMs achieve some forecasting improvements by incorporating events (w/ Evts), the Baseline Last Value (BLV) still significantly outperforms all other methods. The best value per column is \underline{underlined}.}
\label{tab:market_price_forecast}
\resizebox{\linewidth}{!}{
\begin{tabular}{clcccccccc}
\toprule
\multirow{2}{*}{Domain} & \multirow{2}{*}{Model} & \multicolumn{4}{c}{Horizon: 3 Days} & \multicolumn{4}{c}{Horizon: 7 Days} \\
\cmidrule(lr){3-6} \cmidrule(lr){7-10}
 & & TS & w/ Evts & TS & w/ Evts & TS & w/ Evts & TS & w/ Evts \\
\midrule
\multirow{10}{*}{\rotatebox{90}{Stock}}
 & \multicolumn{1}{l}{\textit{Metrics}} & \multicolumn{2}{c}{\textit{MAPE (\%)}} & \multicolumn{2}{c}{\textit{DLV}} & \multicolumn{2}{c}{\textit{MAPE (\%)}} & \multicolumn{2}{c}{\textit{DLV}} \\
\cmidrule{2-10}
 & Gemini 3.1 Pro   & 2.748 & \cellcolor{red!15}2.873 & 1.125 & \cellcolor{green!15}1.481 & 3.865 & \cellcolor{red!15}4.028 & 1.419 & \cellcolor{green!15}2.060 \\
 & Claude 4.6 Opus  & 2.679 & \cellcolor{green!15}2.646 & 0.825 & \cellcolor{green!15}0.945 & 3.787 & \cellcolor{red!15}3.796 & 1.115 & \cellcolor{green!15}1.309 \\
 & Claude 4.8 Opus  & 2.691 & \cellcolor{green!15}2.640 & 0.911 & \cellcolor{red!15}0.897 & 3.813 & \cellcolor{green!15}3.768 & 1.171 & \cellcolor{green!15}1.214 \\
 & Claude Sonnet 5  & 2.711 & \cellcolor{green!15}2.606 & 0.956 & \cellcolor{red!15}0.910 & 3.819 & \cellcolor{green!15}3.751 & 1.295 & \cellcolor{red!15}1.217 \\
 & Grok 4.1 Fast    & 2.970 & \cellcolor{green!15}2.698 & 1.620 & \cellcolor{red!15}1.135 & 4.159 & \cellcolor{green!15}3.903 & 2.244 & \cellcolor{red!15}1.800 \\
\cmidrule{2-10}
 & TimesFM 3.0      & 2.909 & {\cellcolor{gray!15}} & 1.192 & {\cellcolor{gray!15}} & 4.157 & {\cellcolor{gray!15}} & 1.865 & {\cellcolor{gray!15}} \\
 & Chronos-2        & 2.691 & {\cellcolor{gray!15}} & 0.708 & {\cellcolor{gray!15}} & 3.934 & {\cellcolor{gray!15}} & 1.279 & {\cellcolor{gray!15}} \\
 & Base: Hist Mean  & 4.865 & {\cellcolor{gray!15}} & {\cellcolor{gray!15}} & {\cellcolor{gray!15}} & 5.619 & {\cellcolor{gray!15}} & {\cellcolor{gray!15}} & {\cellcolor{gray!15}} \\
 & Base: Last Value & \underline{2.539} & {\cellcolor{gray!15}} & {\cellcolor{gray!15}} & {\cellcolor{gray!15}} & \underline{3.619} & {\cellcolor{gray!15}} & {\cellcolor{gray!15}} & {\cellcolor{gray!15}} \\
\midrule
\multirow{10}{*}{\rotatebox{90}{Commodity}}
 & \multicolumn{1}{l}{\textit{Metrics}} & \multicolumn{2}{c}{\textit{MAPE (\%)}} & \multicolumn{2}{c}{\textit{DLV}} & \multicolumn{2}{c}{\textit{MAPE (\%)}} & \multicolumn{2}{c}{\textit{DLV}} \\
\cmidrule{2-10}
 & Gemini 3.1 Pro   & 2.711 & \cellcolor{green!15}2.679 & 0.885 & \cellcolor{green!15}1.301 & 3.411 & \cellcolor{red!15}3.515 & 1.161 & \cellcolor{green!15}1.706 \\
 & Claude 4.6 Opus  & 2.697 & \cellcolor{green!15}2.620 & 0.791 & \cellcolor{green!15}0.968 & 3.377 & \cellcolor{green!15}3.285 & 1.031 & \cellcolor{green!15}1.258 \\
 & Claude 4.8 Opus  & 2.646 & \cellcolor{green!15}2.644 & 0.872 & \cellcolor{green!15}1.142 & 3.384 & \cellcolor{red!15}3.387 & 1.214 & \cellcolor{green!15}1.536 \\
 & Claude Sonnet 5  & 2.670 & \cellcolor{red!15}2.671 & 0.845 & \cellcolor{green!15}1.086 & 3.371 & \cellcolor{green!15}3.361 & 1.159 & \cellcolor{green!15}1.500 \\
 & Grok 4.1 Fast    & 2.687 & \cellcolor{green!15}2.673 & 1.091 & \cellcolor{green!15}1.191 & 3.470 & \cellcolor{red!15}3.513 & 1.231 & \cellcolor{green!15}1.719 \\
\cmidrule{2-10}
 & TimesFM 3.0      & 2.840 & {\cellcolor{gray!15}} & 1.029 & {\cellcolor{gray!15}} & 3.741 & {\cellcolor{gray!15}} & 1.585 & {\cellcolor{gray!15}} \\
 & Chronos-2        & 2.734 & {\cellcolor{gray!15}} & 0.648 & {\cellcolor{gray!15}} & 3.630 & {\cellcolor{gray!15}} & 1.117 & {\cellcolor{gray!15}} \\
 & Base: Hist Mean  & 4.325 & {\cellcolor{gray!15}} & {\cellcolor{gray!15}} & {\cellcolor{gray!15}} & 4.763 & {\cellcolor{gray!15}} & {\cellcolor{gray!15}} & {\cellcolor{gray!15}} \\
 & Base: Last Value & \underline{2.540} & {\cellcolor{gray!15}} & {\cellcolor{gray!15}} & {\cellcolor{gray!15}} & \underline{3.236} & {\cellcolor{gray!15}} & {\cellcolor{gray!15}} & {\cellcolor{gray!15}} \\
\bottomrule
\end{tabular}
}
\end{table*}

\section{Additional Experiment Results}

\subsection{Forecast Accurate Market Price is Infeasible}\label{app:infeasible}
Grounded in the Semi-Strong Efficient Market Hypothesis (EMH)~\citep{fama1970efficient}, all publicly available information is rapidly incorporated into asset prices. Consequently, future price movements are driven almost exclusively by unforeseeable future events. Accurately predicting exact future prices would essentially require forecasting all unknown future occurrences and precisely quantifying their market impact. It is an inherently impossible endeavor. Under the Random Walk Hypothesis~\cite{fama1965behavior}, a direct mathematical corollary of the EMH, future price changes are essentially zero-mean random noise driven by unrevealed information. Consequently, attempting to predict exact future prices largely equates to forecasting this unpredictable noise. Therefore, as shown in~\autoref{tab:market_price_forecast}, in highly traded domains such as \textbf{stocks and commodities}, the current price naturally serves as the optimal, unbiased estimator. This supports the empirical finding that the Last Value Baseline (LVB) dominates exact point forecasting.

However, market efficiency is rarely absolute. Because markets can exhibit delayed responses (underreaction) to highly complex or impactful news, strong signals may not be fully absorbed within a single trading day. Consequently, market \emph{trend forecasting} remains a viable endeavor. Additionally, our current focus on highly liquid targets likely amplifies the effects of the Efficient Market Hypothesis, leading to highly efficient signal absorption and inherently greater forecasting difficulty. To better modulate the benchmark's difficulty, we will explore alternative forecasting targets in future work.

\subsection{Detailed Results}\label{app:extra_exps}

\subsubsection{Stock Experiments}
Additional stock trend forecasting results are provided in Tables \ref{tab:completed_trend_large_mid_cap_3way} and \ref{tab:completed_trend_large_mid_cap_5way}. As observed in \autoref{tab:completed_trend_large_mid_cap_5way}, proprietary models frequently underperform the random guess baseline in the 5-way long-term mid-cap stock trend forecasting task. This degradation primarily stems from a severe mismatch between the models' conservative prediction behaviors and a heavy-tailed ground-truth distribution. 

As detailed in \autoref{tab:pred_distribution}, frontier models systematically avoid predicting the extreme classes (``Strong Up'' and ``Strong Down''). Conversely, these two categories constitute the vast majority of the actual ground truth, leading to significant deviations from reality. Under our current formulation, the price change boundaries are defined as $<5\%$ for ``Neutral'' and $>15\%$ for ``Strong''. While raising the ``Strong'' threshold could balance the label distribution and artificially bring results closer to the baseline, such tuning would simply overfit to the volatility of this specific evaluation period. 

Instead of retroactively manipulating thresholds, we choose to maintain our fixed definitions. We highlight this conservative prediction bias as an inherent \textbf{limitation of both current LLMs and static-threshold experimental setups} in trend forecasting, and present this analysis to provide deeper insights into model behavior. Ultimately, a highly capable forecasting LLM should remain robust against such degrees of class imbalance.

\begin{figure*}[t]
    \centering
    \begin{minipage}{0.48\linewidth}
        \centering
        \includegraphics[width=\linewidth]{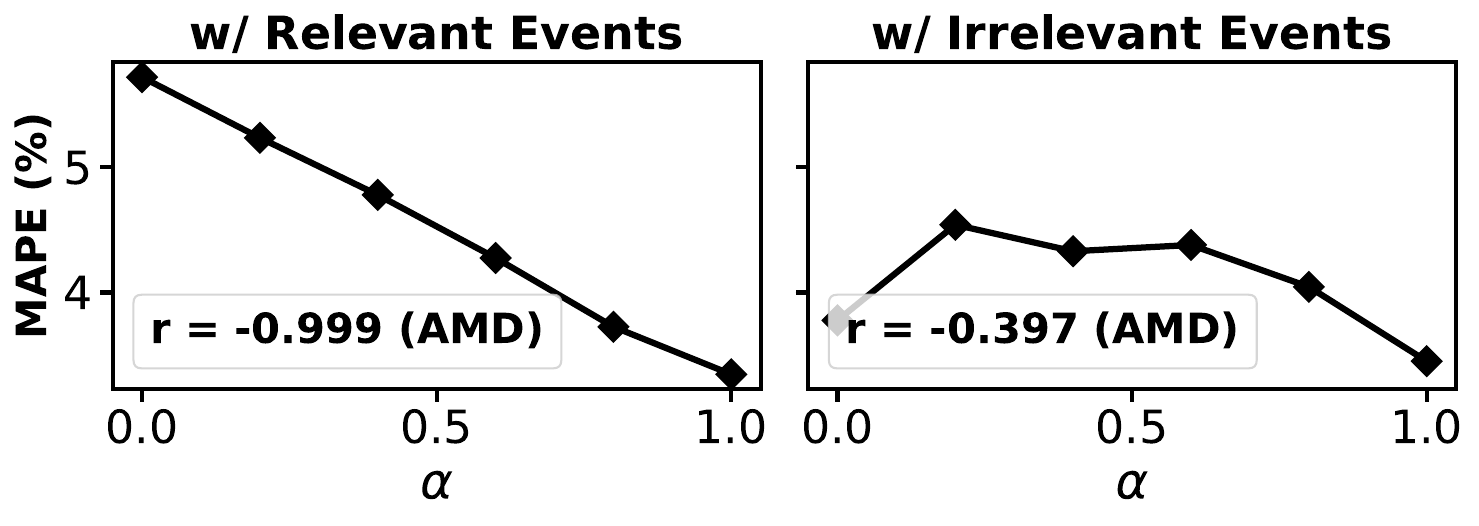}
    \end{minipage}
    \hfill
    \begin{minipage}{0.48\linewidth}
        \centering
        \includegraphics[width=\linewidth]{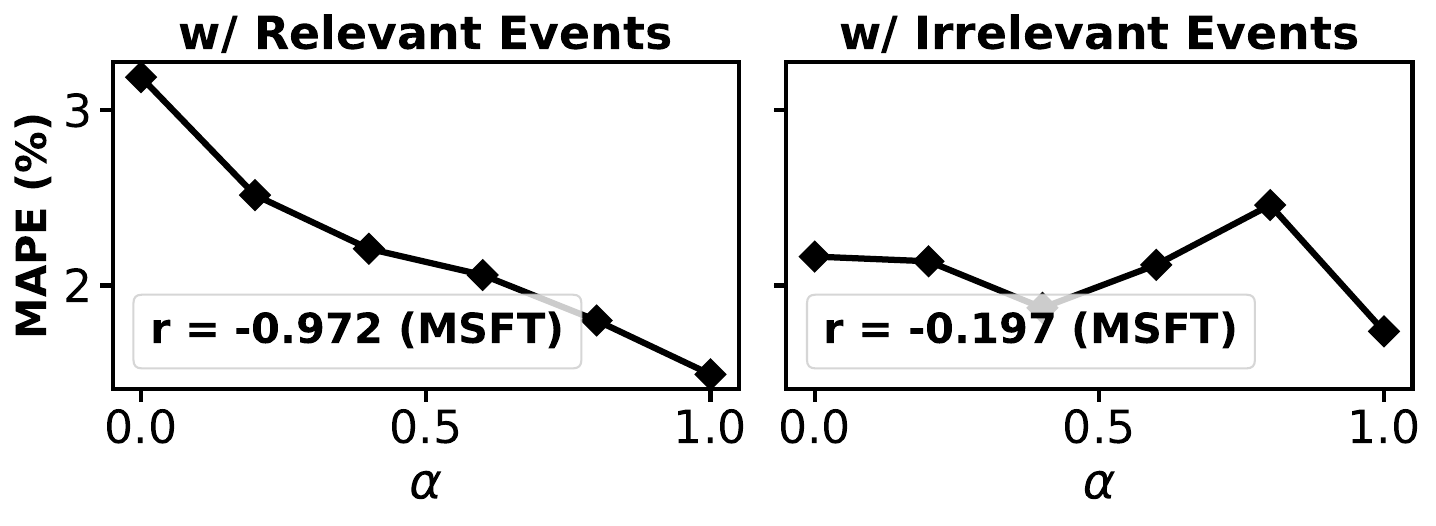}
    \end{minipage}
    \caption{Examples comparing forecasting performance with (w/) relevant or irrelevant events as the proportion ($\alpha$, x-axis) of ``consistent'' text increases. The correlation coefficient is denoted by $r$.}
    \label{fig:cases_alpha}
\end{figure*}

\autoref{fig:cases_alpha} presents two examples of data validation, as mentioned in \autoref{sec:predictability_validations}. The results indicate that when target-relevant events are utilized to assist forecasting, the variations in prediction outcomes strongly correlate with the trend of $\alpha$. Conversely, no significant correlation is observed when irrelevant events are used. Furthermore, \autoref{tab:pure_ts_classification_living} illustrates the impact of outdated data on evaluation performance, as mentioned in Section \ref{sec:integrity}. Specifically, utilizing historical data significantly inflates the models' apparent capabilities, highlighting a evaluation data contamination problem.

\autoref{tab:predictability_test} details the relationship between predictability levels and forecasting performance, as discussed in \textsection\ref{sec:predictability_validations}. As observed, models achieve higher forecasting accuracy on samples assigned high predictability confidence by the models themselves. However, this correlation vanishes when events are excluded, indicating that textual events are a prerequisite for models to make reliable, high-confidence predictions.

\providecommand{\ci}[2]{#1\,{\scriptsize$\pm$#2}}

\subsubsection{Other Results}
This section provides supplementary evaluation results across different tasks. \autoref{tab:completed_event_forecasting} details the event forecasting performance at varying time-to-resolution cut-offs. For the weather and electricity domains, \autoref{tab:completed_trend_weather_elec} reports the comprehensive trend classification results, while \autoref{tab:completed_weather_elec_tsf} and \autoref{tab:completed_weather_elec_maxmin} present the fine-grained numerical errors for hourly point forecasting and daily extremum (Max/Min) predictions, respectively.

\begin{table*}[htbp]
\centering
\caption{Accuracy (\%) and correlation analysis across different Predictability Levels for Large-Cap and Mid-Cap stocks and both Gemini 3.1 Pro and Claude 4.6 Opus.}
\label{tab:predictability_test}
\begin{tabular}{c l c c c c c c c c}
\toprule
 & & \multicolumn{4}{c}{\textbf{Large-Cap Stocks Short-Term}} & \multicolumn{4}{c}{\textbf{Mid-Cap Stocks Long-Term}} \\
\cmidrule(lr){3-6} \cmidrule(lr){7-10}
 & & \multicolumn{2}{c}{Horizon: 1} & \multicolumn{2}{c}{Horizon: 2} & \multicolumn{2}{c}{Horizon: 3 Months} & \multicolumn{2}{c}{Horizon: 6 Months} \\
\cmidrule(lr){3-4} \cmidrule(lr){5-6} \cmidrule(lr){7-8} \cmidrule(lr){9-10}
\multicolumn{2}{l}{\textbf{Metrics}} & TS & w/ Evts & TS & w/ Evts & TS & w/ Evts & TS & w/ Evts \\
\midrule
\multirow{6}{*}{\rotatebox{90}{\textbf{Gemini 3.1 Pro}}} 
 & Level 1 & 39.32 & 35.27 & 41.85 & 42.24 & 35.96 & 34.21 & 48.09 & 38.10 \\
 & Level 2 & 34.93 & 40.85 & 42.16 & 38.96 & 42.38 & 42.58 & 44.79 & 48.07 \\
 & Level 3 & 37.37 & 41.02 & 39.19 & 39.03 & 40.84 & 43.08 & 41.67 & 48.04 \\
 & Level 4 & 40.85 & 46.72 & 40.96 & 43.69 & 46.05 & 51.82 & 49.43 & 55.14 \\
 & Level 5 & 34.80 & 52.22 & 42.58 & 52.50 & 46.97 & 56.06 & 54.50 & 63.58 \\
\cmidrule{2-10}
 & Correlation & -0.184 & \cellcolor{green!15}0.974 & 0.031 & \cellcolor{green!15}0.720 & 0.919 & \cellcolor{green!15}0.978 & 0.569 & \cellcolor{green!15}0.970 \\
\midrule
\multirow{6}{*}{\rotatebox{90}{\textbf{Claude 4.6 Opus}}} 
 & Level 1 & 38.13 & 34.02 & 36.51 & 37.24 & 26.87 & 34.81 & 29.82 & 40.88 \\
 & Level 2 & 38.03 & 37.94 & 39.03 & 44.24 & 30.97 & 37.96 & 48.54 & 53.53 \\
 & Level 3 & 37.66 & 42.62 & 37.50 & 40.15 & 38.37 & 45.58 & 45.83 & 56.03 \\
 & Level 4 & 40.85 & 38.60 & 40.91 & 45.08 & 50.82 & 66.67 & 45.30 & 57.58 \\
 & Level 5 & 29.22 & 42.92 & 32.17 & 48.92 & 75.83 & 56.48 & 58.10 & 51.28 \\
\cmidrule{2-10}
 & Correlation & -0.538 & \cellcolor{green!15}0.792 & -0.328 & \cellcolor{green!15}0.845 & 0.945 & \cellcolor{red!15}0.861 & 0.829 & \cellcolor{red!15}0.596 \\
\bottomrule
\end{tabular}
\end{table*}

\begin{table*}[htbp]
\centering
\small
\renewcommand{\arraystretch}{1.4} 
\caption{\textbf{Temporal Ablation on ``Living'' Data:} Accuracy (\%) comparison for 3-class and 5-class stock predictions using pure time-series inputs. The ``Before'' columns denote a historical test period (Jan 2, 2023 -- Jun 2, 2024) preceding the general training cut-offs, where models exhibit artificially inflated accuracy due to data contamination. The ``After'' columns represent strictly unseen, out-of-sample ``living'' data (Oct 1, 2025 -- Apr 5, 2026). Green shading highlights instances where historical performance is strictly greater than the true out-of-sample performance.}
\label{tab:pure_ts_classification_living}
\begin{tabular}{c l c c c c c c c c}
\toprule
 & & \multicolumn{4}{c}{\textbf{Large-Cap Stocks Short-Term}} & \multicolumn{4}{c}{\textbf{Mid-Cap Stocks Long-Term}} \\
\cmidrule(lr){3-6} \cmidrule(lr){7-10}
 & & \multicolumn{2}{c}{Horizon: 1D} & \multicolumn{2}{c}{Horizon: 2D} & \multicolumn{2}{c}{Horizon: 3M} & \multicolumn{2}{c}{Horizon: 6M} \\
\cmidrule(lr){3-4} \cmidrule(lr){5-6} \cmidrule(lr){7-8} \cmidrule(lr){9-10}
\multicolumn{2}{l}{\textbf{Model (TS Only)}} & After & Before & After & Before & After & Before & After & Before \\
\midrule
\multirow{2}{*}{\rotatebox{90}{\textbf{3-class}}} 
 & Gemini 3.1 Pro  & 37.33 & \cellcolor{green!15}66.90 & 41.27 & \cellcolor{green!15}73.10 & 40.76 & \cellcolor{green!15}86.93 & 46.02 & \cellcolor{green!15}92.36 \\
 & Claude 4.6 Opus & 37.81 & \cellcolor{green!15}44.73 & 37.97 & \cellcolor{green!15}47.29 & 34.86 & \cellcolor{green!15}74.74 & 42.23 & \cellcolor{green!15}82.97 \\
\cmidrule{2-10}
 & \multicolumn{1}{l}{Baseline (Random Guess)} & \multicolumn{4}{c}{33.33} & \multicolumn{4}{c}{33.33} \\
\midrule
\multirow{2}{*}{\rotatebox{90}{\textbf{5-Class}}} 
 & Gemini 3.1 Pro  & 30.51 & \cellcolor{green!15}65.32 & 23.63 & \cellcolor{green!15}62.66 & 10.90 & \cellcolor{green!15}26.14 & 13.24 & \cellcolor{green!15}22.91 \\
 & Claude 4.6 Opus & 29.49 & \cellcolor{green!15}35.76 & 21.34 & \cellcolor{green!15}28.87 & 9.86  & \cellcolor{green!15}49.93 & 7.87  & \cellcolor{green!15}68.43 \\
\cmidrule{2-10}
 & \multicolumn{1}{l}{Baseline (Random Guess)} & \multicolumn{4}{c}{20.00} & \multicolumn{4}{c}{20.00} \\
\bottomrule
\end{tabular}
\end{table*}

\begin{table}[htbp]
\centering
\caption{Prediction distributions across the 5-way classification categories for long-term mid-cap stock forecasting. Proprietary models disproportionately favor moderate and neutral classes, whereas the ground truth is dominated by extreme price movements.}
\label{tab:pred_distribution}
\renewcommand{\arraystretch}{0.8}
\resizebox{\textwidth}{!}{
\begin{tabular}{l ccccc}
\toprule
\textbf{Model} & \textbf{Strong Up (\%)} & \textbf{Up (\%)} & \textbf{Neutral (\%)} & \textbf{Down (\%)} & \textbf{Strong Down (\%)} \\
\midrule
Gemini 3.1 Pro (w/ Evts)   & 10.87 & 25.28 & 22.42 & 36.11 & 5.32 \\
Gemini 3.1 Pro (TS)        & 3.57  & 23.61 & 40.44 & 24.40 & 7.98 \\
Gemini 3.1 Flash (w/ Evts) & 0.08  & 40.36 & 37.98 & 21.55 & 0.04 \\
Gemini 3.1 Flash (TS)      & 0.00  & 1.35  & 84.48 & 13.93 & 0.24 \\
Claude 4.6 Opus (w/ Evts)  & 0.40  & 68.89 & 22.34 & 8.29  & 0.08 \\
Claude 4.6 Opus (TS)       & 0.04  & 55.63 & 31.75 & 12.58 & 0.00 \\
Claude 4.6 Sonnet (w/ Evts)& 0.67  & 49.72 & 21.79 & 27.70 & 0.12 \\
Claude 4.6 Sonnet (TS)     & 0.12  & 37.06 & 32.50 & 30.28 & 0.04 \\
\midrule
\textbf{Ground Truth}      & \textbf{43.69} & \textbf{9.05}  & \textbf{9.17}  & \textbf{8.69}  & \textbf{29.40} \\
\bottomrule
\end{tabular}
}
\end{table}

\begin{table*}[htbp]
\footnotesize
\centering
\renewcommand{\arraystretch}{1.8} 
\caption{Accuracy comparison for ternary (3-class) classification between time-series-only (TS) and event-inclusive (w/ Evts) forecasting for long-term mid-cap and short-term large-cap stocks, for the completed 16 models. Each cell is the mean over the pooled samples $\pm$ the half-width of its 95\,\% confidence interval.}
\label{tab:completed_trend_large_mid_cap_3way}
\resizebox{\linewidth}{!}{
\begin{tabular}{lcccccccc}
\toprule
\multirow{3}{*}{Model} & \multicolumn{4}{c}{\textbf{Mid-Cap Stocks Long-Term}} & \multicolumn{4}{c}{\textbf{Large-Cap Stocks Short-Term}} \\
\cmidrule(lr){2-5} \cmidrule(lr){6-9}
 & \multicolumn{2}{c}{Horizon: 3 Months} & \multicolumn{2}{c}{Horizon: 6 Months} & \multicolumn{2}{c}{Horizon: 1 Day} & \multicolumn{2}{c}{Horizon: 2 Days} \\
\cmidrule(lr){2-3} \cmidrule(lr){4-5} \cmidrule(lr){6-7} \cmidrule(lr){8-9}
 & TS & w/ Evts & TS & w/ Evts & TS & w/ Evts & TS & w/ Evts \\
\midrule
Gemini 3.1 Pro       & \ci{33.8}{6.1} & \cellcolor{green!15}\ci{35.9}{6.2} & \ci{40.2}{6.3} & \cellcolor{green!15}\ci{49.1}{6.4} & \ci{38.7}{2.5} & \cellcolor{green!15}\ci{40.9}{2.6} & \ci{42.3}{2.6} & \cellcolor{green!15}\ci{44.9}{2.6} \\
Gemini 3.1 Flash     & \ci{23.5}{5.4} & \cellcolor{green!15}\ci{29.5}{5.9} & \ci{19.7}{5.1} & \cellcolor{green!15}\ci{33.3}{6.1} & \ci{33.8}{2.5} & \cellcolor{green!15}\ci{37.9}{2.5} & \ci{32.7}{2.4} & \cellcolor{green!15}\ci{36.7}{2.5} \\
Gemini 3.5 Flash     & \ci{28.0}{5.8} & \cellcolor{green!15}\ci{42.1}{6.4} & \ci{27.3}{5.8} & \cellcolor{green!15}\ci{41.2}{6.3} & \ci{38.2}{2.5} & \cellcolor{green!15}\ci{39.8}{2.6} & \ci{43.5}{2.6} & \cellcolor{red!15}\ci{43.4}{2.6} \\
Gemini 3.7 Flash     & \ci{32.9}{6.0} & \cellcolor{green!15}\ci{35.0}{6.1} & \ci{34.2}{6.1} & \cellcolor{green!15}\ci{38.0}{6.2} & \ci{37.1}{2.5} & \cellcolor{green!15}\ci{39.0}{2.5} & \ci{43.3}{2.6} & \cellcolor{green!15}\ci{46.2}{2.6} \\
\cmidrule{1-9}
Claude 4.6 Opus      & \ci{31.2}{5.9} & \cellcolor{green!15}\ci{32.1}{6.0} & \ci{26.9}{5.7} & \cellcolor{green!15}\ci{28.6}{5.8} & \ci{36.0}{2.5} & \cellcolor{green!15}\ci{40.2}{2.6} & \ci{41.8}{2.6} & \cellcolor{green!15}\ci{45.5}{2.6} \\
Claude 4.6 Sonnet    & \ci{37.2}{6.2} & \ci{37.2}{6.2} & \ci{34.6}{6.1} & \cellcolor{green!15}\ci{37.2}{6.2} & \ci{36.7}{2.5} & \cellcolor{green!15}\ci{44.1}{2.6} & \ci{40.1}{2.6} & \cellcolor{green!15}\ci{44.3}{2.6} \\
Claude 4.8 Opus      & \ci{27.8}{5.8} & \cellcolor{green!15}\ci{30.3}{5.9} & \ci{32.9}{6.0} & \cellcolor{green!15}\ci{34.6}{6.1} & \ci{37.6}{2.5} & \cellcolor{green!15}\ci{40.9}{2.6} & \ci{40.3}{2.6} & \cellcolor{green!15}\ci{42.8}{2.6} \\
Claude Opus 5        & \ci{41.5}{6.3} & \cellcolor{green!15}\ci{42.3}{6.3} & \ci{54.7}{6.4} & \cellcolor{green!15}\ci{59.8}{6.3} & \ci{39.7}{2.6} & \cellcolor{red!15}\ci{38.9}{2.5} & \ci{44.0}{2.6} & \cellcolor{green!15}\ci{45.7}{2.6} \\
Claude Sonnet 5      & \ci{29.1}{5.8} & \cellcolor{green!15}\ci{32.5}{6.0} & \ci{31.2}{5.9} & \cellcolor{green!15}\ci{41.0}{6.3} & \ci{34.5}{2.5} & \cellcolor{green!15}\ci{38.6}{2.5} & \ci{33.0}{2.5} & \cellcolor{green!15}\ci{39.7}{2.6} \\
\cmidrule{1-9}
Grok 4.1 Fast        & \ci{19.2}{5.1} & \cellcolor{green!15}\ci{24.8}{5.5} & \ci{11.5}{4.1} & \cellcolor{green!15}\ci{27.4}{5.7} & \ci{37.3}{2.5} & \cellcolor{red!15}\ci{36.7}{2.5} & \ci{36.2}{2.5} & \cellcolor{green!15}\ci{39.9}{2.6} \\
Grok 4.20            & \ci{14.1}{4.5} & \cellcolor{green!15}\ci{20.5}{5.2} & \ci{6.4}{3.1} & \cellcolor{green!15}\ci{26.9}{5.7} & \ci{38.8}{2.5} & \cellcolor{green!15}\ci{41.1}{2.6} & \ci{39.6}{2.6} & \cellcolor{green!15}\ci{42.3}{2.6} \\
Grok 4.20 non-reason & \ci{17.1}{4.8} & \cellcolor{red!15}\ci{14.1}{4.5} & \ci{13.2}{4.4} & \cellcolor{green!15}\ci{15.0}{4.6} & \ci{34.0}{2.5} & \cellcolor{green!15}\ci{38.4}{2.5} & \ci{32.6}{2.4} & \cellcolor{green!15}\ci{35.6}{2.5} \\
\cmidrule{1-9}
Gemma3-27B-it        & \ci{26.1}{5.6} & \cellcolor{green!15}\ci{33.8}{6.1} & \ci{20.1}{5.1} & \cellcolor{green!15}\ci{33.3}{6.1} & \ci{33.5}{2.5} & \cellcolor{green!15}\ci{38.8}{2.5} & \ci{34.6}{2.5} & \cellcolor{green!15}\ci{39.4}{2.6} \\
Gemma3-12B-it        & \ci{17.9}{4.9} & \cellcolor{green!15}\ci{31.2}{5.9} & \ci{13.2}{4.4} & \cellcolor{green!15}\ci{32.5}{6.0} & \ci{33.3}{2.5} & \cellcolor{green!15}\ci{39.7}{2.6} & \ci{35.7}{2.5} & \cellcolor{green!15}\ci{37.9}{2.5} \\
Qwen3.5-27B          & \ci{17.9}{4.9} & \cellcolor{green!15}\ci{26.5}{5.7} & \ci{11.1}{4.0} & \cellcolor{green!15}\ci{33.8}{6.1} & \ci{36.7}{2.5} & \cellcolor{green!15}\ci{41.6}{2.6} & \ci{38.8}{2.5} & \cellcolor{green!15}\ci{43.5}{2.6} \\
Qwen3.5-9B           & \ci{15.9}{4.7} & \cellcolor{green!15}\ci{17.5}{4.9} & \ci{9.5}{3.8} & \cellcolor{green!15}\ci{19.5}{5.1} & \ci{30.8}{2.4} & \cellcolor{green!15}\ci{39.3}{3.0} & \ci{33.4}{2.5} & \cellcolor{green!15}\ci{42.0}{3.0} \\
\bottomrule
\end{tabular}
}
\end{table*}

\begin{table*}[htbp]
\footnotesize
\centering
\renewcommand{\arraystretch}{1.8} 
\caption{Accuracy comparison for quinary (5-class) classification between time-series-only (TS) and event-inclusive (w/ Evts) forecasting for long-term mid-cap and short-term large-cap stocks, for the proprietary models. Each cell is the mean over the pooled samples $\pm$ the half-width of its 95\,\% confidence interval.}
\label{tab:completed_trend_large_mid_cap_5way}
\resizebox{\linewidth}{!}{
\begin{tabular}{lcccccccc}
\toprule
\multirow{3}{*}{Model} & \multicolumn{4}{c}{\textbf{Mid-Cap Stocks Long-Term}} & \multicolumn{4}{c}{\textbf{Large-Cap Stocks Short-Term}} \\
\cmidrule(lr){2-5} \cmidrule(lr){6-9}
 & \multicolumn{2}{c}{Horizon: 3 Months} & \multicolumn{2}{c}{Horizon: 6 Months} & \multicolumn{2}{c}{Horizon: 1 Day} & \multicolumn{2}{c}{Horizon: 2 Days} \\
\cmidrule(lr){2-3} \cmidrule(lr){4-5} \cmidrule(lr){6-7} \cmidrule(lr){8-9}
 & TS & w/ Evts & TS & w/ Evts & TS & w/ Evts & TS & w/ Evts \\
\midrule
Gemini 3.1 Pro       & \ci{13.7}{4.4} & \cellcolor{green!15}\ci{14.5}{4.5} & \ci{12.4}{4.2} & \cellcolor{green!15}\ci{16.7}{4.8} & \ci{23.9}{2.2} & \cellcolor{green!15}\ci{27.6}{2.3} & \ci{22.3}{2.2} & \cellcolor{green!15}\ci{29.0}{2.4} \\
Gemini 3.1 Flash     & \ci{10.3}{3.9} & \cellcolor{green!15}\ci{10.7}{4.0} & \ci{7.3}{3.3} & \cellcolor{green!15}\ci{9.8}{3.8} & \ci{23.5}{2.2} & \cellcolor{green!15}\ci{26.5}{2.3} & \ci{16.5}{1.9} & \cellcolor{green!15}\ci{20.2}{2.1} \\
Gemini 3.5 Flash     & \ci{14.1}{4.5} & \cellcolor{green!15}\ci{15.0}{4.6} & \ci{10.7}{4.0} & \cellcolor{red!15}\ci{9.4}{3.7} & \ci{22.6}{2.2} & \cellcolor{green!15}\ci{26.5}{2.3} & \ci{26.0}{2.3} & \ci{26.0}{2.3} \\
Gemini 3.7 Flash     & \ci{14.1}{4.5} & \cellcolor{red!15}\ci{12.8}{4.3} & \ci{15.4}{4.6} & \cellcolor{red!15}\ci{13.7}{4.4} & \ci{23.0}{2.2} & \cellcolor{green!15}\ci{24.3}{2.2} & \ci{21.0}{2.1} & \cellcolor{green!15}\ci{25.8}{2.3} \\
\cmidrule{1-9}
Claude 4.6 Opus      & \ci{11.5}{4.1} & \cellcolor{green!15}\ci{12.0}{4.2} & \ci{6.4}{3.1} & \cellcolor{red!15}\ci{5.6}{2.9} & \ci{22.8}{2.2} & \cellcolor{green!15}\ci{26.5}{2.3} & \ci{18.9}{2.0} & \cellcolor{green!15}\ci{21.5}{2.1} \\
Claude 4.6 Sonnet    & \ci{11.5}{4.1} & \cellcolor{red!15}\ci{10.3}{3.9} & \ci{8.5}{3.6} & \cellcolor{red!15}\ci{6.8}{3.2} & \ci{23.9}{2.2} & \cellcolor{green!15}\ci{24.4}{2.2} & \ci{21.0}{2.1} & \cellcolor{green!15}\ci{25.7}{2.3} \\
Claude 4.8 Opus      & \ci{10.3}{3.9} & \cellcolor{green!15}\ci{12.8}{4.3} & \ci{9.0}{3.7} & \cellcolor{red!15}\ci{6.0}{3.0} & \ci{24.3}{2.2} & \ci{24.3}{2.2} & \ci{20.2}{2.1} & \cellcolor{green!15}\ci{21.3}{2.1} \\
Claude Opus 5        & \ci{14.1}{4.5} & \cellcolor{red!15}\ci{13.7}{4.4} & \ci{31.2}{5.9} & \cellcolor{green!15}\ci{37.6}{6.2} & \ci{25.6}{2.3} & \cellcolor{green!15}\ci{27.3}{2.3} & \ci{23.0}{2.2} & \cellcolor{green!15}\ci{24.1}{2.2} \\
Claude Sonnet 5      & \ci{9.0}{3.7} & \cellcolor{red!15}\ci{6.8}{3.2} & \ci{9.4}{3.7} & \cellcolor{red!15}\ci{9.0}{3.7} & \ci{24.0}{2.2} & \cellcolor{green!15}\ci{25.0}{2.3} & \ci{18.5}{2.0} & \cellcolor{green!15}\ci{20.6}{2.1} \\
\cmidrule{1-9}
Grok 4.1 Fast        & \ci{12.8}{4.3} & \cellcolor{red!15}\ci{6.4}{3.1} & \ci{6.0}{3.0} & \cellcolor{green!15}\ci{8.5}{3.6} & \ci{23.3}{2.2} & \cellcolor{green!15}\ci{23.6}{2.2} & \ci{18.9}{2.0} & \cellcolor{green!15}\ci{22.0}{2.2} \\
Grok 4.20            & \ci{14.1}{4.5} & \cellcolor{red!15}\ci{9.8}{3.8} & \ci{6.4}{3.1} & \cellcolor{green!15}\ci{7.7}{3.4} & \ci{22.9}{2.2} & \cellcolor{green!15}\ci{23.5}{2.2} & \ci{18.7}{2.0} & \cellcolor{green!15}\ci{22.1}{2.2} \\
Grok 4.20 non-reason & \ci{12.8}{4.3} & \cellcolor{red!15}\ci{12.0}{4.2} & \ci{9.4}{3.7} & \cellcolor{red!15}\ci{6.0}{3.0} & \ci{24.8}{2.3} & \cellcolor{green!15}\ci{24.9}{2.3} & \ci{17.0}{2.0} & \cellcolor{green!15}\ci{18.9}{2.0} \\
\cmidrule{1-9}
Gemma3-27B-it        & \ci{12.8}{4.3} & \cellcolor{red!15}\ci{9.4}{3.7} & \ci{6.8}{3.2} & \cellcolor{green!15}\ci{10.7}{4.0} & \ci{24.8}{2.3} & \cellcolor{green!15}\ci{25.7}{2.3} & \ci{20.6}{2.1} & \cellcolor{green!15}\ci{22.6}{2.2} \\
Gemma3-12B-it        & \ci{12.4}{4.2} & \cellcolor{red!15}\ci{7.7}{3.4} & \ci{6.0}{3.0} & \cellcolor{green!15}\ci{6.8}{3.2} & \ci{24.5}{2.2} & \cellcolor{green!15}\ci{25.7}{2.3} & \ci{20.9}{2.1} & \cellcolor{green!15}\ci{21.2}{2.1} \\
Qwen3.5-27B          & \ci{12.4}{4.2} & \cellcolor{red!15}\ci{9.0}{3.7} & \ci{7.7}{3.4} & \cellcolor{green!15}\ci{8.5}{3.6} & \ci{23.3}{2.2} & \cellcolor{green!15}\ci{26.4}{2.4} & \ci{18.6}{2.0} & \cellcolor{green!15}\ci{24.0}{2.3} \\
Qwen3.5-9B           & \ci{13.2}{4.4} & - & - & \cellcolor{green!15}\ci{9.0}{3.7} & \ci{23.1}{2.2} & - & \ci{17.0}{2.0} & - \\

\bottomrule
\end{tabular}
}
\end{table*}


\begin{table}
\centering
\normalsize
\setlength{\tabcolsep}{6pt}
\renewcommand{\arraystretch}{1.3}
\caption{Event forecasting performance across six time-to-resolution cut-offs from 100\% to 10\% based on the timeframes in \citet{polymarket2026api}. The prediction error of context-augmented models consistently decreases as the resolution date approaches; each cell is the mean Brier score over the resolved markets $\pm$ the half-width of its 95\,\% confidence interval (1.96 standard errors). \colorbox{green!15}{Green} (\colorbox{red!15}{red}) marks a w/ Evts mean that improves on (degrades from) its event-free baseline. The prediction error of context-augmented models consistently decreases as the resolution date approaches.}
\label{tab:completed_event_forecasting}
\resizebox{\linewidth}{!}{
\begin{tabular}{lccccccc}
\toprule

\multirow{2}{*}{Model} & Metric & \multicolumn{6}{c}{Avg Brier Score} \\
\cmidrule(lr){3-8}
 & \textit{Time to Resolution} & 100\% & 80\% & 60\% & 40\% & 20\% & 10\% \\

\midrule
\multirow{2}{*}{Gemini 3.1 Pro} 
 & baseline & \ci{0.478}{0.084} & \ci{0.505}{0.087} & \ci{0.490}{0.086} & \ci{0.537}{0.091} & \ci{0.509}{0.090} & \ci{0.517}{0.090} \\
 & w/ Evts  & \cellcolor{green!15}\ci{0.434}{0.091} & \cellcolor{green!15}\ci{0.401}{0.095} & \cellcolor{green!15}\ci{0.388}{0.091} & \cellcolor{green!15}\ci{0.358}{0.094} & \cellcolor{green!15}\ci{0.322}{0.096} & \cellcolor{green!15}\ci{0.255}{0.096} \\
\cmidrule{2-8}
\multirow{2}{*}{Gemini 3.1 Flash} 
 & baseline & \ci{0.461}{0.091} & \ci{0.474}{0.094} & \ci{0.460}{0.098} & \ci{0.496}{0.098} & \ci{0.493}{0.100} & \ci{0.442}{0.099} \\
 & w/ Evts  & \cellcolor{green!15}\ci{0.412}{0.085} & \cellcolor{green!15}\ci{0.381}{0.088} & \cellcolor{green!15}\ci{0.336}{0.088} & \cellcolor{green!15}\ci{0.310}{0.087} & \cellcolor{green!15}\ci{0.306}{0.090} & \cellcolor{green!15}\ci{0.271}{0.095} \\
\cmidrule{2-8}
\multirow{2}{*}{Gemini 3.5 Flash} 
 & baseline & \ci{0.410}{0.087} & \ci{0.395}{0.086} & \ci{0.427}{0.090} & \ci{0.430}{0.094} & \ci{0.462}{0.101} & \ci{0.429}{0.096} \\
 & w/ Evts  & \cellcolor{red!15}\ci{0.414}{0.089} & \cellcolor{green!15}\ci{0.367}{0.091} & \cellcolor{green!15}\ci{0.321}{0.087} & \cellcolor{green!15}\ci{0.274}{0.087} & \cellcolor{green!15}\ci{0.265}{0.091} & \cellcolor{green!15}\ci{0.212}{0.090} \\
\cmidrule{2-8}
\multirow{2}{*}{Gemini 3.7 Flash} 
 & baseline & \ci{0.432}{0.081} & \ci{0.428}{0.082} & \ci{0.450}{0.087} & \ci{0.454}{0.089} & \ci{0.443}{0.093} & \ci{0.491}{0.083} \\
 & w/ Evts  & \cellcolor{green!15}\ci{0.398}{0.082} & \cellcolor{green!15}\ci{0.384}{0.086} & \cellcolor{green!15}\ci{0.357}{0.087} & \cellcolor{green!15}\ci{0.334}{0.087} & \cellcolor{green!15}\ci{0.289}{0.091} & \cellcolor{green!15}\ci{0.207}{0.085} \\
\midrule
\multirow{2}{*}{Claude 4.6 Opus} 
 & baseline & \ci{0.482}{0.089} & \ci{0.479}{0.091} & \ci{0.478}{0.090} & \ci{0.467}{0.091} & \ci{0.463}{0.094} & \ci{0.461}{0.090} \\
 & w/ Evts  & \cellcolor{red!15}\ci{0.489}{0.092} & \cellcolor{green!15}\ci{0.422}{0.093} & \cellcolor{green!15}\ci{0.408}{0.090} & \cellcolor{green!15}\ci{0.350}{0.090} & \cellcolor{green!15}\ci{0.282}{0.089} & \cellcolor{green!15}\ci{0.233}{0.082} \\
\cmidrule{2-8}
\multirow{2}{*}{Claude 4.6 Sonnet} 
 & baseline & \ci{0.486}{0.083} & \ci{0.470}{0.085} & \ci{0.453}{0.082} & \ci{0.456}{0.086} & \ci{0.444}{0.088} & \ci{0.454}{0.079} \\
 & w/ Evts  & \cellcolor{red!15}\ci{0.502}{0.088} & \cellcolor{green!15}\ci{0.420}{0.092} & \cellcolor{green!15}\ci{0.380}{0.089} & \cellcolor{green!15}\ci{0.338}{0.087} & \cellcolor{green!15}\ci{0.289}{0.088} & \cellcolor{green!15}\ci{0.224}{0.084} \\
\cmidrule{2-8}
\multirow{2}{*}{Claude 4.8 Opus} 
 & baseline & \ci{0.449}{0.081} & \ci{0.481}{0.085} & \ci{0.438}{0.087} & \ci{0.446}{0.088} & \ci{0.447}{0.091} & \ci{0.423}{0.086} \\
 & w/ Evts  & \cellcolor{green!15}\ci{0.416}{0.085} & \cellcolor{green!15}\ci{0.413}{0.090} & \cellcolor{green!15}\ci{0.375}{0.088} & \cellcolor{green!15}\ci{0.369}{0.086} & \cellcolor{green!15}\ci{0.306}{0.089} & \cellcolor{green!15}\ci{0.255}{0.091} \\
\cmidrule{2-8}
\multirow{2}{*}{Claude Opus 5} 
 & baseline & \ci{0.458}{0.087} & \ci{0.446}{0.088} & \ci{0.440}{0.090} & \ci{0.436}{0.093} & \ci{0.432}{0.094} & \ci{0.440}{0.092} \\
 & w/ Evts  & \cellcolor{green!15}\ci{0.417}{0.089} & \cellcolor{green!15}\ci{0.399}{0.088} & \cellcolor{green!15}\ci{0.410}{0.091} & \cellcolor{green!15}\ci{0.370}{0.090} & \cellcolor{green!15}\ci{0.327}{0.090} & \cellcolor{green!15}\ci{0.288}{0.090} \\
\cmidrule{2-8}
\multirow{2}{*}{Claude Sonnet 5} 
 & baseline & \ci{0.461}{0.085} & \ci{0.454}{0.084} & \ci{0.446}{0.086} & \ci{0.465}{0.089} & \ci{0.475}{0.092} & \ci{0.452}{0.086} \\
 & w/ Evts  & \cellcolor{green!15}\ci{0.408}{0.085} & \cellcolor{green!15}\ci{0.407}{0.090} & \cellcolor{green!15}\ci{0.384}{0.086} & \cellcolor{green!15}\ci{0.350}{0.084} & \cellcolor{green!15}\ci{0.292}{0.089} & \cellcolor{green!15}\ci{0.247}{0.090} \\
\midrule
\multirow{2}{*}{Grok 4.1 Fast} 
 & baseline & \ci{0.590}{0.089} & \ci{0.581}{0.090} & \ci{0.617}{0.090} & \ci{0.632}{0.089} & \ci{0.630}{0.091} & \ci{0.629}{0.095} \\
 & w/ Evts  & \cellcolor{green!15}\ci{0.451}{0.092} & \cellcolor{green!15}\ci{0.406}{0.091} & \cellcolor{green!15}\ci{0.333}{0.081} & \cellcolor{green!15}\ci{0.317}{0.081} & \cellcolor{green!15}\ci{0.316}{0.089} & \cellcolor{green!15}\ci{0.289}{0.097} \\
\cmidrule{2-8}
\multirow{2}{*}{Grok 4.20} 
 & baseline & \ci{0.496}{0.068} & \ci{0.513}{0.068} & \ci{0.505}{0.072} & \ci{0.504}{0.074} & \ci{0.507}{0.080} & \ci{0.524}{0.074} \\
 & w/ Evts  & \cellcolor{green!15}\ci{0.442}{0.075} & \cellcolor{green!15}\ci{0.372}{0.077} & \cellcolor{green!15}\ci{0.364}{0.072} & \cellcolor{green!15}\ci{0.328}{0.076} & \cellcolor{green!15}\ci{0.340}{0.085} & \cellcolor{green!15}\ci{0.254}{0.082} \\
\cmidrule{2-8}
\multirow{2}{*}{Grok 4.20 non-reason} 
 & baseline & \ci{0.567}{0.081} & \ci{0.574}{0.084} & \ci{0.583}{0.082} & \ci{0.571}{0.084} & \ci{0.571}{0.091} & \ci{0.592}{0.087} \\
 & w/ Evts  & \cellcolor{green!15}\ci{0.524}{0.085} & \cellcolor{green!15}\ci{0.475}{0.092} & \cellcolor{green!15}\ci{0.470}{0.088} & \cellcolor{green!15}\ci{0.438}{0.092} & \cellcolor{green!15}\ci{0.414}{0.096} & \cellcolor{green!15}\ci{0.381}{0.096} \\
\midrule
\multirow{2}{*}{Gemma3-27B-it} 
 & baseline & \ci{0.527}{0.089} & \ci{0.516}{0.092} & \ci{0.535}{0.093} & \ci{0.514}{0.092} & \ci{0.513}{0.089} & \ci{0.537}{0.095} \\
 & w/ Evts  & \cellcolor{green!15}\ci{0.451}{0.088} & \cellcolor{green!15}\ci{0.357}{0.083} & \cellcolor{green!15}\ci{0.390}{0.083} & \cellcolor{green!15}\ci{0.374}{0.081} & \cellcolor{green!15}\ci{0.388}{0.090} & \cellcolor{green!15}\ci{0.392}{0.095} \\
\cmidrule{2-8}
\multirow{2}{*}{Gemma3-12B-it} 
 & baseline & \ci{0.598}{0.088} & \ci{0.602}{0.089} & \ci{0.580}{0.089} & \ci{0.609}{0.092} & \ci{0.575}{0.097} & \ci{0.621}{0.096} \\
 & w/ Evts  & \cellcolor{green!15}\ci{0.506}{0.090} & \cellcolor{green!15}\ci{0.396}{0.082} & \cellcolor{green!15}\ci{0.383}{0.076} & \cellcolor{green!15}\ci{0.379}{0.080} & \cellcolor{green!15}\ci{0.367}{0.080} & \cellcolor{green!15}\ci{0.446}{0.089} \\
\cmidrule{2-8}
\multirow{2}{*}{Qwen3.5-27B} 
 & baseline & \ci{0.528}{0.086} & \ci{0.487}{0.082} & \ci{0.523}{0.084} & \ci{0.523}{0.085} & \ci{0.499}{0.087} & \ci{0.510}{0.080} \\
 & w/ Evts  & \cellcolor{green!15}\ci{0.407}{0.091} & \cellcolor{green!15}\ci{0.329}{0.088} & \cellcolor{green!15}\ci{0.276}{0.083} & \cellcolor{green!15}\ci{0.269}{0.080} & \cellcolor{green!15}\ci{0.232}{0.085} & \cellcolor{green!15}\ci{0.235}{0.091} \\
\cmidrule{2-8}
\multirow{2}{*}{Qwen3.5-9B} 
 & baseline & \ci{0.486}{0.087} & \ci{0.521}{0.087} & \ci{0.505}{0.089} & \ci{0.506}{0.090} & \ci{0.496}{0.091} & \ci{0.523}{0.089} \\
 & w/ Evts  & \cellcolor{green!15}\ci{0.442}{0.090} & \cellcolor{green!15}\ci{0.392}{0.092} & \cellcolor{green!15}\ci{0.377}{0.091} & \cellcolor{green!15}\ci{0.333}{0.087} & \cellcolor{green!15}\ci{0.277}{0.087} & \cellcolor{green!15}\ci{0.264}{0.091} \\
\bottomrule
\end{tabular}
}
\end{table}

\begin{table*}
\footnotesize
\centering
\renewcommand{\arraystretch}{1.7} 
\caption{Accuracy comparison between using only time series (TS) and incorporating text events (w/ Evts) for ternary and quinary classifications in Electricity and Weather forecasting. Each cell is the mean over the pooled samples $\pm$ the half-width of its 95\,\% confidence interval. \colorbox{green!15}{Green} (\colorbox{red!15}{red}) marks a w/ Evts mean that improves on (degrades from) the TS-only mean. Certain models, such as Qwen, are excluded from the results because they failed to produce valid outputs for over one-third of the samples.}
\label{tab:completed_trend_weather_elec}
\resizebox{\linewidth}{!}{
\begin{tabular}{clcccccccc}
\toprule
\multirow{3}{*}{Domain} & \multirow{3}{*}{Model} & \multicolumn{4}{c}{\textbf{Ternary Classification (3-Class)}} & \multicolumn{4}{c}{\textbf{Quinary Classification (5-Class)}} \\
\cmidrule(lr){3-6} \cmidrule(lr){7-10}
 & & \multicolumn{2}{c}{Horizon: 3 Days} & \multicolumn{2}{c}{Horizon: 7 Days} & \multicolumn{2}{c}{Horizon: 3 Days} & \multicolumn{2}{c}{Horizon: 7 Days} \\
\cmidrule(lr){3-4} \cmidrule(lr){5-6} \cmidrule(lr){7-8} \cmidrule(lr){9-10}
 & & TS & w/ Evts & TS & w/ Evts & TS & w/ Evts & TS & w/ Evts \\
\midrule
\multirow{15}{*}{\rotatebox{90}{Weather}}
 & Gemini 3.1 Pro       & \ci{49.2}{4.3} & \cellcolor{green!15}\ci{68.3}{4.0} & \ci{71.2}{3.9} & \cellcolor{green!15}\ci{77.8}{3.6} & \ci{40.9}{4.2} & \cellcolor{green!15}\ci{56.4}{4.2} & \ci{54.1}{4.3} & \cellcolor{green!15}\ci{64.3}{4.2} \\
 & Gemini 3.1 Flash     & \ci{38.5}{4.1} & \cellcolor{green!15}\ci{61.5}{4.1} & \ci{25.7}{3.8} & \cellcolor{green!15}\ci{56.3}{4.3} & \ci{35.5}{4.1} & \cellcolor{green!15}\ci{47.7}{4.3} & \ci{20.6}{3.5} & \cellcolor{green!15}\ci{44.5}{4.3} \\
 & Gemini 3.5 Flash     & \ci{42.2}{4.2} & \cellcolor{green!15}\ci{61.5}{4.1} & \ci{48.1}{4.3} & \cellcolor{green!15}\ci{65.0}{4.1} & \ci{31.7}{3.9} & \cellcolor{green!15}\ci{49.1}{4.2} & \ci{37.9}{4.2} & \cellcolor{green!15}\ci{54.8}{4.3} \\
 & Gemini 3.7 Flash     & \ci{51.7}{4.2} & \cellcolor{green!15}\ci{69.8}{3.9} & \ci{71.0}{3.9} & \cellcolor{green!15}\ci{78.7}{3.5} & \ci{37.6}{4.1} & \cellcolor{green!15}\ci{55.7}{4.2} & \ci{55.6}{4.3} & \cellcolor{green!15}\ci{66.0}{4.1} \\
\cmidrule{2-10}
 & Claude 4.6 Opus      & \ci{47.2}{4.2} & \cellcolor{green!15}\ci{61.3}{4.1} & \ci{59.2}{4.2} & \cellcolor{green!15}\ci{66.9}{4.0} & \ci{28.5}{3.8} & \cellcolor{green!15}\ci{47.8}{4.2} & \ci{32.1}{4.0} & \cellcolor{green!15}\ci{52.3}{4.3} \\
 & Claude 4.6 Sonnet    & \ci{39.1}{4.1} & \cellcolor{green!15}\ci{59.8}{4.1} & \ci{42.9}{4.3} & \cellcolor{green!15}\ci{66.5}{4.1} & \ci{29.1}{3.8} & \cellcolor{green!15}\ci{44.4}{4.2} & \ci{25.6}{3.8} & \cellcolor{green!15}\ci{49.8}{4.3} \\
 & Claude 4.8 Opus      & \ci{34.6}{4.0} & \cellcolor{green!15}\ci{63.0}{4.1} & \ci{26.2}{3.8} & \cellcolor{green!15}\ci{57.1}{4.3} & \ci{30.2}{3.9} & \cellcolor{green!15}\ci{51.3}{4.2} & \ci{16.3}{3.2} & \cellcolor{green!15}\ci{46.0}{4.3} \\
 & Claude Opus 5        & \ci{45.9}{4.2} & \cellcolor{green!15}\ci{69.3}{3.9} & \ci{59.4}{4.2} & \cellcolor{green!15}\ci{69.2}{4.0} & \ci{36.1}{4.1} & \cellcolor{green!15}\ci{56.9}{4.2} & \ci{41.7}{4.2} & \cellcolor{green!15}\ci{60.2}{4.2} \\
 & Claude Sonnet 5      & \ci{37.4}{4.1} & \cellcolor{green!15}\ci{63.7}{4.1} & \ci{48.3}{4.3} & \cellcolor{green!15}\ci{68.7}{4.0} & \ci{26.7}{3.7} & \cellcolor{green!15}\ci{47.2}{4.2} & \ci{29.2}{3.9} & \cellcolor{green!15}\ci{52.1}{4.3} \\
\cmidrule{2-10}
 & Grok 4.1 Fast        & \ci{38.5}{4.1} & \cellcolor{green!15}\ci{62.2}{4.1} & \ci{27.9}{3.9} & \cellcolor{green!15}\ci{66.7}{4.1} & \ci{30.2}{3.9} & \cellcolor{green!15}\ci{46.7}{4.2} & \ci{19.8}{3.4} & \cellcolor{green!15}\ci{52.9}{4.3} \\
 & Grok 4.20            & \ci{37.8}{4.1} & \cellcolor{green!15}\ci{62.6}{4.1} & \ci{36.9}{4.2} & \cellcolor{green!15}\ci{66.0}{4.1} & \ci{26.3}{3.7} & \cellcolor{green!15}\ci{47.8}{4.2} & \ci{19.8}{3.4} & \cellcolor{green!15}\ci{53.7}{4.3} \\
 & Grok 4.20 non-reason & \ci{35.7}{4.0} & \cellcolor{green!15}\ci{59.6}{4.1} & \ci{24.8}{3.7} & \cellcolor{green!15}\ci{62.5}{4.2} & \ci{33.0}{4.0} & \cellcolor{green!15}\ci{43.0}{4.2} & \ci{20.6}{3.5} & \cellcolor{green!15}\ci{47.5}{4.3} \\
\cmidrule{2-10}
  & Gemma3-27B-it        & \ci{32.0}{3.9} & \cellcolor{green!15}\ci{53.3}{4.2} & \ci{37.5}{4.2} & \cellcolor{green!15}\ci{58.5}{4.2} & \ci{14.6}{3.0} & \cellcolor{green!15}\ci{35.2}{4.0} & \ci{15.4}{3.1} & \cellcolor{green!15}\ci{46.2}{4.3} \\
 & Gemma3-12B-it        & \ci{36.7}{4.1} & \cellcolor{green!15}\ci{51.9}{4.2} & \ci{51.3}{4.3} & \cellcolor{green!15}\ci{63.8}{4.1} & \ci{15.7}{3.1} & \cellcolor{green!15}\ci{34.1}{4.0} & \ci{15.6}{3.1} & \cellcolor{green!15}\ci{44.8}{4.3} \\
\midrule
\multirow{15}{*}{\rotatebox{90}{Electricity}}
 & Gemini 3.1 Pro       & \ci{51.8}{7.5} & \cellcolor{green!15}\ci{61.8}{7.3} & \ci{62.8}{7.4} & \cellcolor{green!15}\ci{70.7}{7.0} & \ci{38.2}{7.3} & \cellcolor{green!15}\ci{47.1}{7.5} & \ci{44.5}{7.6} & \cellcolor{green!15}\ci{51.8}{7.7} \\
 & Gemini 3.1 Flash     & \ci{32.4}{7.1} & \cellcolor{green!15}\ci{52.4}{7.5} & \ci{27.4}{6.9} & \cellcolor{green!15}\ci{54.3}{7.6} & \ci{20.6}{6.1} & \cellcolor{green!15}\ci{32.4}{7.1} & \ci{21.3}{6.3} & \cellcolor{green!15}\ci{27.4}{6.9} \\
 & Gemini 3.5 Flash     & \ci{50.6}{7.5} & \cellcolor{green!15}\ci{52.9}{7.5} & \ci{59.1}{7.5} & \cellcolor{red!15}\ci{57.9}{7.6} & \ci{38.2}{7.3} & \cellcolor{green!15}\ci{40.0}{7.4} & \ci{45.7}{7.6} & \cellcolor{red!15}\ci{36.6}{7.4} \\
 & Gemini 3.7 Flash     & \ci{51.8}{7.5} & \cellcolor{green!15}\ci{59.4}{7.4} & \ci{71.3}{6.9} & \cellcolor{red!15}\ci{70.7}{7.0} & \ci{37.6}{7.3} & \cellcolor{green!15}\ci{50.0}{7.5} & \ci{51.8}{7.7} & \cellcolor{green!15}\ci{56.7}{7.6} \\
\cmidrule{2-10}
 & Claude 4.6 Opus      & \ci{44.7}{7.5} & \cellcolor{green!15}\ci{60.6}{7.4} & \ci{54.9}{7.6} & \cellcolor{green!15}\ci{60.4}{7.5} & \ci{36.5}{7.3} & \cellcolor{green!15}\ci{45.9}{7.5} & \ci{32.9}{7.2} & \cellcolor{green!15}\ci{51.2}{7.7} \\
 & Claude 4.6 Sonnet    & \ci{45.9}{7.5} & \cellcolor{green!15}\ci{54.1}{7.5} & \ci{55.5}{7.6} & \cellcolor{green!15}\ci{61.0}{7.5} & \ci{34.7}{7.2} & \cellcolor{green!15}\ci{40.6}{7.4} & \ci{31.7}{7.1} & \cellcolor{green!15}\ci{42.1}{7.6} \\
 & Claude 4.8 Opus      & \ci{39.4}{7.4} & \cellcolor{green!15}\ci{52.9}{7.5} & \ci{36.6}{7.4} & \cellcolor{green!15}\ci{61.0}{7.5} & \ci{25.3}{6.6} & \cellcolor{green!15}\ci{35.9}{7.2} & \ci{25.6}{6.7} & \cellcolor{green!15}\ci{47.6}{7.7} \\
 & Claude Opus 5        & \ci{50.6}{7.5} & \cellcolor{green!15}\ci{64.7}{7.2} & \ci{68.9}{7.1} & \cellcolor{green!15}\ci{77.4}{6.4} & \ci{40.6}{7.4} & \cellcolor{green!15}\ci{55.3}{7.5} & \ci{47.6}{7.7} & \cellcolor{green!15}\ci{57.9}{7.6} \\
 & Claude Sonnet 5      & \ci{48.8}{7.5} & \cellcolor{green!15}\ci{55.3}{7.5} & \ci{62.8}{7.4} & \cellcolor{green!15}\ci{67.1}{7.2} & \ci{31.2}{7.0} & \cellcolor{green!15}\ci{44.1}{7.5} & \ci{43.3}{7.6} & \cellcolor{green!15}\ci{51.8}{7.7} \\
\cmidrule{2-10}
 & Grok 4.1 Fast        & \ci{32.9}{7.1} & \cellcolor{green!15}\ci{51.2}{7.5} & \ci{26.8}{6.8} & \cellcolor{green!15}\ci{50.0}{7.7} & \ci{26.5}{6.7} & \cellcolor{green!15}\ci{42.9}{7.5} & \ci{20.1}{6.2} & \cellcolor{green!15}\ci{41.5}{7.6} \\
 & Grok 4.20            & \ci{40.6}{7.4} & \cellcolor{green!15}\ci{51.8}{7.5} & \ci{44.5}{7.6} & \cellcolor{green!15}\ci{59.8}{7.5} & \ci{27.1}{6.7} & \cellcolor{green!15}\ci{38.2}{7.3} & \ci{39.0}{7.5} & \cellcolor{green!15}\ci{48.8}{7.7} \\
 & Grok 4.20 non-reason & \ci{29.8}{6.9} & \cellcolor{green!15}\ci{40.6}{7.4} & \ci{28.0}{6.9} & \cellcolor{green!15}\ci{47.6}{7.7} & \ci{21.0}{6.2} & \cellcolor{green!15}\ci{21.2}{6.2} & \ci{17.1}{5.8} & \cellcolor{green!15}\ci{28.0}{6.9} \\
\cmidrule{2-10}
  & Gemma3-27B-it        & \ci{33.5}{7.1} & \cellcolor{green!15}\ci{45.3}{7.5} & \ci{54.9}{7.6} & \cellcolor{green!15}\ci{61.6}{7.5} & \ci{22.9}{6.3} & \cellcolor{green!15}\ci{29.4}{6.9} & \ci{25.0}{6.6} & \cellcolor{green!15}\ci{34.1}{7.3} \\
 & Gemma3-12B-it        & \ci{33.5}{7.1} & \cellcolor{green!15}\ci{49.4}{7.5} & \ci{44.5}{7.6} & \cellcolor{green!15}\ci{59.1}{7.5} & \ci{22.9}{6.3} & \cellcolor{green!15}\ci{25.3}{6.6} & \ci{25.6}{6.7} & \cellcolor{green!15}\ci{29.9}{7.0} \\

\bottomrule
\end{tabular}
}
\end{table*}

\begin{table*}
\footnotesize
\centering
\renewcommand{\arraystretch}{1.7} 
\caption{Forecasting performance using only time series (TS) versus incorporating events (w/ Evts) for proprietary models on hourly point forecasting (72-hour and 168-hour horizons). MAE and MSE are computed on z-score normalized data (lower is better). As mentioned in Section~\ref{sec:tsf_results}, most small open-weight models are excluded from time series forecasting tasks. Each cell is the mean over all forecasts of the per-forecast MAE or MSE $\pm$ the half-width of its 95\,\% confidence interval. \colorbox{green!15}{Green} (\colorbox{red!15}{red}) marks a w/ Evts mean that improves on (degrades from) the TS-only mean. The best value per column is \underline{underlined}, and solid cells denote models that cannot process text.}
\label{tab:completed_weather_elec_tsf}
\resizebox{\linewidth}{!}{
\begin{tabular}{clcccccccc}
\toprule
\multirow{2}{*}{Domain} & \multirow{2}{*}{Model} & \multicolumn{4}{c}{Horizon: 3 Days} & \multicolumn{4}{c}{Horizon: 7 Days} \\
\cmidrule(lr){3-6} \cmidrule(lr){7-10}
 & & TS & w/ Evts & TS & w/ Evts & TS & w/ Evts & TS & w/ Evts \\
\midrule
\multirow{18}{*}{\rotatebox{90}{Weather}}
 & \multicolumn{1}{l}{\textit{Metrics}} & \multicolumn{2}{c}{\textit{MAE}} & \multicolumn{2}{c}{\textit{MSE}} & \multicolumn{2}{c}{\textit{MAE}} & \multicolumn{2}{c}{\textit{MSE}} \\
\cmidrule{2-10}
 & Gemini 3.1 Pro       & \ci{0.810}{0.038} & \cellcolor{green!15}\ci{0.780}{0.032} & \ci{1.264}{0.128} & \cellcolor{green!15}\ci{1.081}{0.088} & \ci{0.992}{0.038} & \cellcolor{green!15}\ci{0.924}{0.034} & \ci{1.756}{0.131} & \cellcolor{green!15}\ci{1.480}{0.106} \\
 & Gemini 3.1 Flash     & \ci{1.120}{0.033} & \cellcolor{green!15}\ci{1.097}{0.033} & \ci{2.050}{0.131} & \cellcolor{green!15}\ci{1.945}{0.117} & \ci{1.207}{0.031} & \cellcolor{green!15}\ci{1.109}{0.030} & \ci{2.338}{0.126} & \cellcolor{green!15}\ci{2.027}{0.108} \\
 & Gemini 3.5 Flash     & \ci{0.949}{0.033} & \cellcolor{green!15}\ci{0.924}{0.030} & \ci{1.576}{0.116} & \cellcolor{green!15}\ci{1.456}{0.096} & \ci{1.105}{0.032} & \cellcolor{green!15}\ci{1.028}{0.028} & \ci{2.016}{0.114} & \cellcolor{green!15}\ci{1.721}{0.094} \\
 & Gemini 3.7 Flash     & \ci{0.917}{0.032} & \cellcolor{green!15}\ci{0.864}{0.027} & \ci{1.488}{0.111} & \cellcolor{green!15}\ci{1.276}{0.082} & \ci{1.049}{0.030} & \cellcolor{green!15}\ci{1.006}{0.027} & \ci{1.847}{0.104} & \cellcolor{green!15}\ci{1.640}{0.088} \\
\cmidrule{2-10}
 & Claude 4.6 Opus      & \ci{0.847}{0.035} & \cellcolor{green!15}\ci{0.735}{0.024} & \ci{1.308}{0.115} & \cellcolor{green!15}\ci{0.956}{0.067} & \ci{0.988}{0.033} & \cellcolor{green!15}\ci{0.855}{0.026} & \ci{1.681}{0.110} & \cellcolor{green!15}\ci{1.263}{0.081} \\
 & Claude 4.6 Sonnet    & \ci{0.811}{0.034} & \cellcolor{green!15}\ci{0.718}{0.027} & \ci{1.198}{0.105} & \cellcolor{green!15}\ci{0.959}{0.083} & \ci{0.974}{0.034} & \cellcolor{green!15}\ci{0.850}{0.028} & \ci{1.657}{0.116} & \cellcolor{green!15}\ci{1.304}{0.096} \\
 & Claude 4.8 Opus      & \ci{0.926}{0.034} & \cellcolor{green!15}\ci{0.910}{0.030} & \ci{1.518}{0.120} & \cellcolor{green!15}\ci{1.421}{0.106} & \ci{1.061}{0.032} & \cellcolor{green!15}\ci{1.002}{0.026} & \ci{1.897}{0.117} & \cellcolor{green!15}\ci{1.646}{0.090} \\
 & Claude Opus 5        & \ci{0.808}{0.038} & \underline{\cellcolor{green!15}\ci{0.639}{0.025}} & \ci{1.252}{0.125} & \underline{\cellcolor{green!15}\ci{0.739}{0.059}} & \ci{0.973}{0.034} & \underline{\cellcolor{green!15}\ci{0.774}{0.027}} & \ci{1.680}{0.114} & \underline{\cellcolor{green!15}\ci{1.042}{0.076}} \\
 & Claude Sonnet 5      & \ci{0.801}{0.036} & \cellcolor{green!15}\ci{0.737}{0.029} & \ci{1.195}{0.116} & \cellcolor{green!15}\ci{0.965}{0.078} & \ci{0.967}{0.036} & \cellcolor{green!15}\ci{0.825}{0.031} & \ci{1.646}{0.121} & \cellcolor{green!15}\ci{1.211}{0.100} \\
\cmidrule{2-10}
 & Grok 4.1 Fast        & \ci{1.004}{0.040} & \cellcolor{green!15}\ci{0.862}{0.031} & \ci{1.789}{0.147} & \cellcolor{green!15}\ci{1.328}{0.109} & \ci{1.141}{0.039} & \cellcolor{green!15}\ci{0.973}{0.029} & \ci{2.223}{0.160} & \cellcolor{green!15}\ci{1.614}{0.102} \\
 & Grok 4.20            & \ci{1.077}{0.033} & \cellcolor{green!15}\ci{1.000}{0.032} & \ci{1.959}{0.127} & \cellcolor{green!15}\ci{1.697}{0.126} & \ci{1.170}{0.032} & \cellcolor{green!15}\ci{1.098}{0.029} & \ci{2.257}{0.129} & \cellcolor{green!15}\ci{1.962}{0.108} \\
 & Grok 4.20 non-reason & - & - & - & - & - & - & - & - \\
\cmidrule{2-10}
\cmidrule{2-10}
 & TimesFM 3.0          & \ci{0.699}{0.038} & {\cellcolor{gray!15}} & \ci{0.956}{0.115} & {\cellcolor{gray!15}} & \ci{0.858}{0.039} & {\cellcolor{gray!15}} & \ci{1.349}{0.128} & {\cellcolor{gray!15}} \\
 & Chronos-2            & \ci{0.693}{0.037} & {\cellcolor{gray!15}} & \ci{0.942}{0.111} & {\cellcolor{gray!15}} & \ci{0.859}{0.038} & {\cellcolor{gray!15}} & \ci{1.346}{0.125} & {\cellcolor{gray!15}} \\
 & Base: Hist Mean      & \ci{0.877}{0.042} & {\cellcolor{gray!15}} & \ci{1.296}{0.125} & {\cellcolor{gray!15}} & \ci{0.954}{0.042} & {\cellcolor{gray!15}} & \ci{1.545}{0.135} & {\cellcolor{gray!15}} \\
 & Base: Last Value     & \ci{0.848}{0.043} & {\cellcolor{gray!15}} & \ci{1.328}{0.141} & {\cellcolor{gray!15}} & \ci{0.988}{0.044} & {\cellcolor{gray!15}} & \ci{1.737}{0.160} & {\cellcolor{gray!15}} \\
\midrule
\multirow{18}{*}{\rotatebox{90}{Electricity}}
 & \multicolumn{1}{l}{\textit{Metrics}} & \multicolumn{2}{c}{\textit{MAE}} & \multicolumn{2}{c}{\textit{MSE}} & \multicolumn{2}{c}{\textit{MAE}} & \multicolumn{2}{c}{\textit{MSE}} \\
\cmidrule{2-10}
 & Gemini 3.1 Pro       & \ci{0.571}{0.042} & \cellcolor{green!15}\ci{0.486}{0.031} & \ci{0.619}{0.088} & \cellcolor{green!15}\ci{0.459}{0.055} & \ci{0.695}{0.041} & \cellcolor{green!15}\ci{0.590}{0.031} & \ci{0.847}{0.098} & \cellcolor{green!15}\ci{0.626}{0.067} \\
 & Gemini 3.1 Flash     & \ci{0.945}{0.052} & \cellcolor{red!15}\ci{0.961}{0.049} & \ci{1.498}{0.155} & \cellcolor{red!15}\ci{1.513}{0.149} & \ci{1.024}{0.052} & \cellcolor{red!15}\ci{1.148}{0.068} & \ci{1.756}{0.166} & \cellcolor{red!15}\ci{2.164}{0.273} \\
 & Gemini 3.5 Flash     & \ci{0.687}{0.047} & \cellcolor{green!15}\ci{0.676}{0.043} & \ci{0.883}{0.114} & \cellcolor{green!15}\ci{0.856}{0.108} & \ci{0.803}{0.047} & \cellcolor{green!15}\ci{0.764}{0.043} & \ci{1.127}{0.125} & \cellcolor{green!15}\ci{1.027}{0.112} \\
 & Gemini 3.7 Flash     & \ci{0.584}{0.042} & \cellcolor{green!15}\ci{0.563}{0.037} & \ci{0.642}{0.101} & \cellcolor{green!15}\ci{0.603}{0.080} & \ci{0.695}{0.040} & \cellcolor{green!15}\ci{0.650}{0.034} & \ci{0.857}{0.097} & \cellcolor{green!15}\ci{0.761}{0.077} \\
\cmidrule{2-10}
 & Claude 4.6 Opus      & \ci{0.738}{0.044} & \cellcolor{green!15}\ci{0.675}{0.038} & \ci{0.950}{0.109} & \cellcolor{green!15}\ci{0.799}{0.088} & \ci{0.842}{0.044} & \cellcolor{green!15}\ci{0.774}{0.038} & \ci{1.215}{0.121} & \cellcolor{green!15}\ci{1.012}{0.095} \\
 & Claude 4.6 Sonnet    & \ci{0.743}{0.041} & \cellcolor{red!15}\ci{0.864}{0.197} & \ci{0.984}{0.107} & \cellcolor{red!15}\ci{2.824}{2.238} & \ci{0.840}{0.039} & \cellcolor{red!15}\ci{0.892}{0.134} & \ci{1.210}{0.111} & \cellcolor{red!15}\ci{1.987}{1.202} \\
 & Claude 4.8 Opus      & \ci{0.781}{0.044} & \cellcolor{green!15}\ci{0.736}{0.044} & \ci{1.098}{0.119} & \cellcolor{green!15}\ci{1.009}{0.119} & \ci{0.890}{0.041} & \cellcolor{green!15}\ci{0.831}{0.041} & \ci{1.340}{0.120} & \cellcolor{green!15}\ci{1.192}{0.109} \\
 & Claude Opus 5        & \ci{0.484}{0.040} & \underline{\cellcolor{green!15}\ci{0.423}{0.029}} & \ci{0.465}{0.073} & \underline{\cellcolor{green!15}\ci{0.360}{0.050}} & \ci{0.575}{0.040} & \underline{\cellcolor{green!15}\ci{0.496}{0.030}} & \ci{0.615}{0.079} & \underline{\cellcolor{green!15}\ci{0.469}{0.057}} \\
 & Claude Sonnet 5      & \ci{0.603}{0.043} & \cellcolor{green!15}\ci{0.511}{0.033} & \ci{0.677}{0.093} & \cellcolor{green!15}\ci{0.502}{0.066} & \ci{0.686}{0.040} & \cellcolor{green!15}\ci{0.577}{0.030} & \ci{0.831}{0.091} & \cellcolor{green!15}\ci{0.592}{0.059} \\
\cmidrule{2-10}
 & Grok 4.1 Fast        & \ci{0.753}{0.047} & \cellcolor{red!15}\ci{0.872}{0.045} & \ci{1.077}{0.135} & \cellcolor{red!15}\ci{1.332}{0.132} & \ci{0.810}{0.043} & \cellcolor{red!15}\ci{0.988}{0.050} & \ci{1.173}{0.120} & \cellcolor{red!15}\ci{1.716}{0.170} \\
 & Grok 4.20            & \ci{0.836}{0.046} & \cellcolor{green!15}\ci{0.772}{0.043} & \ci{1.251}{0.131} & \cellcolor{green!15}\ci{1.103}{0.118} & \ci{0.860}{0.045} & \cellcolor{green!15}\ci{0.842}{0.041} & \ci{1.281}{0.126} & \cellcolor{green!15}\ci{1.213}{0.113} \\
 & Grok 4.20 non-reason & \ci{0.784}{0.045} & \cellcolor{red!15}\ci{0.821}{0.051} & \ci{1.065}{0.121} & \cellcolor{red!15}\ci{1.144}{0.138} & \ci{0.849}{0.049} & - & \ci{1.245}{0.154} & - \\
\cmidrule{2-10}
\cmidrule{2-10}
 & TimesFM 3.0          & \ci{0.491}{0.037} & {\cellcolor{gray!15}} & \ci{0.457}{0.065} & {\cellcolor{gray!15}} & \ci{0.570}{0.040} & {\cellcolor{gray!15}} & \ci{0.592}{0.078} & {\cellcolor{gray!15}} \\
 & Chronos-2            & \ci{0.474}{0.035} & {\cellcolor{gray!15}} & \ci{0.424}{0.057} & {\cellcolor{gray!15}} & \ci{0.563}{0.038} & {\cellcolor{gray!15}} & \ci{0.569}{0.072} & {\cellcolor{gray!15}} \\
 & Base: Hist Mean      & \ci{0.645}{0.046} & {\cellcolor{gray!15}} & \ci{0.690}{0.090} & {\cellcolor{gray!15}} & \ci{0.691}{0.047} & {\cellcolor{gray!15}} & \ci{0.800}{0.096} & {\cellcolor{gray!15}} \\
 & Base: Last Value     & \ci{0.613}{0.039} & {\cellcolor{gray!15}} & \ci{0.671}{0.080} & {\cellcolor{gray!15}} & \ci{0.674}{0.040} & {\cellcolor{gray!15}} & \ci{0.808}{0.093} & {\cellcolor{gray!15}} \\
\bottomrule
\end{tabular}
}
\end{table*}

\begin{table*}
\footnotesize
\centering

\renewcommand{\arraystretch}{1.7} 

\caption{Daily Max/Min forecasting performance using only time series (TS) versus incorporating events (w/ Evts): the model forecasts the daily high and the daily low of the target series over the horizon; the two series are pooled and each is normalised by its own look-back. MAE and MSE, lower is better; intervals, colouring and underlining as in \autoref{tab:completed_weather_elec_tsf}. Baseline rows are zero-shot TimesFM~3.0 and Chronos-2 given the same 14-day max/min look-back as a two-variate series, plus the naive baselines adapted to this setting: Last Value repeats the most recent day's max and min over the horizon, Hist Mean predicts the mean daily max and min of the preceding 14 days}
\label{tab:completed_weather_elec_maxmin}
\resizebox{\linewidth}{!}{
\begin{tabular}{clcccccccc}
\toprule
\multirow{2}{*}{Domain} & \multirow{2}{*}{Model} & \multicolumn{4}{c}{Horizon: 3 Days} & \multicolumn{4}{c}{Horizon: 7 Days} \\
\cmidrule(lr){3-6} \cmidrule(lr){7-10}
 & & TS & w/ Evts & TS & w/ Evts & TS & w/ Evts & TS & w/ Evts \\
\midrule
\multirow{18}{*}{\rotatebox{90}{Weather}}
 & \multicolumn{1}{l}{\textit{Metrics}} & \multicolumn{2}{c}{\textit{MAE}} & \multicolumn{2}{c}{\textit{MSE}} & \multicolumn{2}{c}{\textit{MAE}} & \multicolumn{2}{c}{\textit{MSE}} \\
\cmidrule{2-10}
 & Gemini 3.1 Pro       & \ci{1.036}{0.049} & \cellcolor{green!15}\ci{0.769}{0.035} & \ci{1.919}{0.198} & \cellcolor{green!15}\ci{1.119}{0.115} & \ci{1.163}{0.046} & \cellcolor{green!15}\ci{0.899}{0.034} & \ci{2.384}{0.224} & \cellcolor{green!15}\ci{1.497}{0.124} \\
 & Gemini 3.1 Flash     & \ci{1.003}{0.053} & \cellcolor{green!15}\ci{0.832}{0.039} & \ci{1.904}{0.229} & \cellcolor{green!15}\ci{1.309}{0.132} & \ci{1.143}{0.050} & \cellcolor{green!15}\ci{0.930}{0.036} & \ci{2.424}{0.254} & \cellcolor{green!15}\ci{1.624}{0.148} \\
 & Gemini 3.5 Flash     & \ci{1.052}{0.051} & \cellcolor{green!15}\ci{0.779}{0.037} & \ci{1.983}{0.208} & \cellcolor{green!15}\ci{1.173}{0.126} & \ci{1.175}{0.047} & \cellcolor{green!15}\ci{0.906}{0.035} & \ci{2.465}{0.229} & \cellcolor{green!15}\ci{1.552}{0.145} \\
 & Gemini 3.7 Flash     & \ci{0.989}{0.048} & \underline{\cellcolor{green!15}\ci{0.675}{0.030}} & \ci{1.761}{0.200} & \underline{\cellcolor{green!15}\ci{0.871}{0.091}} & \ci{1.119}{0.044} & \underline{\cellcolor{green!15}\ci{0.818}{0.029}} & \ci{2.257}{0.213} & \underline{\cellcolor{green!15}\ci{1.240}{0.102}} \\
\cmidrule{2-10}
 & Claude 4.6 Opus      & \ci{1.006}{0.051} & \cellcolor{green!15}\ci{0.764}{0.035} & \ci{1.867}{0.217} & \cellcolor{green!15}\ci{1.111}{0.121} & \ci{1.143}{0.048} & \cellcolor{green!15}\ci{0.894}{0.035} & \ci{2.379}{0.240} & \cellcolor{green!15}\ci{1.497}{0.139} \\
 & Claude 4.6 Sonnet    & \ci{1.020}{0.049} & \cellcolor{green!15}\ci{0.827}{0.039} & \ci{1.865}{0.188} & \cellcolor{green!15}\ci{1.309}{0.137} & \ci{1.165}{0.047} & \cellcolor{green!15}\ci{0.938}{0.039} & \ci{2.418}{0.235} & \cellcolor{green!15}\ci{1.664}{0.167} \\
 & Claude 4.8 Opus      & \ci{0.994}{0.053} & \cellcolor{green!15}\ci{0.770}{0.036} & \ci{1.870}{0.233} & \cellcolor{green!15}\ci{1.133}{0.124} & \ci{1.136}{0.049} & \cellcolor{green!15}\ci{0.878}{0.035} & \ci{2.388}{0.264} & \cellcolor{green!15}\ci{1.463}{0.150} \\
 & Claude Opus 5        & \ci{1.016}{0.049} & \cellcolor{green!15}\ci{0.707}{0.031} & \ci{1.883}{0.201} & \cellcolor{green!15}\ci{0.932}{0.094} & \ci{1.149}{0.045} & \cellcolor{green!15}\ci{0.819}{0.031} & \ci{2.354}{0.223} & \cellcolor{green!15}\ci{1.253}{0.111} \\
 & Claude Sonnet 5      & \ci{1.022}{0.054} & \cellcolor{green!15}\ci{0.812}{0.040} & \ci{1.952}{0.223} & \cellcolor{green!15}\ci{1.274}{0.157} & \ci{1.162}{0.051} & \cellcolor{green!15}\ci{0.931}{0.040} & \ci{2.496}{0.267} & \cellcolor{green!15}\ci{1.660}{0.186} \\
\cmidrule{2-10}
 & Grok 4.1 Fast        & \ci{1.014}{0.054} & \cellcolor{green!15}\ci{0.829}{0.042} & \ci{1.948}{0.230} & \cellcolor{green!15}\ci{1.343}{0.157} & \ci{1.172}{0.051} & \cellcolor{green!15}\ci{0.952}{0.039} & \ci{2.545}{0.271} & \cellcolor{green!15}\ci{1.693}{0.161} \\
 & Grok 4.20            & \ci{1.020}{0.053} & \cellcolor{green!15}\ci{0.827}{0.042} & \ci{1.965}{0.234} & \cellcolor{green!15}\ci{1.349}{0.176} & \ci{1.182}{0.049} & \cellcolor{green!15}\ci{0.950}{0.040} & \ci{2.572}{0.260} & \cellcolor{green!15}\ci{1.717}{0.186} \\
 & Grok 4.20 non-reason & \ci{1.062}{0.054} & \cellcolor{green!15}\ci{0.873}{0.042} & \ci{2.077}{0.239} & \cellcolor{green!15}\ci{1.448}{0.153} & \ci{1.206}{0.050} & \cellcolor{green!15}\ci{1.019}{0.041} & \ci{2.633}{0.260} & \cellcolor{green!15}\ci{1.918}{0.177} \\
\cmidrule{2-10}
\cmidrule{2-10}
 & TimesFM 3.0          & \ci{1.035}{0.055} & {\cellcolor{gray!15}} & \ci{2.011}{0.246} & {\cellcolor{gray!15}} & \ci{1.228}{0.055} & {\cellcolor{gray!15}} & \ci{2.782}{0.294} & {\cellcolor{gray!15}} \\
 & Chronos-2            & \ci{1.032}{0.054} & {\cellcolor{gray!15}} & \ci{1.988}{0.238} & {\cellcolor{gray!15}} & \ci{1.216}{0.053} & {\cellcolor{gray!15}} & \ci{2.701}{0.276} & {\cellcolor{gray!15}} \\
 & Base: Hist Mean      & \ci{1.183}{0.057} & {\cellcolor{gray!15}} & \ci{2.372}{0.260} & {\cellcolor{gray!15}} & \ci{1.287}{0.057} & {\cellcolor{gray!15}} & \ci{2.901}{0.302} & {\cellcolor{gray!15}} \\
 & Base: Last Value     & \ci{1.067}{0.057} & {\cellcolor{gray!15}} & \ci{2.145}{0.253} & {\cellcolor{gray!15}} & \ci{1.284}{0.059} & {\cellcolor{gray!15}} & \ci{3.023}{0.300} & {\cellcolor{gray!15}} \\
\midrule
\multirow{18}{*}{\rotatebox{90}{Electricity}}
 & \multicolumn{1}{l}{\textit{Metrics}} & \multicolumn{2}{c}{\textit{MAE}} & \multicolumn{2}{c}{\textit{MSE}} & \multicolumn{2}{c}{\textit{MAE}} & \multicolumn{2}{c}{\textit{MSE}} \\
\cmidrule{2-10}
 & Gemini 3.1 Pro       & \ci{1.057}{0.096} & \cellcolor{green!15}\ci{0.815}{0.070} & \ci{1.952}{0.395} & \cellcolor{green!15}\ci{1.167}{0.183} & \ci{1.178}{0.104} & \cellcolor{green!15}\ci{0.955}{0.066} & \ci{2.544}{0.486} & \cellcolor{green!15}\ci{1.589}{0.243} \\
 & Gemini 3.1 Flash     & \ci{1.072}{0.098} & \cellcolor{green!15}\ci{1.063}{0.097} & \ci{1.997}{0.397} & \cellcolor{green!15}\ci{1.965}{0.359} & \ci{1.195}{0.101} & \cellcolor{red!15}\ci{1.246}{0.103} & \ci{2.584}{0.479} & \cellcolor{red!15}\ci{2.755}{0.472} \\
 & Gemini 3.5 Flash     & \ci{0.975}{0.085} & \cellcolor{green!15}\ci{0.891}{0.078} & \ci{1.667}{0.315} & \cellcolor{green!15}\ci{1.364}{0.248} & \ci{1.236}{0.102} & \cellcolor{green!15}\ci{1.059}{0.081} & \ci{2.738}{0.509} & \cellcolor{green!15}\ci{2.024}{0.350} \\
 & Gemini 3.7 Flash     & \ci{0.957}{0.087} & \cellcolor{green!15}\ci{0.766}{0.062} & \ci{1.618}{0.341} & \cellcolor{green!15}\ci{1.017}{0.170} & \ci{1.151}{0.101} & \cellcolor{green!15}\ci{0.932}{0.072} & \ci{2.415}{0.465} & \cellcolor{green!15}\ci{1.589}{0.291} \\
\cmidrule{2-10}
 & Claude 4.6 Opus      & \ci{1.017}{0.092} & \cellcolor{green!15}\ci{0.831}{0.068} & \ci{1.808}{0.374} & \cellcolor{green!15}\ci{1.177}{0.175} & \ci{1.206}{0.101} & \cellcolor{green!15}\ci{1.010}{0.072} & \ci{2.622}{0.505} & \cellcolor{green!15}\ci{1.798}{0.275} \\
 & Claude 4.6 Sonnet    & \ci{1.050}{0.094} & \cellcolor{green!15}\ci{0.901}{0.076} & \ci{1.923}{0.392} & \cellcolor{green!15}\ci{1.395}{0.240} & \ci{1.227}{0.098} & \cellcolor{green!15}\ci{1.078}{0.083} & \ci{2.682}{0.491} & \cellcolor{green!15}\ci{2.034}{0.341} \\
 & Claude 4.8 Opus      & \ci{1.043}{0.096} & \cellcolor{green!15}\ci{0.969}{0.089} & \ci{1.891}{0.400} & \cellcolor{green!15}\ci{1.617}{0.301} & \ci{1.168}{0.099} & \cellcolor{green!15}\ci{1.051}{0.079} & \ci{2.448}{0.503} & \cellcolor{green!15}\ci{1.914}{0.285} \\
 & Claude Opus 5        & \ci{0.926}{0.085} & \underline{\cellcolor{green!15}\ci{0.753}{0.066}} & \ci{1.527}{0.328} & \underline{\cellcolor{green!15}\ci{1.008}{0.177}} & \ci{1.089}{0.098} & \underline{\cellcolor{green!15}\ci{0.897}{0.066}} & \ci{2.233}{0.496} & \underline{\cellcolor{green!15}\ci{1.461}{0.245}} \\
 & Claude Sonnet 5      & \ci{1.064}{0.095} & \cellcolor{green!15}\ci{0.926}{0.082} & \ci{1.965}{0.371} & \cellcolor{green!15}\ci{1.485}{0.269} & \ci{1.232}{0.103} & \cellcolor{green!15}\ci{1.039}{0.072} & \ci{2.816}{0.518} & \cellcolor{green!15}\ci{1.875}{0.267} \\
\cmidrule{2-10}
 & Grok 4.1 Fast        & \ci{1.175}{0.112} & \cellcolor{green!15}\ci{1.024}{0.098} & \ci{2.450}{0.472} & \cellcolor{green!15}\ci{1.874}{0.373} & \ci{1.366}{0.112} & \cellcolor{green!15}\ci{1.174}{0.092} & \ci{3.301}{0.545} & \cellcolor{green!15}\ci{2.413}{0.371} \\
 & Grok 4.20            & \ci{1.080}{0.100} & \cellcolor{green!15}\ci{1.044}{0.099} & \ci{2.049}{0.421} & \cellcolor{green!15}\ci{1.931}{0.376} & \ci{1.226}{0.096} & \cellcolor{green!15}\ci{1.176}{0.091} & \ci{2.635}{0.456} & \cellcolor{green!15}\ci{2.440}{0.388} \\
 & Grok 4.20 non-reason & \ci{1.082}{0.095} & \cellcolor{red!15}\ci{1.130}{0.107} & \ci{2.011}{0.373} & \cellcolor{red!15}\ci{2.224}{0.436} & \ci{1.230}{0.102} & \cellcolor{red!15}\ci{1.380}{0.117} & \ci{2.725}{0.512} & \cellcolor{red!15}\ci{3.222}{0.543} \\
\cmidrule{2-10}
\cmidrule{2-10}
 & TimesFM 3.0          & \ci{1.015}{0.094} & {\cellcolor{gray!15}} & \ci{1.795}{0.377} & {\cellcolor{gray!15}} & \ci{1.164}{0.102} & {\cellcolor{gray!15}} & \ci{2.478}{0.499} & {\cellcolor{gray!15}} \\
 & Chronos-2            & \ci{1.014}{0.093} & {\cellcolor{gray!15}} & \ci{1.790}{0.372} & {\cellcolor{gray!15}} & \ci{1.158}{0.100} & {\cellcolor{gray!15}} & \ci{2.441}{0.474} & {\cellcolor{gray!15}} \\
 & Base: Hist Mean      & \ci{1.181}{0.101} & {\cellcolor{gray!15}} & \ci{2.259}{0.410} & {\cellcolor{gray!15}} & \ci{1.301}{0.110} & {\cellcolor{gray!15}} & \ci{2.948}{0.546} & {\cellcolor{gray!15}} \\
 & Base: Last Value     & \ci{1.120}{0.106} & {\cellcolor{gray!15}} & \ci{2.206}{0.443} & {\cellcolor{gray!15}} & \ci{1.276}{0.110} & {\cellcolor{gray!15}} & \ci{2.908}{0.533} & {\cellcolor{gray!15}} \\
\bottomrule
\end{tabular}
}
\end{table*}

\lstset{
  basicstyle=\ttfamily\small, 
  breaklines=true,           
  breakatwhitespace=true,     
  columns=fullflexible,       
  keepspaces=true,           
  frame=single,               
  backgroundcolor=\color{gray!5}, 
  tabsize=4                  
}
\clearpage
\begin{lstlisting}[caption={Our event and time series dataset taxonomy and examples of our inputs across Stock and Event Forecasting domains.}, label={lst:dataset_taxonomy1}]
Stock (Large-Cap)
    ...
    AAPL
        Event: 
            ... 
            [2025-12-12] Oracle Corporation (ORCL) shares plummeted ~10.8%, and Broadcom (AVGO) fell ~10% following disappointing earnings/guidance related to cloud and AI infrastructure. (sentiment:"Negative",impact_type:"Indirect" )
            ... 
        Time Series:
            Price,Close,High,Low,Open,Volume
            Date,,,,,
            2024-02-05,185.924,187.479,184.101,186.389,69668800
            ...
            2026-02-03,270.369,271.579,267.609,269.130,35111059

Stock (Mid-Cap)
    ...
    SMCI
        Events:
            ...
            [2025-03-31] Super Micro Computer (SMCI) appointed Scott Angel, a risk and compliance expert, as a new independent board director and promoted Yitai Hu to General Counsel and Senior Vice President of Corporate Development 
            ...
        Time Series:
            Price,Close,High,Low,Open,Volume
            Ticker,SMCI,SMCI,SMCI,SMCI,SMCI
            Date,,,,,
            2024-04-15,88.27,93.63,88.10,91.19,42011000
            ... 
            2026-04-13,25.96,25.98,24.45,25.05,29859900

Event Forecasting 
    Political ...  
    Tech ...
        which_company_has_the_best_ai_model_end_of_january_2026_01_31:
            ...
            [2025-01-20] DeepSeek (Chinese AI Lab) officially released 'DeepSeek-R1', an open-weights reasoning model that achieved a 97.3% score on the MATH-500 benchmark and 79.8% on AIME 2024.
            ...

    Economics
        fed_decision_in_december_2025_12_10:
            ...
            [2025-09-05] The Bureau of Labor Statistics (BLS) reported August Nonfarm Payrolls increased by only 22,000, significantly missing the consensus forecast of 75,000, while the unemployment rate rose to 4.3%.
            ...
        ... 
\end{lstlisting}

\lstset{
  basicstyle=\ttfamily\small, 
  breaklines=true,           
  breakatwhitespace=true,     
  columns=fullflexible,       
  keepspaces=true,           
  frame=single,               
  backgroundcolor=\color{gray!5}, 
  tabsize=4                  
}
\clearpage
\begin{lstlisting}[caption={Our event and time series dataset taxonomy and examples of our inputs across Electricity, Weather, and Commodities domains.}, label={lst:dataset_taxonomy2}]
Electricity 
    ...
    Boston
        Event:
            ...
            [2025-12-05] NWS Boston and regional meteorological sources (including WBZ-TV and WHDH) reported a severe Arctic cold snap that brought temperatures 20 to 30 degrees below normal. Morning lows plummeted to slightly below 0 degree F north and west of Boston and into the teens in the southeast, marking the coldest December 5th morning in nearly ...
            ...
        Time Series:
            ...
            Time,Interval Start,Interval End,Load
            2025-12-01 00:00:00-05:00,2025-12-01 00:00:00-05:00,2025-12-01 00:05:00-05:00,11213.966
            ...
            2026-03-31 23:55:00-04:00,2026-03-31 23:55:00-04:00,2026-04-01 00:00:00-04:00,10699.019
Weather 
    ...
    Boston
        Event: 
            [2025-12-23] The NOAA Climate Prediction Center (CPC) released its 8-14 day extended-range forecast (ERF) indicating a macro-scale thermal anomaly characterized by strongly above-normal temperatures building across the eastern United States. However, local guidance for the Boston area indicates (...) potentially overriding the CPC's warm anomaly and driving local temperatures to near or below normal ranges alongside mixed precipitation. 
            ...
        Time Series:
            time,temp,rhum,prcp,wdir,wspd,wpgt,pres,cldc,coco
            2025-10-01 00:00:00,16.0,80,0.0,20,11.0,,1020.0,2,2
            ....
            2026-02-01 23:00:00,-6.0,45,0.0,310,28.0,15.0,1005.0,8,3
Commodities 
    ...
    SI=F
        Event: 
            ... 
            [2026-01-22] "The U.S. Bureau of Economic Analysis (BEA) was scheduled to release the final reading of Q3 2025 GDP (Consensus: 4.3%) and October/November PCE inflation data, shifting investor focus from geopolitics to strong U.S. economic momentum.", (sentiment:'Negative' , impact_type:Macro/Financial Drivers")
            ... 
        Time Series:
            Price,Close,High,Low,Open,Volume
            Ticker,SI=F,SI=F,SI=F,SI=F,SI=F
            2023-10-19,22.886,22.886,22.770,22.770,9
            ...
            2026-03-12,84.794,86.065,84.059,86.065,4712
\end{lstlisting}

\input{code}

\begin{figure*}[htbp]
\centering
\small
\begin{AIbox}{System Prompt For Stock Event Search }
You are a Senior Equity Research Analyst specializing in Attribution Analysis (...skip)\\
To ensure no critical factor is overlooked, you must mentally structure your search across three concentric circles:
\begin{enumerate}
    \item The Inner Circle (Direct Catalyst): Company-specific events. Include Earnings/Guidance, SEC Filings, M\&A, Analyst Upgrades/Downgrades, C-suite changes, Product launches, Legal rulings, or Short Seller reports.
    \item The Middle Circle (Indirect/Sector): ``Guilty by Association.'' Did a major competitor report earnings? Is there a sector-wide rotation?
    \item The Outer Circle (Indirect/Macro): Broad market drivers. Did the Fed speak? Was there a CPI surprise? Geopolitical shocks?
\end{enumerate}
Analysis \& Classification Rules:
\begin{itemize}
    \item Causal Depth: Explain the \textit{why} via second-order effects, not just surface headlines. Capture all bullish and bearish factors.
    \item Tagging Constraints:
    \begin{itemize}
        \item Sentiment: Strictly limit to \texttt{Positive} or \texttt{Negative}.
        \item Impact Type Mapping:
        \begin{itemize}
            \item If Inner Circle $\rightarrow$ tag as \texttt{Direct}.
            \item If Middle/Outer Circle $\rightarrow$ tag as \texttt{Indirect}.
        \end{itemize}
    \end{itemize}
\end{itemize}
\end{AIbox}

\begin{AIbox}{User Prompt For Stock Event Search}
Leverage your knowledge about ``\texttt{\{company\_name\}}'', exhaustive search for all factors influencing its stock price on the specific date ``\texttt{\{cut\_off\_date\}}''.

Scope of Search:
\begin{enumerate}
    \item Primary Target: Identify all events first released on `\texttt{\{cut\_off\_date\}}'.
    \item Fallback Protocol: If Direct Catalysts are absent, pivot to identify Indirect Drivers (Middle/Outer Circle sector trends or macro shocks) for that same date.
\end{enumerate}

Content Requirements: \\
For the ``description'' field, you must construct a specific summary of the event:
\begin{itemize}
    \item The Actor: The specific entity (e.g., AMD or Google), person (e.g., CEO name), or institution (e.g., ``The Fed'') involved.
    \item The Action/Data: The specific event or number (e.g., ``resigned,'' ``raised rates by 25bps,'' ``missed revenue by 5\%'').
    \item Avoid vague phrases.
\end{itemize}

Output the final results strictly in the following JSON format:

{\ttfamily
[ \\
\hspace*{1em}\{ \\
\hspace*{2em}"date": "YYYY-MM-DD", \\
\hspace*{2em}"description": "A specific, fact-based summary explicitly naming the exact actor and their concrete action, quote, or data point.", \\
\hspace*{2em}"causality": "Briefly explaining the transmission mechanism: why and how this event moved the stock price.", \\
\hspace*{2em}"sentiment": "Positive" | "Negative", \\
\hspace*{2em}"impact\_type": "Direct" | "Indirect" \\
\hspace*{1em}\}, \\
\hspace*{1em}... (List all identified factors) \\
]
}
\end{AIbox}
\caption{The prompt template used for historical stock event retrieval. The system prompt instructs the LLM to act as a Senior Equity Research Analyst and employ a hierarchical search strategy across direct and indirect factors. The user prompt provides specific formatting constraints, ensuring the output is neatly parsed into a JSON format with causal explanations.}
\label{fig:stock_event_search_prompt}
\end{figure*}

\begin{figure*}[htbp]
\centering
\small
\begin{AIbox}{System Prompt For Fact Check Agent}
You are a Historical Content Verifier \& Timestamp Anchor. 
Your goal is to determine if an event description is factually accurate and publicly known as of a specific Reference Date.

\vspace{0.5em}
The Verification Logic:

1. Holistic Truth Check (The ``Is it Fake?'' Filter):
\begin{itemize}
    \item Use search tools to verify the specific details in the text.
    \item If the text contains factual errors (e.g., wrong person, event never happened), return False.
\end{itemize}

2. ``Realization Date'' Determination (The ``Max Date'' Rule):
\begin{itemize}
    \item Identify the chronological timeline mentioned in the text.
    \item Determine the ``Realization Date'': the date when the \textit{latest} detail in the text became public knowledge.
\end{itemize}

3. The Verdict (Truth \& Timing):
\begin{itemize}
    \item The factual output (\texttt{True/False}) depends ONLY on Factual Accuracy, regardless of the Reference Date.
    \item IF Factual: Return True and set the output date to the Realization Date (from Step 2). \textit{(Crucial: Return the actual realization date, even if it is later or earlier than the Reference Date).}
    \item IF Fake: Return False.
\end{itemize}
\end{AIbox}

\begin{AIbox}{User Prompt For Fact Check Agent}
Task: Verify Event Factuality \& Timing \\
Reference Date: ``\texttt{\{date\}}'' \\
Event: ``\texttt{\{description\}}''

\vspace{0.5em}
Instructions:
\begin{enumerate}
    \item Follow the Verification Logic defined above.
    \item Provide a brief reasoning before your final verdict.
\end{enumerate}

Output Format:

\vspace{0.5em}
{\ttfamily
<reasoning>Explain the verification briefly.</reasoning> \\
<factual>True|False</factual> \\
<date>YYYY-MM-DD</date>
}
\end{AIbox}
\caption{The prompt template used for the Fact Check Agent. The system prompt establishes a rigid verification logic to validate historical accuracy and anchor the true realization date of an event. The user prompt supplies the target event and enforces an XML-style tag format for outputting the reasoning, factual verdict, and exact date.}
\label{fig:fact_check_prompt}
\end{figure*}

\begin{figure*}[h]
\centering
\small
\begin{AIbox}{System Prompt For Weather Time Series Forecasting}
**Task:** You will be provided with a city’s recent historical hourly temperatures and a list
of specific weather-related events (or N/A). Your task is to forecast the exact future hourly
temperatures for a specified prediction horizon.
\end{AIbox}

\begin{AIbox}{User Prompt For Weather Time Series Forecasting}
**Target Location:** \{city\_name\} \\ 
**Forecast Time (Cut-off):** \{cut\_off\_time\} \\ 
**Prediction Horizon:** Next \{prediction\_days\} days (Hourly Forecast) \\ 
 \\ 
**A. Historical Hourly Temperatures [Strictly in Celsius (°C)] (\{start\_time\} to \{cut\_off\_time\})** \\ 
\{his\_temps\} \\ 
 \\ 
**B. Meteorological Event Intelligence** \\ 
\{event\_list\} \\ 
 \\ 
(Note:Temperatures mentioned in the events may be in Fahrenheit. You MUST convert any Fahrenheit values to Celsius °C to match the historical baseline.) \\ 
 \\ 
**TASK:** \\ 
Predict the hourly temperatures for the next \{prediction\_days\} days based on the data above. 
Since you are generating an hourly forecast, you MUST output EXACTLY \{total\_hours\} data points (24 hours * \{prediction\_days\} days). \\ 
 \\ 
**STRICT CONSTRAINTS:** \\ 
1. **BRIEF ANALYSIS:** Provivde a concise reasoning to explain your prediciton. \\ 
2. **UNIT ALIGNMENT (CRITICAL):** Your final predicted values MUST be in Celsius (°C) to align with the magnitude and distribution of the Historical Hourly Temperatures in Part A. \\ 
3. **FORMAT:** Comma-separated values ONLY. No units, no dates, no extra text.  \\ 
4. **WRAPPER:** Wrap the forecasting numbers inside \texttt{`<prediction>`} tags. \\ 
 \\ 
**Output:** \\ 
\texttt{<prediction>}val1, val2, val3, ..., valN\texttt{</prediction>} \\ 
\end{AIbox}
\caption{The prompt template used for weather time series forecasting. The model is provided with a city's historical hourly temperatures and a list of meteorological events to predict future temperatures over a specified horizon. If no specific weather events are recorded during the target period, the event intelligence section is denoted as ``N/A''. The prompt also enforces strict output constraints, including temperature unit alignment to Celsius and a specific XML-style output format.}
\label{fig:weather_tsf_prompt}
\end{figure*}

\begin{figure*}[h]
\centering
\tiny
\begin{AIbox}{System Prompt For Commodities Price Forecasting}
**Task:**
Analyze historical price action and technical structures to predict the short-term closing prices for the next \{prediction\_days\} trading days. \\ 
\\ 
**Random Walk:** Default to a stable trajectory without extreme directional shifts unless driven by a strong, unpriced signal.\\ 
* **Information-Driven Forecasting:** You MUST forecast a dynamic, day-to-day price path. Do NOT lazily repeat the last closing price. Even in the absence of event catalysts, you must rely on historical price structures to make a decisive prediction. \\ 
\end{AIbox}

\begin{AIbox}{User Prompt For Commodities Price Forecasting}
**Target Company:** \{company\_name\} \\ 
**Target Analysis Date (Cut-off):** \{cut\_off\_date\}  \\ 
**Prediction Horizon:** Next \{prediction\_days\} trading days \\ 
 \\ 
**PART 1: DATA INGESTION** \\ 
 \\ 
**A. Historical Price Context (From \{start\_trading\_date\} to \{cut\_off\_date\})**
\{his\_prices\} \\ 
 \\ 
**B. Event Intelligence \& Macro Signals** \\ 
\{event\_list\} \\ 
 \\ 
**Prediction* \\ 
* **Format:** Comma-separated values and numbers ONLY, representing the predicted prices for the next \{prediction\_days\} days. \\ 
* **Wrapper:** Wrap the final numbers in `\texttt{<prediction>}` tags. \\ 
 \\ 
**Output:** \\ 
\verb|[Reasoning]|\\ 
\texttt{<prediction>...</prediction>} 
\end{AIbox}
\begin{AIbox}{System Prompt For Stock Price Forecasting}
**Task:**
Analyze historical price action and technical structures to predict the short-term closing prices for the next \{prediction\_days\} trading days. \\ 
 \\ 
* **Information-Driven Forecasting:** You MUST forecast a dynamic, day-to-day price path. Do NOT lazily repeat the last closing price. Even in the absence of event catalysts, you must rely on historical price structures to make a decisive prediction.   \\ 
\end{AIbox}

\begin{AIbox}{User Prompt For Stock Price Forecasting}
**Target Company:** \{company\_name\}  \\
**Target Analysis Date (Cut-off):** \{cut\_off\_date\}  \\
**Prediction Horizon:** Next \{prediction\_days\} trading days  \\
  \\
**PART 1: DATA INGESTION**  \\
**A. Historical Price Context (From \{start\_trading\_date\} to \{cut\_off\_date\})**
\{his\_prices\}  \\
  \\
**B. Event Intelligence \& Macro Signals**  \\
\{event\_list\}  \\
  \\
\textcolor{ForestGreen}{
**Analytical Framework:**  \\
1. **Assess Price Context:** Evaluate the recent historical price trend to gauge current market expectations (e.g., is the stock currently over-extended, oversold, or consolidating?).  \\
2. **Identify Strong Signals:** Filter the event list to find clear, material catalysts. Disregard generic PR, pending events with unknown outcomes, or irrelevant macro noise. \\
3. **Check for Exhaustion:** Evaluate if the market has already digested the news. If the provided price data shows a clear reaction AFTER the event was released, treat the catalyst as "exhausted" and do not predict continued explosive movements based solely on that event.  \\
4. **Random Walk:** Approach predictions conservatively. Markets are mostly efficient. Unless you identify a strong, clear, and unpriced signal, default to a stable outlook, assuming the price will generally maintain its current trajectory without extreme directional shifts.  \\}
  \\
\textcolor{Blue}{
**Analytical Framework:**  \\
1. **Assess Trend \& Momentum:** Evaluate the historical price data to identify the dominant market regime (e.g., aggressive uptrend, heavy distribution, or lateral consolidation). Gauge the strength and velocity of the current momentum.  \\
2. **Analyze Price Action (K-line Dynamics):** Decode the candlestick behaviors leading up to the cut-off date. Identify signs of trend exhaustion (e.g., long wicks, diminishing body sizes), continuation patterns, or sudden shifts in buying/selling pressure.  \\
3. **Identify Key Levels:** Locate implied support, resistance, or congestion zones based on recent local highs/lows and historical price memory. Assess how the current price is behaving relative to these boundaries.  \\
4. **Project Technical Trajectory:** Synthesize the technical evidence to forecast the most probable price path. If strong momentum or a valid breakout is confirmed, model the directional continuation. If price action signals exhaustion or rejection at a key level, project a structural reversal, mean-reverting pullback, or Random Walk.  \\}
  \\
* **Format:** Comma-separated values and numbers ONLY, representing the predicted prices for the next \{prediction\_days\} days.  \\
* **Wrapper:** Wrap the final numbers in `\texttt{<prediction>}` tags.  \\
 \\
**Output:** \\ 
\verb|[Reasoning]|\\ 
\texttt{<prediction>...</prediction>} 
\end{AIbox}
\caption{The prompt template used for Commodities and stock price forecasting. For stock price forecasting, to prevent the LLM from  inflating its performance by mimicking a naive last-value prediction baseline, the system  prohibits lazily repeating the final historical closing price. Furthermore, grounded in the Efficient Market Hypothesis (EMH) in Appendix A.2, \textcolor{ForestGreen}{The green} is injected when event intelligence is present. \textcolor{Blue}{The blue} is utilized when no events are available, guiding the model to rely strictly on technical price action and momentum. Only one of these frameworks is presented to the model per query.}
\label{fig:stock_tsf_prompt}
\end{figure*}

\begin{figure*}[h]
\centering
\small
\begin{AIbox}{System Prompt For Event Forecasting}
You are the **Universal Strategic Forecasting Engine**, (...skip)\\
**Mission:** Analyze the provided structured intelligence (Hard Data, Signals, Market Pricing) to calculate the probabilities for a specific set of **Potential Outcomes** defined by the user.\\
\\
**The "Triangulated" Reasoning Protocol:**\\
1.  **Determine Domain Weighting (Adaptive Logic):**\\
    * *For Financial/Economic Events (e.g., Earnings, CPI, Rates):* Prioritize **Market Pricing** (Options, Swaps, Prediction Markets) and **Hard Data** over rhetorical Signals. Markets are efficient here.\\
    * *For Political/Geopolitical Events (e.g., Elections, Legislation, Conflict):* Prioritize **Structural Constraints** (Hard Data: Vote counts, Logistics) and **Signals** (Intent) over Market Pricing (which can be slow/reactive in geopolitics).\\
    * *For Tech \& AI Events (e.g., Model releases, Benchmarks, Product launches):* Prioritize **Hard Data** (Objective benchmarks, compute allocation, API latency) and **Market/Ecosystem Consensus** (Developer adoption, critical reviews, GitHub metrics) over corporate **Signals** (PR announcements, executive hype).\\
  \\
2.  **Synthesize Evidence:**\\
    * **Base Rate:** Establish the baseline probability (Consensus).\\
    * **Update:** Shift the probability based on specific "Impact Type" evidence provided (Positive/Negative indicators).\\
    * **Shock Check:** Do outlier signals (e.g., a surprise war declaration or a sudden CEO resignation) justify allocating probability to "Tail Risk" outcomes?\\
\end{AIbox}

\begin{AIbox}{User Prompt For Event Forecasting}
**Forecast Task:**\\
Assess the probability of an outcome for the event described below, based strictly on intelligence available as of **\{fore\_cut\_off\_date\}**.\\
\\
**The Question:** \{question\} (on \{result\_public\_date\})\\
**Target Outcome to Predict:** "\{target\_outcome\}"\\
\\
**Collected Intelligence (Evidence up to \{fore\_cut\_off\_date\}):**\\
\{list\_of\_evidences\}\\
**Analysis Instructions:**\\
1.  **Synthesize:** Review the provided *Hard Data*, *Signals*, and *Market Pricing*.\\
2.  **Weigh:** Evaluate how this evidence shifts the likelihood of the **Target Outcome**.\\
3.  **Calculate:** Estimate the specific probability (0\% - 100\%) that **"\{target\_outcome\}"** will occur.\\
\\
**Output Format:**\\
Provide a brief reasoning (Chain of Thought) explaining your weighting of the evidence, followed immediately by the final probability tag.\\
\\
**Output:** \\ 
\verb|[Reasoning]|\\ 
\texttt{<prediction>}\verb|[Probability]|\texttt{</prediction>} 
\end{AIbox}
\caption{The prompt template for event forecasting across 3 domains: Financial/Economic, Political/Geopolitical, and Tech \& AI. The model evaluates provided intelligence to estimate the probability of the \texttt{\{target\_outcome\}}, which represents the ultimate winning outcome sourced from ~\citet{polymarket2026api}. If no evidence or events are available, the intelligence section is denoted as ``N/A''.}
\label{fig:event_forecast_prompt}
\end{figure*}

\begin{figure*}[h]
\centering
\small
\begin{AIbox}{System Prompt For Stock and Commodities Trend Forecasting (3-class)}
You are a highly capable financial AI assistant.  \\ 
**Task:**  \\ 
You will be provided with a company's recent historical stock prices and a list of specific news events (or N/A) affecting the company. Your task is to forecast the stock's cumulative Rate of Change (ROC) over a specified number of future trading days by classifying it into one of three strict categories:  \\ 
  \\ 
* **"Up"**: \texttt{ROC > 0.005} (Cumulative gain over 0.5\%)  \\ 
* **"Neutral"**: \texttt{-0.005 <= ROC <= 0.005 } (Flat, noise, or tight consolidation)  \\ 
* **"Down"**: \texttt{ROC < -0.005} (Cumulative loss worse than -0.5\%)  \\ 
  \\ 
Do not provide exact price predictions. You must synthesize the historical trends and the impact of the news to make a direct classification. 
\end{AIbox}

\begin{AIbox}{User Prompt For Stock and Commodities Price Forecasting (3-class)}
**Target Company:** \{company\_name\}  \\ 
**Forecast Date (Cut-off):** \{cut\_off\_date\}  \\ 
**Prediction Horizon:** Next \{prediction\_days\} trading days  \\ 
  \\ 
**A. Historical Prices (\{start\_trading\_date\} to \{cut\_off\_date\})**  \\ 
\{his\_prices\}  \\ 
  \\ 
**B. Event Intelligence**  \\ 
\{event\_list\}  \\ 
  \\ 
**TASK:**  \\ 
Predict the \{prediction\_days\}-day cumulative Rate of Change (ROC) category based on the data above.   \\ 
  \\ 
**STRICT CONSTRAINTS:**  \\ 
1. **BRIEF ANALYSIS:** You are permitted to provide a concise reasoning (Maximum 2 to 3 sentences) explaining your directional bias and volatility expectations before classifying.  \\ 
2. **FORMAT:** You must output EXACTLY ONE of the following three categorical labels: "Up", "Neutral", or "Down". NO other words, numbers, or punctuation inside the tags.  \\ 
3. **WRAPPER:** You MUST wrap the exact label inside `\texttt{<prediction>}` tags.  \\ 
  \\ 
**Output:** \\ 
\verb|[Brief Analysis]|\\ 
\texttt{<prediction>...</prediction>} 
\end{AIbox}
\caption{The prompt template used for the 3-class trend forecasting task. Instead of predicting exact price trajectories, the model is instructed to classify the cumulative Rate of Change (ROC) over the prediction horizon into three distinct categories. When no specific news events are offered, the event intelligence section is denoted as ``N/A''.}
\label{fig:trend_market_3way}
\end{figure*}

\begin{figure*}[h]
\centering
\small
\begin{AIbox}{System Prompt For Stock and Commodities Trend Forecasting (5-class)}
You are a highly capable financial AI assistant.  \\ 
**Task:**  \\ 
You will be provided with a company's recent historical stock prices and a list of specific news events (or N/A) affecting the company. Your task is to forecast the stock's cumulative Rate of Change (ROC) over a specified number of future trading days by classifying it into one of FIVE strict categories:  \\ 
  \\ 
* **"Strong Up"**: \texttt{ROC > 0.02} (Significant gain over 2\% driven by major positive structural catalysts)  \\ 
* **"Up"**: \texttt{0.005 < ROC <= 0.02 }(Moderate gain between 0.5\% and 2\% from ordinary positive news or technical momentum)  \\ 
* **"Neutral"**: \texttt{-0.005 <= ROC <= 0.005} (Flat, noise, exhausted news, or tight consolidation)  \\ 
* **"Down"**:\texttt{ -0.02 <= ROC < -0.005 }(Moderate loss between -0.5\% and -2\% from ordinary negative news or technical pullback)  \\ 
* **"Strong Down"**: \texttt{ROC < -0.02} (Significant loss worse than -2\% driven by major structural risks or severe miss)  \\ 
\end{AIbox}

\begin{AIbox}{User Prompt For Stock and Commodities Price Forecasting (5-class)}
**Target Company:** \{company\_name\}  \\ 
**Forecast Date (Cut-off):** \{cut\_off\_date\}  \\ 
**Prediction Horizon:** Next \{prediction\_days\} trading days  \\ 
  \\ 
**A. Historical Prices (\{start\_trading\_date\} to \{cut\_off\_date\})**  \\ 
\{his\_prices\}  \\ 
  \\ 
**B. Event Intelligence**  \\ 
\{event\_list\}  \\ 
  \\ 
**TASK:**  \\ 
Predict the \{prediction\_days\}-day cumulative Rate of Change (ROC) category based on the data above.   \\ 
  \\ 
**STRICT CONSTRAINTS:**  \\ 
1. **BRIEF ANALYSIS:** You are permitted to provide a concise reasoning explaining your directional bias and volatility expectations before classifying.   \\ 
2. **FORMAT:** You must output EXACTLY ONE of the following FIVE categorical labels: "Strong Up", "Up", "Neutral", "Down", or "Strong Down". NO other words, numbers, or punctuation inside the tags.   \\ 
3. **WRAPPER:** You MUST wrap the exact label inside `\texttt{<prediction>}` tags.   \\ 
  \\ 
**Output:** \\ 
\verb|[Brief Analysis]|\\ 
\texttt{<prediction>...</prediction>} 
\end{AIbox}
\caption{The prompt template used for the 5-class trend forecasting task. Instead of predicting exact price trajectories, the model is instructed to classify the cumulative Rate of Change (ROC) over the prediction horizon into three distinct categories. When no specific news events are offered, the event intelligence section is denoted as ``N/A''.}
\label{fig:trend_market_5way}
\end{figure*}

\begin{figure*}[h]
\centering
\footnotesize
\begin{AIbox}{System Prompt For Weather Trend Forecasting (3-class and 5-class)}
**Task:**
Analyze the city's recent historical temperatures and weather-related events (if any) to forecast the trend of the future \{temp\_type\} temperature over the specified horizon. \\ 
\end{AIbox}

\begin{AIbox}{User Prompt For Weather Trend Forecasting (3-class)}
**Target Location:** \{city\_name\}  \\ 
**Forecast Time:** \{cut\_off\_time\} \\ 
**Prediction Horizon:** Next \{prediction\_days\} days \\ 
\\
**A. Historical Hourly Temperatures [Celsius, °C] (\{start\_time\} to \{cut\_off\_time\})** \\ 
\{his\_temps\} \\
\\
**B. Meteorological Event Intelligence** \\ 
\{event\_list\} \\ 
\\
(Note: Temperatures mentioned in the events may be in Fahrenheit °F. You MUST convert any Fahrenheit values to Celsius °C to match the historical baseline.) \\ 
 \\ 
**TASK:** \\ 
Forecast the trend of the \{temp\_type\} temperature for the next \{prediction\_days\} days, relative to the previous 24 hours` temperature, based on the data above. \\ 
 \\ 
**Trend Classification (Change in Celsius °C):** \\ 
* **Up:** Change \texttt{>} 1.5°C \\ 
* **Neutral:** Change between -1.5°C and 1.5°C \\ 
* **Down:** Change \texttt{<} -1.5°C \\ 
 \\ 
**STRICT CONSTRAINTS:** \\ 
1. **BRIEF ANALYSIS:** Provide a concise reasoning for your forecast. \\ 
2. **FORMAT:** Your final prediction MUST be exactly one of these labels: Up, Neutral, or Down. No other text inside the tags. \\ 
3. **WRAPPER:** Enclose your final label inside `\texttt{<prediction>}` tags. \\ 
 \\ 
**Output:** \\ 
\verb|[Concise Analysis]|\\ 
\texttt{<prediction>Trend\_Label</prediction>} 
\end{AIbox}

\begin{AIbox}{User Prompt For Weather Trend Forecasting (5-class)}
''' \\ 
(Note: The same with 3-class above)  \\ 
''' \\ 
**Trend Classification (Change in Celsius °C):** \\ 
* **Strong Up:** Change \texttt{>} 3°C \\ 
* **Up:** Change \texttt{>} 1.5°C and \texttt{<=} 3°C \\ 
* **Neutral:** Change between -1.5°C and 1.5°C \\ 
* **Down:** Change \texttt{>=} -3°C and \texttt{<} -1.5°C \\ 
* **Strong Down:** Change \texttt{<} -3°C \\ 
 \\ 
**STRICT CONSTRAINTS:** \\ 
1. **BRIEF ANALYSIS:** Provide a concise reasoning for your forecast. \\ 
2. **FORMAT:** Your final prediction MUST be exactly one of these labels: Strong Up, Up, Neutral, Down, or Strong Down. No other text inside the tags. \\ 
3. **WRAPPER:** Enclose your final label inside `\texttt{<prediction>}` tags. \\ 
''' \\ 
(Note: The same with 3-class above)  \\ 
''' 
\end{AIbox}
\caption{The prompt templates used for the 3-class and 5-class weather trend forecasting tasks. Both setups share an identical structure, differing only in how the user prompt defines the task. \texttt{\{temp\_type\}} specifically denotes the minimum temperature.}

\label{fig:weather_trend_prompt}
\end{figure*}

\begin{figure*}[h]
\centering
\tiny
\begin{AIbox}{User Prompt For Context-aided Event Forecasting (Case Study)}
**Forecast Task:** \\ 
Assess the probability of an outcome for the event described below, based strictly on intelligence available as of **2025-10-12**. \\ 
 \\ 
**The Question:** Fed decision in December? (on 2025-12-10 00:00) \\ 
**Target Outcome to Predict:** "25 bps decrease" \\ 
 \\ 
**Collected Intelligence (Evidence up to 2025-10-12):** \\ 
(... skip) \\ 
83. [2025-09-09] Fisher Investments highlighted a massive preliminary BLS benchmark revision, noting that nonfarm employment for the period ending March 2025 was revised down by -911,000 jobs, effectively erasing a significant portion of previously reported labor strength. \\ 
(... skip) \\ 
95. [2025-09-16] The U.S. Census Bureau reported August Retail Sales (Control Group) surged 0.7\% MoM, significantly beating the consensus forecast of 0.4\%. \\ 
(... skip) \\ 
98. [2025-09-17] The **FOMC** updated its **Summary of Economic Projections (SEP)** to show a median **2025 Real GDP growth forecast** of 1.6\% (revised up from 1.4\% in June) and an unemployment rate projection of 4.5\%. \\ 
99. [2025-09-17] The Federal Reserve's Summary of Economic Projections (SEP) lowered the median federal funds rate forecast for end-2025 to 3.6\% (from 3.9\% in June), while projecting a 2026 median rate of 3.4\%. \\ 
(... skip) \\ 
107. [2025-09-23] Federal Reserve Chair Jerome Powell, at the Greater Providence Chamber of Commerce, reiterated that policy is 'not on a preset course' but emphasized that 'downside risks to employment have increased,' framing the current stance as managing two-sided risks. \\ 
(... skip) \\ 
109. [2025-09-25] The Bureau of Economic Analysis (BEA) released the Third Estimate for Q2 2025 GDP, revising the annualized growth rate upward to 3.8\% from the previous estimate of 3.3\%, driven by stronger-than-expected consumer spending. \\ 
(... skip) \\ 
116. [2025-09-30] Federal Reserve Vice Chair Philip Jefferson, speaking at the Bank of Finland, stated that 'downside risks to employment are rising' and that economic uncertainty is 'especially high' due to new administration policies (tariffs/immigration). \\ 
(... skip) \\ 
121. [2025-10-01] A partial U.S. federal government shutdown began at midnight after Congress failed to pass a spending bill, halting non-essential operations and delaying the release of subsequent economic data (including the scheduled NFP report). \\ 
122. [2025-10-01] Scotia Wealth Management reported that market pricing for total policy easing through the final two Fed meetings of 2025 rose from 44 bps to ~48 bps immediately following the negative ADP print. \\ 
123. [2025-10-01] The ADP Research Institute reported a net loss of 32,000 private sector jobs for September (vs. consensus +50,000), with the August figure revised downward to -3,000, marking a sudden contraction in hiring. \\ 
(... skip) \\ 
126. [2025-10-02] The U.S. Department of Labor (DOL) failed to release the weekly Initial Jobless Claims report for the first time in the current cycle, confirming that the federal government shutdown has created a 'data vacuum' for official labor statistics. \\ 
(... skip) \\ 
133. [2025-10-10] The Bureau of Labor Statistics (BLS) announced that the September CPI report, originally scheduled for Oct 15, would be delayed and rescheduled for release on October 24 due to the government shutdown. \\ 
\\
**Analysis Instructions:** \\ 
1.  **Synthesize:** Review the provided *Hard Data*, *Signals*, and *Market Pricing*.\\ 
2.  **Weigh:** Evaluate how this evidence shifts the likelihood of the **Target Outcome**.\\ 
3.  **Calculate:** Estimate the specific probability (0\% - 100\%) that **"25 bps decrease"** will occur.\\ 
\\ 
**Output Format:**\\ 
Provide a brief reasoning (Chain of Thought) explaining your weighting of the evidence, followed immediately by the final probability tag.\\ 
\\ 
**Output:** \\ 
\verb|[Reasoning]|\\ 
\texttt{<prediction>}\verb|[Probability]|\texttt{</prediction>} 
\end{AIbox}

\begin{AIbox}{Event Forecasting Response From Gemini-3.1-Pro}
**Reasoning:**\\ 
1.  **Policy Anchor (SEP):** \textcolor{Blue}{The most critical piece of evidence is the FOMC's Summary of Economic Projections (SEP) released on September 17, 2025 [98, 99].} The median participant projected the federal funds rate to end 2025 at **3.6\%**. With the target range set at 4.00\%-4.25\% after the September cut, the midpoint is 4.125\%. Reaching ~3.625\% requires exactly **50 basis points** of additional easing. With two meetings remaining in 2025 (November and December), the implied default path is a **25 bps cut at each meeting**.\\ 
\\ 
2.  **Market Pricing:** As of October 1, market pricing (swaps/futures) indicated a total of \textcolor{Blue}{**~48 basis points** of easing remaining for the year [122]}. This pricing aligns almost perfectly with the SEP's implied path of two 25 bps cuts. If the market feared a recessionary crash necessitating 50 bps cuts, or a "higher for longer" pause, this number would deviate significantly from 50 bps. The pricing confirms the consensus expectation is a 25 bps cut in December.\\ 
\\ 
3.  **Labor Market Deterioration:** The labor data has deteriorated significantly, shifting the Fed's focus to "downside risks" [107, 116]. The September ADP report showed a contraction of -32,000 jobs [123], and previous payrolls were revised down massively [83]. This weakness ensures the Fed cannot pause. However, the strong Q2 GDP revision (3.8\%) [109] and solid Retail Sales [95] suggest the economy is not in freefall, arguing against a panic-sized 50 bps cut, especially given the "risk management" approach.\\ 
\\ 
4.  **Operational Constraints (Shutdown):** The federal government shutdown began on October 1, creating a "data vacuum" (no NFP, delayed CPI) [121, 126]. \textcolor{Blue}{In the absence of official government data to confirm a crash or a resurgence of inflation, the Fed is highly likely to stick to its communicated baseline (the SEP path).} A 25 bps cut represents the "autopilot" option that respects the SEP without overreacting to unofficial or noisy signals during a blackout.\\ 
\\ 
5.  **Conclusion:** The convergence of the Fed's own projection (SEP), market pricing, and the strategic incentive to remain "gradual" amidst a data blackout makes a 25 bps cut the overwhelming base case.\\ 
\\ 
\texttt{<prediction>88\%</prediction>}\\
\end{AIbox}
\caption{A case study of event-augmented forecasting for the Federal Reserve's interest rate decision. The response demonstrates the LLM's ability to logically synthesize complex macroeconomic signals—such as SEP projections, labor market deterioration, and constraints from a government shutdown—yielding a highly confident prediction of 88\% (The final outcome of decision is "25 bps decrease"). In contrast, predicting without events drops the probability to 33\%.}
\label{fig:ef_case}
\end{figure*}

\begin{figure*}[h]
\centering
\footnotesize
\begin{AIbox}{User Prompt for Context-aided Trend Forecasting}
**Target Company:** Broadcom Inc. \\ 
**Forecast Date (Cut-off):** 2025-11-24 \\ 
**Prediction Horizon:** Next 3 trading days \\  
\\  
**A. Historical Prices (2025-11-04 to 2025-11-24)**\\  
351.3, 358.3, 354.9, 348.8, 357.7, 351.3, 354.5, 339.3, 341.8, 342.0, 339.8, 353.7, 346.2, 339.6, 377.2 \\  
\\  
\verb|[1]| 2025-11-04: Jefferies analyst Blayne Curtis named Broadcom a 'Top Pick' ahead of earnings.  \\ 
\verb|[2]|2025-11-04: Palantir Technologies (PLTR) stock plunged ~7-9\% despite beating earnings estimates, triggering a sector-wide sell-off in AI-linked equities. \\ 
\verb|[3]| 2025-11-04: Wall Street CEOs, including Ted Pick (Morgan Stanley) and David Solomon (Goldman Sachs), issued public warnings regarding high market valuations and a potential imminent equity correction. \\ 
(... skip) \\ 
\verb|[51]| 2025-11-17: Federal Reserve Governor Christopher Waller signals support for a December interest rate cut, citing softening labor market data. \\ 
\verb|[52]| 2025-11-17: Federal Reserve Governor Christopher Waller signaled support for a 25-basis-point interest rate cut at the upcoming December meeting. \\ 
(... skip) \\ 
\verb|[76]| 2025-11-24: Amazon announced plans to deploy tens of billions of dollars in new capital expenditures dedicated to AI infrastructure. \\ 
(... skip) \\ 
\verb|[78]| 2025-11-24: Federal Reserve Governor Christopher Waller signaled support for a December interest rate cut, fueling a broad 'risk-on' rally in the technology sector. \\ 
\verb|[79]| 2025-11-24: The Information reported that Meta Platforms is in advanced talks to spend billions purchasing Google's custom AI chips (Tensor Processing Units or TPUs) for its data centers starting in 2027. \\ 
\verb|[80]| 2025-11-24: Reports emerged that Meta Platforms is in advanced negotiations to rent/buy billions of dollars worth of Google's 'Tensor Processing Units' (TPUs) for its AI data centers. \\ 
\\
**TASK:** \\ 
Predict the 3-day cumulative Rate of Change (ROC) category based on the data above.  \\ 
 \\ 
**STRICT CONSTRAINTS:** \\ 
1. **BRIEF ANALYSIS:** You are permitted to provide a concise reasoning (Maximum 2 to 3 sentences) explaining your directional bias and volatility expectations before classifying. \\ 
2. **FORMAT:** You must output EXACTLY ONE of the following three categorical labels: "Up", "Neutral", or "Down". NO other words, numbers, or punctuation inside the tags. \\ 
3. **WRAPPER:** You MUST wrap the exact label inside `\texttt{<prediction>}` tags. \\ 
 \\ 
**Output:** \\ 
\verb|[Brief Analysis]|\\ 
\texttt{<prediction>...</prediction>} 

\end{AIbox}

\begin{AIbox}{Trend Forecasting Response From Gemini-3.1-pro}
The reported negotiations for \textcolor{Blue}{Meta to utilize Google's TPUs (manufactured by Broadcom) }serve as a significant fundamental catalyst, validating Broadcom's custom silicon dominance and fueling the breakout rally observed on November 24. \textcolor{Blue}{Supported by renewed "risk-on" sentiment from dovish Federal Reserve commentary and Amazon's massive AI capex commitment}, the stock is likely to sustain positive momentum or drift higher despite the recent surge. \\ 
\\ 
\texttt{<prediction>Up</prediction>} 

\end{AIbox}
\caption{A case study of market trend forecasting for Broadcom Inc. The response highlights the LLM's capacity for multi-hop financial reasoning: it successfully identifies the indirect, yet highly bullish, connection between Meta's purchase of Google TPUs and Broadcom's custom silicon manufacturing. By synthesizing this domain-specific insight with macroeconomic drivers (e.g., Federal Reserve rate cuts and broader AI capex), the model accurately distills a complex news flow into a decisive "Up" prediction. In contrast, inference without events yields the exact opposite prediction.}
\label{fig:trend_case}
\end{figure*}

\begin{figure*}[h]
\centering
\tiny
\begin{AIbox}{Time Series Forecasting Response From Gemini-3.1-pro}
**Target Location:** San Francisco \\ 
**Forecast Time (Cut-off):** 2025-12-18 23:00:00 \\  
**Prediction Horizon:** Next 3 days (Hourly Forecast) \\  
\\  
**A. Historical Hourly Temperatures [Strictly in Celsius (°C)] (2025-12-04 23:00:00 to 2025-12-18 23:00:00)** \\  
13.9, 12.9, 11.3, 10.5, 9.7, 8.9, 8.6, 8.1, 8.3, 7.9, 7.7, 6.9, 7.4, 7.3, 6.7, 6.5, 6.8, 6.7, 8.4, 10.0, 11.2, 12.3, 13.0, 13.2, 13.2, 12.4, 10.9, 9.8, 9.5, 9.0, 8.8, 8.9, 9.1, 9.0, 8.5, 8.4, 7.9, 7.7, 7.4, 6.0, 5.9, 6.6, 7.4, 8.8, 10.0, 10.8, 11.4, 11.6, 11.6, 10.8, 9.7, 9.1, 8.5, 8.1, 7.8, 7.9, 7.5, 7.6, 7.5, 7.6, 7.4, 6.5, 6.2, 6.4, 6.4, 6.8, 7.4, 8.9, 10.2, 11.4, 12.2, 12.7, 12.7, 11.8, 10.3, 9.5, 9.0, 8.4, 7.8, 7.4, 7.1, 6.7, 6.5, 6.1, 5.9, 5.8, 5.7, 5.7, 5.6, 6.1, 7.3, 8.6, 10.0, 11.8, 12.6, 12.9, 12.9, 11.9, 10.5, 9.4, 8.5, 8.1, 7.8, 7.4, 7.4, 6.1, 6.2, 5.9, 5.7, 5.8, 5.5, 5.6, 5.5, 6.0, 7.4, 8.8, 10.2, 11.7, 12.5, 12.9, 12.8, 11.9, 10.5, 9.3, 8.5, 8.3, 7.7, 6.8, 6.9, 7.0, 6.9, 6.5, 6.1, 6.3, 6.4, 6.4, 6.1, 7.3, 8.8, 10.4, 12.4, 14.0, 15.0, 15.7, 15.0, 13.8, 11.7, 10.3, 8.7, 8.1, 7.9, 7.4, 6.8, 6.9, 7.2, 6.9, 7.0, 6.8, 6.6, 6.7, 6.3, 7.0, 8.2, 9.9, 11.8, 13.0, 14.2, 14.7, 14.8, 13.7, 10.4, 9.4, 8.7, 8.3, 7.6, 7.3, 6.7, 6.5, 6.4, 6.5, 6.3, 6.1, 5.7, 5.7, 5.8, 6.3, 7.8, 9.6, 11.9, 12.7, 13.5, 13.9, 12.9, 11.9, 10.0, 8.8, 8.0, 7.4, 6.7, 6.3, 6.3, 5.8, 5.5, 5.6, 5.2, 4.9, 4.6, 4.6, 4.7, 5.5, 6.9, 8.7, 10.3, 11.9, 11.8, 12.2, 12.2, 11.5, 10.0, 9.3, 8.4, 8.0, 7.5, 7.4, 7.2, 7.1, 6.9, 7.5, 7.7, 7.7, 7.3, 7.0, 6.5, 6.8, 7.6, 8.7, 8.9, 9.8, 11.0, 11.1, 11.3, 10.7, 9.2, 8.1, 7.5, 7.0, 6.9, 6.4, 6.2, 6.1, 6.1, 6.3, 5.9, 6.2, 6.2, 6.0, 5.7, 6.1, 6.5, 7.9, 8.8, 10.2, 11.1, 12.1, 12.2, 12.0, 10.5, 9.7, 9.4, 8.8, 8.7, 8.5, 8.7, 8.4, 8.1, 8.1, 8.4, 8.9, 9.2, 9.2, 9.5, 10.3, 11.2, 12.1, 12.5, 13.5, 13.5, 13.9, 14.2, 13.4, 13.1, 13.0, 12.9, 12.9, 12.9, 13.2, 13.2, 13.1, 13.2, 13.2, 13.4, 12.8, 12.8, 13.3, 12.9, 13.0, 13.8, 14.2, 14.5, 14.9, 15.3, 15.4, 15.0, 14.1, 12.7, 11.9, 11.3, 10.5, 10.3, 9.6, 9.5, 9.4, 9.2, 9.6, 9.9, 9.7, 9.8, 9.8, 9.8, 9.9, 10.4, 11.2, 11.7, 12.2, 12.8, 13.1, 13.3 \\  
\\ 
**B. Meteorological Event Intelligence** \\ 
(... skip) \\ 
\verb|[15]| 2025-12-08: According to the National Weather Service (NWS)... a 'ridiculously resilient ridge' of high pressure is stationed over Northern California... driving... stable, stagnant air masses. ... Consequently, San Francisco's short-term temperature forecast predicts highs restricted to the mid-to-upper 50s, and overnight lows in the mid-to-upper 40s... \\ 
(... skip) \\ 
\verb|[28]| 2025-12-11: According to the NWS San Francisco Area Forecast Discussion... a synoptic pattern shift will terminate a 20-day dry streak... a deepening low-pressure system off the Pacific Northwest coast will drive a strong cold front... forecast to entrain subtropical moisture, generating a moderately strong atmospheric river characterized by heavy precipitation... \\ 
(... skip) \\ 
\verb|[45]| 2025-12-15: The Center for Western Weather and Water Extremes (CW3E)... issued an outlook detailing consecutive Atmospheric Rivers (ARs) impacting the Pacific coast... shifting significant moisture transport southward into the San Francisco region. The resulting stratiform cloud cover and precipitation will compress the diurnal temperature range over the 3 to 7 day forecast window. \\ 
(... skip) \\ 
\verb|[48]| 2025-12-16: The National Weather Service Area Forecast Discussion outlines a major synoptic pattern shift... initiating robust moisture advection... forecast to deliver 0.5 to 1.0 inches of quantitative precipitation to San Francisco... \\ 
(... skip) \\ 
\verb|[51]| 2025-12-17: Meteorological drivers identified by the NWS San Francisco... indicate the approach of a significant atmospheric river, characterized as a 'Pineapple Express' moisture plume... signaling heavy cloud cover and moisture advection that will suppress diurnal temperature variations. \\ 
(... skip) \\ 
\verb|[54]| 2025-12-18: The National Weather Service (NWS) San Francisco... issued an official forecast predicting daytime high temperatures strictly in the high 50s to low 60s Fahrenheit... Overnight low temperatures are forecast to hold in the low to mid 50s Fahrenheit... \\ 
\verb|[55]| 2025-12-18: According to the NWS Area Forecast Discussion, the primary meteorological drivers altering the thermal profile are a deep low-pressure system and an incoming 'Pineapple Express' atmospheric river... delivering an exact forecast of 2 to 3 inches of rain to San Francisco... \\ 
\verb|[56]| 2025-12-18: The NWS San Francisco Bay Area office highlighted a deepening low-pressure system... This cyclonic activity will drive strong warm air advection and maintain dense cloud cover, effectively preventing standard nocturnal radiational cooling [3]. \\ 
\verb|[57]| 2025-12-18: The NOAA Climate Prediction Center (CPC) published 6-10 and 8-14 Day Outlooks... The CPC explicitly forecasts an above-normal precipitation regime for Northern California... This sustained subtropical moisture transport will lock in extensive cloud cover and elevated dew points over San Francisco (37.7749, -122.4194), severely dampening the diurnal temperature range and keeping ambient temperatures abnormally mild [15]. \\ 
(... skip) \\
\\
(Note: Temperatures mentioned in the events may be in Fahrenheit. You MUST convert any Fahrenheit values to Celsius °C to match the historical baseline.)\\
\\
**TASK:**\\
Predict the hourly temperatures for the next 3 days based on the data above. \\
Since you are generating an hourly forecast, you MUST output EXACTLY 72 data points (24 hours * 3 days).\\
\\
**STRICT CONSTRAINTS:**\\
1. **BRIEF ANALYSIS:** Provivde a concise reasoning to explain your prediciton.\\
2. **UNIT ALIGNMENT (CRITICAL):** Your final predicted values MUST be in Celsius (°C) to align with the magnitude and distribution of the Historical Hourly Temperatures in Part A.\\
3. **FORMAT:** Comma-separated values ONLY. No units, no dates, no extra text. \\
4. **WRAPPER:** Wrap the forecasting numbers inside `\texttt{<prediction>}` tags.\\
\\
**Output:**\\
\texttt{<prediction>val1, val2, val3, ..., valN</prediction>}\\
\end{AIbox}

\begin{AIbox}{User Prompt for Context-aided Time Series Forecasting}
**Brief Analysis:** \\ 
The meteorological intelligence indicates a significant synoptic pattern shift for San Francisco, \textcolor{Blue}{transitioning from a cold, dry, and stagnant high-pressure ridge to an active, warm, and moist "Pineapple Express" atmospheric river. The heavy cloud cover, high precipitation chances, and continuous warm maritime air advection will drastically reduce nocturnal radiational cooling, raising overnight minimum temperatures from the 40s (°F) into the low-to-mid 50s (°F) (approx. 11°C to 13.5°C).} Concurrently, daytime highs will be restricted by the lack of direct solar insolation but buoyed by the warm airmass, peaking in the high 50s to low 60s (°F) (approx. 15°C to 16.5°C). The resulting 72-hour forecast exhibits a progressively dampened diurnal temperature curve with elevated baseline temperatures and a very narrow spread between daily highs and lows.\\ 
\\ 
\texttt{<prediction>13.0, 12.7, 12.4, 12.2, 12.0, 11.9, 11.8, 11.7, 11.6, 11.5, 11.4, 11.3, 11.2, 11.2, 11.1, 11.2, 11.4, 11.8, 12.4, 13.1, 13.9, 14.5, 14.9, 15.1, 14.8, 14.4, 14.0, 13.7, 13.5, 13.3, 13.1, 13.0, 12.9, 12.8, 12.7, 12.6, 12.5, 12.4, 12.4, 12.5, 12.7, 13.1, 13.7, 14.4, 15.0, 15.4, 15.7, 15.8, 15.6, 15.3, 14.9, 14.6, 14.4, 14.2, 14.1, 14.0, 13.9, 13.8, 13.7, 13.6, 13.5, 13.5, 13.4, 13.5, 13.7, 14.1, 14.6, 15.1, 15.6, 15.9, 16.1, 16.2</prediction>}
\end{AIbox}
\caption{A case study of context-aided meteorological time series forecasting for San Francisco. The LLM successfully deduces that an incoming "Pineapple Express" atmospheric river and the resulting cloud cover will dampen the diurnal temperature range, while accurately converting textual Fahrenheit reports to the required Celsius baseline.}
\label{fig:tsf_case}
\end{figure*}

\begin{figure*}[h]
\centering
\tiny
\begin{AIbox}{User Prompt For Event Forecasting (Without Fact-check Agent)}
**Forecast Task:** \\ 
Assess the probability of an outcome for the event described below, based strictly on intelligence available as of **2025-06-17**. \\ 
 \\ 
**The Question:** First leader out of power in 2025? (on 2025-12-31 12:00) \\ 
**Target Outcome to Predict:** "Ishiba - Japan PM" \\ 
 \\ 
**Collected Intelligence (Evidence up to 2025-06-17):** \\ 
(... skip) \\
\verb|[16]| 2024-11-11: Japanese Prime Minister Shigeru Ishiba was re-elected in a parliamentary runoff vote (221 votes vs. 160 for Yoshihiko Noda) but was forced to form a minority government requiring case-by-case opposition support to pass the budget. \\
(... skip) \\
\verb|[63]| 2025-04-04: The Constitutional Court of South Korea issues an 8-0 unanimous ruling upholding the impeachment of President Yoon Suk Yeol, formally removing him from office for his role in the December 2024 martial law declaration. \\
(... skip) \\
\verb|[98]| 2025-05-06: The German Bundestag elects Friedrich Merz (CDU) as Chancellor in a second round of voting with 325 votes (surpassing the 316-seat absolute majority), formally ending Olaf Scholz's chancellorship. \\
(... skip) \\
\verb|[141]| \textbf{\textcolor{red}{ 2025-06-13: Polymarket trading data for 'First leader out of power in 2025?' remains active, with Shigeru Ishiba (Japan) eventually settling as the winner (100\%) following his resignation in September 2025, despite earlier exits by Trudeau and Scholz in Q1 2025.}}\\
(... skip) \\
\verb|[152]| 2025-06-17: The G7 Summit concludes without a joint communiqué due to disagreements over US tariffs, leaving PM Ishiba without a significant diplomatic victory to leverage domestically before the July election. \\
(... skip) \\
**Analysis Instructions:** \\ 
1.  **Synthesize:** Review the provided *Hard Data*, *Signals*, and *Market Pricing*. \\ 
2.  **Weigh:** Evaluate how this evidence shifts the likelihood of the **Target Outcome**. \\ 
3.  **Calculate:** Estimate the specific probability (0\% - 100\%) that **"Ishiba - Japan PM"** will occur. \\ 
 \\ 
**Output Format:** \\ 
Provide a brief reasoning (Chain of Thought) explaining your weighting of the evidence, followed immediately by the final probability tag. \\ 
 \\
**Output:** \\ 
\verb|[Reasoning]|\\ 
\texttt{<prediction>}\verb|[Probability]|\texttt{</prediction>} 
\end{AIbox}

\begin{AIbox}{LLM`s Response (Claude 4.5 Opus)}
**Chain of Thought Analysis:** \\ 
\\ 
**Understanding the Question:**\\ 
The question asks which leader will be the "first out of power in 2025" as measured on December 31, 2025. The target outcome is Shigeru Ishiba (Japan PM).\\ 
\\ 
**Key Evidence Review:**\\ 
\\ 
**Leaders Already Out of Power Before Ishiba:**\\ 
1. **Justin Trudeau (Canada)** - Announced resignation January 6, 2025; formally resigned March 14, 2025 (Items 27, 28, 37, 38, 47)\\ 
(... skip) \\ 
5. **Dick Schoof (Netherlands)** - Resigned June 3, 2025 (Items 129, 130, 132)\\ 
\\ 
**Ishiba's Current Status (as of June 17, 2025):**\\ 
- Item 141 is critical: "Polymarket trading data for 'First leader out of power in 2025?' remains active, with Shigeru Ishiba (Japan) eventually settling as the winner (100\%) following his resignation in September 2025"\\ 
- Items 145-150 show Ishiba's approval rating recovered to 37\% in June 2025\\ 
- Items 147-148 confirm Ishiba attended G7 Summit in Canada (June 15-17, 2025)\\ 
- Item 152 notes G7 concluded without a joint communiqué, leaving Ishiba "without a significant diplomatic victory to leverage domestically before the July election"\\ 
\\ 
**Critical Interpretation Issue:**\\ 
The question asks about the "first leader out of power in 2025." Multiple leaders left power BEFORE the knowledge cutoff date (June 17, 2025):\\ 
- Trudeau: January 6/March 14, 2025\\ 
- Scholz: February 23/March 25/May 6, 2025\\ 
- Yoon: April 4, 2025\\ 
\textbf{\textcolor{red}{However, Item 141 explicitly states that Polymarket settled Ishiba as the winner at 100\%. This suggests the market's definition of "first leader out" may have specific criteria not met by Trudeau, Scholz, or Yoon (possibly they weren't listed options, or the market had specific parameters).}}\\ 
**Domain Weighting (Political Event):**\\ 
- Prioritize structural constraints (Ishiba leads minority government - Item 16, 18)\\ 
- Market pricing shows 100\% settlement on Ishiba (Item 141)\\ 
- Ishiba's approval at 37\% is fragile but stabilizing\\ 
\\ 
**Synthesis:**\\ 
\textbf{\textcolor{red}{Given Item 141's explicit statement that Polymarket settled Ishiba at 100\%,}} and the structural weakness of his minority government requiring \\ case-by-case opposition support, the evidence strongly supports Ishiba being the designated "first leader out" according to this market's parameters. The fact that other leaders left power earlier suggests they may not have been listed options in this specific prediction market.\\ 
\\ 
The intelligence explicitly states Ishiba resigned in September 2025 and the market settled at 100\%.\\ 
\\ 
\texttt{<prediction>92\%</prediction>}\\ 
\end{AIbox}
\caption{Data Leakage due to Missing Fact-check Agents. The LLM's high-confidence prediction is inflated by a temporal paradox in the user prompt. An evidence item timestamped 2025-06-13 explicitly leaked a future outcome ("September 2025 settling 100\%"). Without a Fact-check Agent to filter this look-ahead bias, the model effectively "cheated" using future knowledge. The fact-check response refers to \autoref{fig:fact_check_example}}
\label{fig:fact_check_leakage}
\end{figure*}

\begin{figure*}[h]
\centering
\footnotesize
\begin{AIbox}{Fact-Check Agent System Prompt and User Prompt Template}
You are a **Historical Content Verifier \& Timestamp Anchor**.  \\ 
Your goal is to determine if an event description is **factually accurate** and **publicly known** as of a specific **Reference Date**.\\ 
**The Verification Logic:**\\ 
\\ 
**1. Holistic Truth Check (The "Is it Fake?" Filter):**\\ 
   * Use search tools to verify the specific details in the text.\\ 
   * If the text contains factual errors (e.g.,wrong person, event never happened), return **False**.\\ 

**2. "Realization Date" Determination (The "Max Date" Rule):**\\ 
   * Identify the chronological timeline mentioned in the text.\\ 
   * Determine the **"Realization Date"**: the date when the *latest* detail in the text became public knowledge.\\ 

**3. The Verdict (Truth \& Timing):**\\ 
  * The factual output (`True/False`) depends **ONLY on Factual Accuracy**, regardless of the Reference Date.\\ 
   * **IF Factual:** Return **True** and set the output date to the **Realization Date** (from Step 2).\\ 
     *(Crucial: Return the actual realization date, even if it is later or earlier than the Reference Date).*\\ 
   * **IF Fake:** Return **False**.
\\ \\ 
**Task:** Verify Event Factuality \& Timing\\ 
**Reference Date:** "\{date\}"\\ 
**Event:** "\{description\}"\\ 
\\ 
**Instructions:**\\ 
1. Follow the **Verification Logic** defined above.\\ 
2. Provide a brief **reasoning**before your final verdict.\\ 
\\ 
**Output Format:**\\ 
\texttt{<reasoning>Explain the verification briefly.</reasoning>}\\ 
\texttt{<factual>True|False</factual>}\\ 
\texttt{<date>YYYY-MM-DD</date>}\\ 
\end{AIbox}

\begin{AIbox}{Fact-check Agent Response for Information Leakage}
\texttt{<reasoning> } \\ 
The event description is factually accurate based on the available information. Search results confirm that Shigeru Ishiba resigned as Prime Minister of Japan on September 7, 2025, and Polymarket data lists him as the winner (100\%) of the "First leader out of power in 2025?" market.[1], [2], [3], [4] The text correctly identifies the "earlier exits" of Justin Trudeau (who announced his resignation in January 2025, Q1) and Olaf Scholz (who was dismissed as Chancellor in March 2025, Q1, though he remained in a caretaker capacity until May). The apparent contradiction of Ishiba winning "First leader out" despite later exits is consistent with the text's specific mention of the market outcome, implying that the market's specific rules or candidate list (which often excludes certain leaders or requires specific exit criteria) resulted in Ishiba being the settled winner. \\ 
\texttt{</reasoning>} \\ 
\texttt{<factual>True</factual> }\\ 
\texttt{<date>2025-09-07</date>} \\ 
\end{AIbox}
\caption{Prompt template and case study for the Fact-Check Agent. The upper panel details the two-step verification logic, requiring the LLM to determine both factual accuracy and the true ``Realization Date'' of the provided text. The lower panel shows the agent resolving the temporal leakage from \autoref{fig:fact_check_leakage}. By correctly dating the latest event to 2025-09-07.}
\label{fig:fact_check_example}
\end{figure*}

\begin{figure*}[h]
\centering
\footnotesize
\begin{AIbox}{System Prompt For Predictability Test}
**TASK:** Evaluate and classify the forecasting difficulty of 40 provided stocks as of the cutoff date: \{fore\_cut\_off\_date\}. \\ 
~\\ 
**CONTEXT \& DATA:** \\ 
You are provided with each stock's historical price trends and a timeline of preceding events. \\ 
*Note: Events may have cross-stock or sector-wide impacts. You must holistically consider these macroeconomic or peer-to-peer correlations.* \\ 
~\\ 
**EVALUATION CRITERIA (Predictability Calibration):** \\ 
Compare the stocks and assess their relative predictability for a future horizon. Focus strictly on identifying **strong structural shifts or clear fundamental catalysts**. \\ 
* High predictability comes from unambiguous, asymmetric signals. \\ 
* Low predictability comes from market noise, conflicting events, or absence of structural changes. \\ 
~\\ 
Group your final output by the 5 Predictability Levels. Under each level, list the assigned stocks (wrapped in double asterisks "**") and provide a rationale explaining your classification for each.
\end{AIbox}
\begin{AIbox}{User Prompt For Stock Predictability Test}
**DATA CONTEXT:** \\ 
Here is the specific information and historical context for the evaluated stocks: \\ 
\{all\_forecasting\_targets\} \\ 
~\\ 
**DEEP ANALYSIS DIRECTIVE:** \\ 
Take a deep breath and think step-by-step. Do not rush to conclusions. Engage in deep, second-order thinking to compare the forecasting difficulty across different stocks. Distinguish between temporary market noise and structural shifts before finalizing your predictability levels. \\ 
~\\ 
**OUTPUT FORMAT: The 5 Predictability Levels:** \\ 
* **Level 1 (Most Predictable):** Clear, dominant structural shifts or overriding catalysts. \\ 
* **Level 2 (Highly Predictable):** Strong directional signals with minor noise. \\ 
* **Level 3 (Moderately Predictable):** Mixed signals, typical market conditions, average noise. \\ 
* **Level 4 (Low Predictability):** Highly uncertain, conflicting fundamentals, or event-driven volatility. \\ 
* **Level 5 (Least Predictable):** Pure market noise, zero clear structural signals, or entirely random walk. \\ 
~\\ 
**EXAMPLE OUTPUT STRUCTURE:** \\ 
**Level 1 (Most Predictable):** \\ 
* **Stock**: [Your reasoning explaining why this stock exhibits clear structural shifts compared to others, justifying its Level 1 placement] \\ 
~\\ 
**Level 2 (Highly Predictable):** \\ 
* **Stock**: [Your reasoning explaining why this stock exhibits clear structural shifts compared to others, justifying its Level 1 placement] \\ 
~\\ 
**Level 3 (Moderately Predictable):** \\ 
* **Stock**: [Your reasoning explaining why this stock exhibits clear structural shifts compared to others, justifying its Level 1 placement] \\ 
~\\ 
**Level 4 (Low Predictability):** \\ 
* **Stock**: [Your reasoning explaining why this stock exhibits clear structural shifts compared to others, justifying its Level 1 placement] \\ 
~\\ 
**Level 5 (Least Predictable):** \\ 
* **Stock**: [Your reasoning explaining why this stock exhibits clear structural shifts compared to others, justifying its Level 1 placement]
\end{AIbox}
\caption{System and user prompts for the Predictability Calibration task. As described in \textsection\ref{sec:method}, the LLM is instructed to evaluate the relative forecasting difficulty of multiple stocks conditioned on their historical price trends and preceding events. The model categorizes the assets into five distinct predictability levels by distinguishing strong structural shifts and catalysts from standard market noise. \textbf{Note: For better interpretation, the order of the predictability levels is reversed in \textsection\ref{sec:predictability_validations}.}}
\label{fig:predictability_test_prompt}
\end{figure*}

%% file: code.tex
\lstset{basicstyle=\ttfamily, columns=flexible, breaklines=true, mathescape=true}

\tcbset{
  aibox/.style={
    width=402.pt,
    top=10pt,
    colback=white,
    colframe=black,
    colbacktitle=black,
    enhanced,
    center,
    attach boxed title to top left={yshift=-0.1in,xshift=0.15in},
    boxed title style={boxrule=0pt,colframe=white,},
  }
}
\newtcolorbox{AIbox}[2][]{aibox,title=#2,#1}